\documentclass[twoside,11pt]{article}
\usepackage{titletoc}

\usepackage{jmlr2e}

\usepackage[title]{appendix}
\usepackage{enumerate}
\usepackage{xcolor}
\usepackage{amsfonts, amsmath}
\usepackage{mathtools}
\usepackage{breakcites}
\usepackage{multirow, multicol, tabularx, threeparttable}
\usepackage{accents}
\usepackage{lipsum}
\usepackage{array}
\usepackage{cases} 
\usepackage{url} 
\usepackage{textcomp}
\usepackage[bottom]{footmisc}
\newcommand{\textappr}{\raisebox{0.5ex}{\texttildelow}}
\usepackage{booktabs}
\usepackage{algorithm}
\usepackage{algpseudocode}
\newcommand{\p}{{\rm I}\kern-0.18em{\rm P}}
\newcommand{\1}{{\rm 1}\kern-0.24em{\rm I}}
\newcommand{\E}{{\rm I}\kern-0.18em{\rm E}}
\DeclareMathOperator{\tr}{tr}

\makeatletter
\def\BState{\State\hskip-\ALG@thistlm}
\makeatother

\DeclareMathOperator*{\argmax}{arg\,max}

\newcommand{\R}{{\rm I}\kern-0.18em{\rm R}}
\newcommand{\N}{{\rm I}\kern-0.18em{\rm N}}

\usepackage[english]{babel}
\newcommand{\bX}{\boldsymbol{X}}

\newcommand{\dunif}{\text{discrete uniform}}

\newcommand{\truecomb}{\pi_{K^*}^*}
\newcommand{\escomb}{\pi_{K_{\rm{ES}}}^{\rm{ES}}}
\newcommand{\gscomb}{\pi_{K_{\rm{GS}}}^{\rm{GS}}}
\newcommand{\bfscomb}{\pi_{K_{\rm{BFS}}}^{\rm{BFS}}}
\newcommand{\partition}[2]{\pi_{#1}^{-1}(#2)}
\newcommand{\truepartition}[1]{\pi^{*-1}_{K^*}(#1)}

\newcommand{\genstirlingII}[3]{%
	\genfrac{\{}{\}}{0pt}{#1}{#2}{#3}%
}

\newcommand{\stirlingII}[2]{\genstirlingII{}{#1}{#2}}

\newtheorem{assumption}{Assumption}
\newcommand{\uderbar}[1]{\underset{\raise0.3em\hbox{$\smash{\scriptscriptstyle-}$}}{#1}}
\jmlrheading{X}{XXXX}{1-XX}{XX/XX}{XX/XX}{XXX}{Chihao Zhang, Yiling Elaine Chen, Shihua Zhang, and Jingyi Jessica Li}

\ShortHeadings{Information-theoretic Classification Accuracy}{Zhang, Chen, Zhang, and Li}
\firstpageno{1}
\begin{document}
	
	\title{Information-theoretic Classification Accuracy: A Criterion that Guides Data-driven Combination of Ambiguous Outcome Labels in Multi-class Classification}
	
	\author{\name Chihao Zhang \email zhangchihao@amss.ac.cn \\
		\addr Academy of Mathematics and Systems Science\\
	Chinese Academy of Sciences\\
	School of Mathematical Sciences\\
	University of Chinese Academy of Sciences
		\AND
		\name  Yiling Elaine Chen  \email yiling0210@ucla.edu  \\
	\addr  Department of Statistics\\
	University of California, Los Angeles 
		\AND
		\name Shihua Zhang \email zsh@amss.ac.cn \\
		\addr Academy of Mathematics and Systems Science\\
		Chinese Academy of Sciences\\
		School of Mathematical Sciences\\
		University of Chinese Academy of Sciences
		\AND 
		\name Jingyi Jessica Li  \email jli@stat.ucla.edu \\
		\addr Department of Statistics\\
		University of California, Los Angeles 
}
	
	\editor{XXX}
	
	\maketitle

\begin{abstract}
 Outcome labeling ambiguity and subjectivity are ubiquitous in real-world datasets.  
While practitioners commonly combine ambiguous outcome labels for all data points (instances) in an ad hoc way to improve the accuracy of multi-class classification, there lacks a principled approach to guide the label combination for all data points by any optimality criterion. To address this problem, we propose the information-theoretic classification accuracy (ITCA), a criterion that balances the trade-off between prediction accuracy (how well do predicted labels agree with actual labels) and classification resolution (how many labels are predictable), to guide practitioners on how to combine ambiguous outcome labels.  
To find the optimal label combination indicated by ITCA, we propose two search strategies: greedy search and breadth-first search. 
Notably, ITCA and the two search strategies are adaptive to all machine-learning classification algorithms. Coupled with a classification algorithm and a search strategy, ITCA has two uses: improving prediction accuracy and identifying ambiguous labels. We first verify that ITCA achieves high accuracy with both search strategies in finding the correct label combinations on synthetic and real data. Then we demonstrate the effectiveness of ITCA in diverse applications including medical prognosis, cancer survival prediction,  user demographics prediction, and cell type classification.
We also provide theoretical insights into ITCA by studying the oracle and the linear discriminant analysis classification algorithms. 
Python package \texttt{itca} (available at \url{https://github.com/JSB-UCLA/ITCA}) implements ITCA and the search strategies.
\end{abstract}

\begin{keywords}https://www.overleaf.com/project/621dee15e114a10d24c8ac90
multi-class classification, information theory, noisy labels, supervised learning, class label combination. 
\end{keywords}
\section{Introduction}
\par
Machine-learning prediction algorithms play an increasingly important role in data-driven, computer-based scientific research and industrial applications, thanks to the rapid advances in data availability, computing power, and algorithm development. 
Prominent examples include fraud detection based on historical transactions \citep{Brockett2002}, cardiovascular risk prediction \citep{Wilson1998,Weng2017}, and risk evaluation for multiple diseases using genomics data \citep{Chen2016}. 
Accurate algorithm prediction carries great promise because powerful algorithms can extract wisdom from human experts' numerous decisions made over the years.
\par
However, a bottleneck in the development of reliable algorithms is the availability of high-quality data, especially in medical diagnosis/prognosis and other biomedical applications. 
For example, medical records are inherently noisy, containing diagnostic/prognostic outcomes that are mislabeled or labeled inconsistently by graders \citep{Krause2018}. 
Further, labeling ambiguity is common for ordinal outcomes—whose ordered levels represent degrees of symptom severity or treatment effectiveness—because of graders’ subjectivity in assigning patients to levels.
\par
Ambiguous outcome labels would inevitably deteriorate the prediction accuracy of algorithms. 
Nevertheless, prediction accuracy may be boosted by combining the outcome labels that are hard to distinguish in training data, at the cost of losing classification resolution because label combination reduces the number of predictable outcome labels. 
Hence, how to find a balance between prediction accuracy and classification resolution is a computational challenge. 
Although outcome labels are often combined in an ad hoc way to train algorithms in practices \citep{Feldmann2000,Hemingway2013}, there lacks a principled approach to guide the combination by any optimality criterion.
\par
Besides outcome prediction, another critical application of machine learning is to refine the outcome labels that are predefined by human experts.
For example, in medical informatics, an important task is to use treatment outcomes to retrospectively refine diagnosis categories \citep{Lindenauer2012,Kale2018}. This task can be formulated as a multi-class classification problem, where the features are treatment outcomes and the response is a categorical variable indicating diagnosis categories. 
In this task, if patients in different diagnosis categories exhibit indistinguishable treatment outcomes, these categories should be combined. 
Such data-driven prediction has been used to update existing grading systems for diagnosis, such as the Gleason score for prostatic carcinoma \citep{Epstein2015}, the glomerular filtration rate (GFR) grade for chronic kidney disease, and the ACC/AHA classification for high blood pressure \citep{Muntner2018}.
For another example, in single-cell gene expression data analysis, a typical procedure is to cluster cells based on gene expression levels and subsequently annotate the cell clusters using domain knowledge \citep{Butler2018}.
This procedure is inevitably subjective because how to determine the number of clusters remains a challenge, and some cell clusters may be hardly distinguishable by gene expression levels.    
Hence, a principled method is called to guide the decision of combining ambiguous labels defined by human experts.
\par
How to find an ``optimal'' class combination is not a trivial problem. The reason is that, even if prediction is completely random, i.e., assigning data points with random labels irrespective of features, prediction accuracy would still be boosted by label combination. In such an extreme case, the increase in prediction accuracy does not outweigh the decrease in classification  resolution.
Hence, our rationale is that label combination must be guided by a criterion that reasonably balances prediction accuracy and classification resolution.

\par
Motivated by this rationale, we propose a criterion from an information theory perspective to evaluate prediction accuracy together with classification  resolution. 
This data-driven criterion, called the \textit{information-theoretic classification accuracy} (ITCA), can guide the combination of class labels given a multi-class classification algorithm. 
ITCA also allows choosing a multi-class classification algorithm among the available algorithms based on their respective optimal label combinations.
\par
There are three lines of research seemingly related to our work. 
	The first line is classification in the presence of labeling noise \citep{Frenay2013}.
	It includes three major approaches for handling labeling noise: (1) using robust losses or ensemble learning \citep{Freund2001, Beigman2009}; 
	(2) removing data points that are likely mislabeled \citep{Zhang2006,Thongkam2008}; 
	(3) modeling labeling noise using data generative models \citep{Swartz2004,Kim2008}.
	\par
	The second line is set-valued prediction, which predicts a set of labels, instead of a single label, for each data point. It includes two major approaches: (1) conformal prediction and (2) set-based utility maximization.
	Conformal prediction constructs a set of labels that contains the actual label with probability no less than the pre-specified confidence level \citep{Vovk2005, Balasubramanian2014}. 
	Set-based utility maximization constructs a set of labels that maximizes a set-valued utility function,
	which evaluates the utility of the set and typically decreases as the set's cardinality increases \citep{Corani2008, DelCoz2009, Zaffalon2012, Mortier2021}.
	\par
	The third line is nested dichotomies (ND), which recursively splits the classes into two subsets, inducing a binary tree of classes; then the multi-class classification problem can be solved recursively by a binary classification algorithm \citep{Frank2004, Leathart2016, Melnikov2018}. ND aims to find such a binary tree of classes that the multi-class classification accuracy is maximized \citep{Melnikov2018}.  
\par
ITCA differs from these three lines of research. 
First, ITCA is not specific to a loss, algorithm, or generative model, and ITCA does not require data removal.
Second, set-valued prediction approaches assume that the observed class labels are accurate at the global level, and they aim to find a specific class combination for each data point. 
As a result, they cannot suggest how to combine class labels at the global level for all data points.
Third, while the ND methods take a global approach and build a binary tree of classes so that the multi-class classification problem can be solved recursively by a binary classification algorithm, they cannot output an optimal class combination without an optimality criterion. 
\par
ITCA is a criterion for global class combination.
Given a multi-class prediction algorithm, say random forest, ITCA is defined as a weighted prediction accuracy, in which each data point is weighted by the entropy attributable to its class. 
As a result, ITCA balances the trade-off between prediction accuracy and classification resolution, thus offering guidance for finding an ``optimal" class label combination. 
In particular, ITCA can guide where to cut in an ND method's binary tree of classes so that a class combination can be determined. ITCA has broad applications, including medical diagnosis and prognosis, cancer survival prediction, user demographics prediction, and cell type classification. We will demonstrate these applications in Section~\ref{sec:applications}.
\par A prominent advantage of ITCA is its adaptivity  to all classification algorithms, thus allowing practitioners to choose the most suitable classification algorithm for a specific task.
As a side note, one may intuitively consider using a clustering algorithm to combine similar classes; for example, one may use the $K$-means algorithm or the hierarchical clustering algorithm to cluster the $K_0$ class centers into $K < K_0$ clusters, so that the $K_0$ observed classes are correspondingly combined into $K$ classes (see Section \ref{sec:exp}).
However, this intuitive approach has a drawback: since a distance metric is required to define the class centers and their distances, a gap exists between the choices of a metric and a classification algorithm. In other words, clustering-guided class combination based on a certain metric (e.g., Euclidean distance) does not guarantee to optimize the classification accuracy of a specific algorithm (e.g., support vector machine with Gaussian kernel). In contrast, ITCA does not have this drawback because it is defined based on the classification accuracy of the algorithm.   
\par
The rest of this paper is structured as follows.
In Section \ref{sec:method}, we first formulate the problem, define ITCA, and explain the intuition behind the definition. Then we introduce two search strategies---greedy search and breadth-first search---to find the optimal class combination guided by ITCA given a classification algorithm. 
In Section \ref{sec:exp}, we use extensive simulation studies to verify the effectiveness of ITCA and the two search strategies.
In Section \ref{sec:applications}, we demonstrate the broad applications of ITCA by applying it to multiple real-world datasets, including prognosis data of traumatic brain injury patients, glioblastoma cancer survival data, mobile phone user behavioral data, and single-cell RNA-seq data. In these applications, we also show the versatility of ITCA in working with various classification algorithms. Section~\ref{sec:conlustion} is the conclusion. Some key details are in Appendices. 
In Appendix~\ref{sec:alternative_criteria}, we propose five alternative criteria that may also guide class combination and are compared with ITCA.
In Appendix~\ref{sec:remarks}, we theoretically analyze the property of ITCA and the search strategies;
specifically, in Appendix~\ref{sec:p-ITCA}, we define ITCA at the population level and conduct theoretical analysis on two classification algorithms---the oracle and the linear discriminant analysis (LDA)---to characterize ITCA and provide insights for its use in practice; in Appendix~\ref{subsec:enhance}, we propose the soft LDA algorithm to improve LDA for finding the optimal combination defined by ITCA;
in Appendix~\ref{subsec:search property}, we theoretically analyze the optimality of the breadth-first search with the oracle algorithm.
In Appendix~\ref{sec:pruning}, we propose a pruning procedure to further reduce the search spaces of the two search strategies. Further details are provided in the Supplementary Material.

\section{Method}\label{sec:method}
\subsection{Problem formulation}
Let $(\bX, Y)\sim \mathcal{P}$ be a random pair where $\bX\in \mathcal{X} \subset \R^d$ is a feature vector, $Y \in [K_0] := \{1, \dots, K_0\}$ is a class label indicating one of $K_0$ observed classes that are potentially ambiguous, and $\mathcal{P}$ is the joint distribution of $(\bX, Y)$.
For a fixed positive integer $K$ $(< K_0)$, a class combination is represented
by an onto mapping: $\pi_K: [K_0] \to [K]$.
For example, if $K_0=4$ classes are combined into $K = 3$ classes by merging the original classes 3 and 4, then $\pi_3(1)=1$, $\pi_3(2)=2$, $\pi_3(3)=3$, and $\pi_3(4)=3$. 
We define $\pi_K^{-1}$ as follows: $\pi_K^{-1}(k):=\{k_0\in [K_0]:\pi_K(k_0)=k\}$, $\forall k\in[K]$.
Then in this example, $\pi_3^{-1} (1)=\{1\}$, $\pi_3^{-1}(2)=\{2\}$, $\pi_3^{-1} (3) =\{3,4\}$. 
For notation simplicity, we write $\pi_3$ as $\{1, 2, (3, 4)\}$.
Given a class combination $\pi_K$, a classification algorithm $\mathcal{C}$, and a training dataset $\mathcal{D}_t$, we denote by $\phi_{\pi_K}^{\mathcal{C},\mathcal{D}_t}: \mathcal{X}\to [K]$ a multi-class classifier trained by $\mathcal{C}$ on $\mathcal{D}_t$ to predict $K$ combined classes.
The prediction is accurate if and only if $\phi_{\pi_K}^{\mathcal{C},\mathcal{D}_t}(\bX)=\pi_K(Y)$. 
\par
\begin{figure}[hpt]
	\centering
	\includegraphics[width=\linewidth]{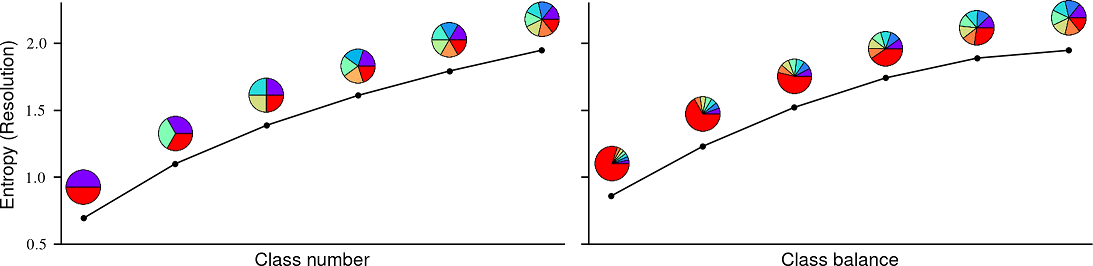}
	\caption{Entropy of the class label distribution reflects the classification resolution. Each pie chart indicates a class label distribution. Left: the number of balanced classes increases from left to right. 
		Right: the class label distribution is increasingly  balanced from left to right. The classification resolution increases from left to right in both plots, reflected by the entropy increase.\label{fig:entropy}}
\end{figure}

Given $K$, how to find an ``optimal" $\pi_K$ is a twofold problem. 
First, we need an optimality criterion of $\pi_K$ that balances the trade-off between prediction accuracy and classification resolution. 
Mathematically, given a dataset $\mathcal{D}:=\{(\bX_i, Y_i)\}_{i=1}^n$, a class combination $\pi_K$, and a classification algorithm $\mathcal{C}$, we split $\mathcal{D}$ into training data $\mathcal{D}_t$ and validation data $\mathcal{D}_v$, train a classifier $\phi_{\pi_K}^{\mathcal C, \mathcal{D}_t}$ on $\mathcal{D}_t$, and evaluate the prediction accuracy of $\phi_{\pi_K}^{\mathcal C, \mathcal{D}_t}$ on $\mathcal{D}_v$. 
Then we define an optimality criterion of $\pi_K$ given $\mathcal{D}_t$, $\mathcal{D}_v$, and $\mathcal{C}$, denoted by $m(\pi_K; \mathcal{D}_t, \mathcal{D}_v, \mathcal{C})$, based on the prediction accuracy of $\phi_{\pi_K}^{\mathcal C, \mathcal{D}_t}$ and the resolution of $\pi_K$'s $K$ combined classes. To define the classification resolution, we adopt the entropy concept in information theory.
The entropy of $\pi_K$'s $K$ combined classes' empirical distribution in $\mathcal{D}_v$ is $\sum_{k=1}^K \left[ -p_{\pi_K}^{\mathcal{D}_v}(k) \cdot \log p_{\pi_K}^{\mathcal{D}_v}(k)\right]$, where $p_{\pi_K}^{\mathcal{D}_v}(k) := \frac{1}{|\mathcal{D}_v|} \sum_{(\bX_i, Y_i) \in \mathcal{D}_v} \1 (\pi_K(Y_i)=k)$ is the proportion of the $k$-th combined class in $\mathcal{D}_v$.
Note that this entropy measures the uncertainty of $\pi_K(Y)$ based on its empirical distribution in $\mathcal{D}_v$, and it increases as $K$ increases or as the balance of the $K$ classes increases (Figure \ref{fig:entropy}). Hence, we find it reasonable to use this entropy to describe the resolution of $\pi_K$.
In Section~\ref{sec:ITCA_def}, we will define $m(\pi_K; \mathcal{D}_t, \mathcal{D}_v, \mathcal{C})$ as ITCA, a criterion that can be interpreted as an entropy-weighted accuracy, to balance the trade-off between prediction accuracy and classification resolution.

Second, given the criterion $m(\cdot; \mathcal{D}_t, \mathcal{D}_v, \mathcal{C})$, we need a search strategy to find the optimal class combination $\pi_K^*$ that maximizes the criterion:
\begin{equation}\label{eq:optimize}
	\pi_K^* = \argmax_{\pi_K\in \mathcal{A}} m(\pi_K; \mathcal{D}_t, \mathcal{D}_v, \mathcal{C})\,,
\end{equation}  
where $\mathcal{A}$ is the set of allowed class combinations. 
For example, if the outcome labels are ordinal, $\mathcal{A}$ contains all combinations that only combine adjacent class labels. We will address this search problem in Section~\ref{sec:search}.
\par

\subsection{Information-theoretic classification accuracy (ITCA)}\label{sec:ITCA_def}
\par
To explain the intuition of ITCA, we first introduce the classification accuracy (ACC), a widely-used evaluation criterion for a classifier. ACC is defined on a validation dataset $\mathcal{D}_v$ for a classifier $\phi_{\pi_K}^{\mathcal C, \mathcal{D}_t}$, which is trained by the algorithm $\mathcal C$ on the training dataset $\mathcal{D}_t$ given the combined classes defined by $\pi_K$.
\begin{equation}\label{eq:ACC-decompostion} 
\begin{split}
\text{ACC}(\pi_K; \mathcal{D}_t, \mathcal{D}_v, \mathcal{C})		&:= \frac{1}{|\mathcal{D}_v|} \sum\limits_{(\bX_i, Y_i)\in \mathcal{D}_v} \1\left(\phi_{\pi_K}^{\mathcal C, \mathcal{D}_t}(\bX_i)=\pi_k(Y_i)\right)\\
&= \sum_{k=1}^K p_{\pi_K}^{\mathcal{D}_v}(k) \cdot  \frac{\sum\limits_{(\bX_i, Y_i)\in \mathcal{D}_v} \1(\phi_{\pi_K}^{\mathcal C, \mathcal{D}_t}(\bX_i)=k,\, \pi_K(Y_i)=k)}{1 \bigvee \sum\limits_{(\bX_i, Y_i) \in \mathcal{D}_v} \1 (\pi_K(Y_i)=k)}\,,
\end{split}
\end{equation}
where $p_{\pi_K}^{\mathcal{D}_v}(k) := \frac{1}{|\mathcal{D}_v|} \sum\limits_{(\bX_i, Y_i) \in \mathcal{D}_v} \1 (\pi_K(Y_i)=k)$ is the proportion of $\pi_K$'s  $k$-th (combined) class in $\mathcal{D}_v$.
That is, ACC is a weighted sum of class-conditional out-of-sample prediction accuracies, with the $k$-th class weighted by its proportion $p_{\pi_K}^{\mathcal{D}_v}(k)$.
Note that ACC does not reflect the classification resolution because its maximal value of $1$ can be achieved by combining all classes into one, the scenario with the lowest classification resolution.
The reason is that the class weights in ACC are class proportions, making ACC dominated by the major classes with large proportions.
This issue motivates us to modify the class weights to reflect the classification resolution.
\par
\begin{figure}[hpbt]
	\includegraphics[width=\linewidth]{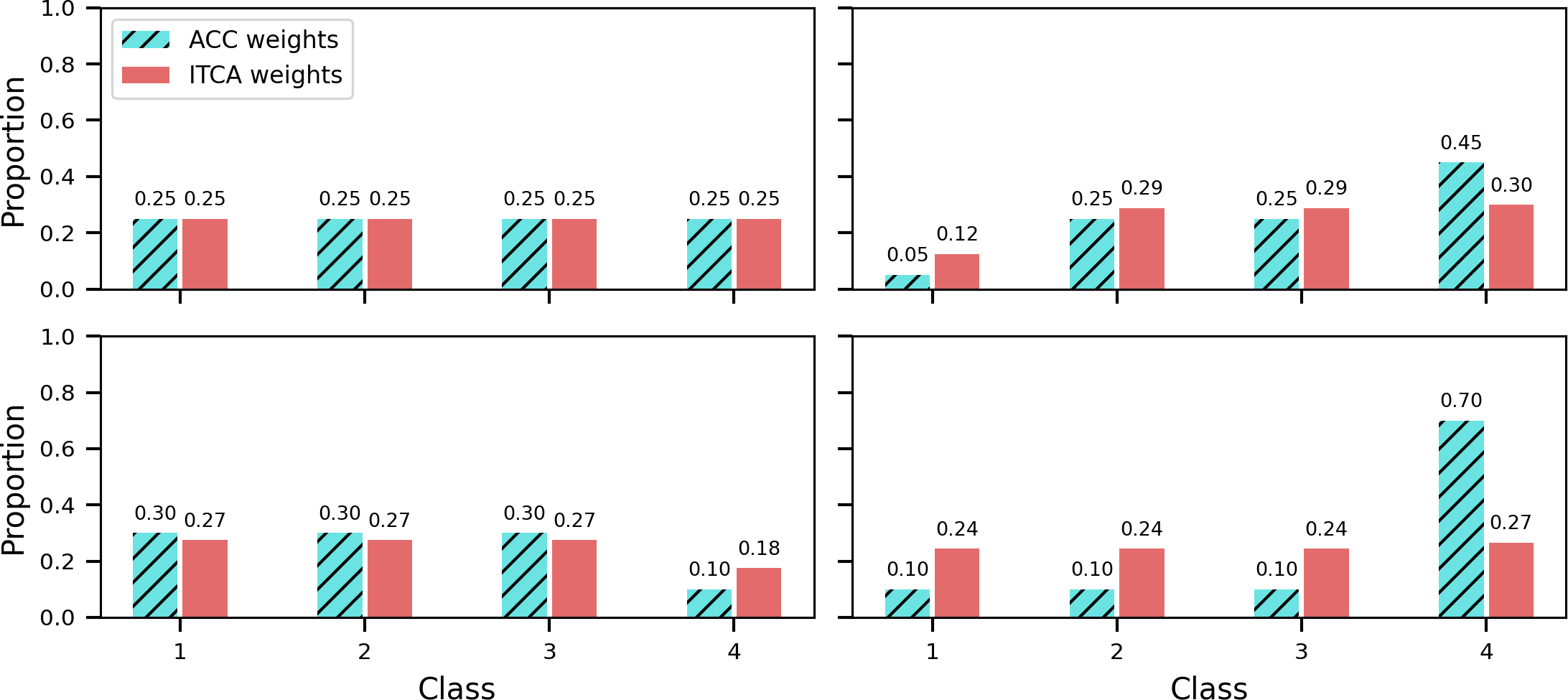}
	\caption{Comparison of class weights in ACC and ITCA under four scenarios of class proportions. The class weights (green bars with slanting lines) in ACC are equal to the class proprotions. The class weights (red bars) in ITCA are shown after being normalized to sum up to $1$. Mathematically, if the $k$-th class has weight $p_k$ in ACC, its weight would be $-p_k \log p_k$ in ITCA. }\label{fig:itca_acc_weights}
\end{figure}

 
Based on the ACC definition, we define ITCA by weighting each class using its contribution to the classification resolution, defined as the entropy of the distribution of $\pi_K(Y)$.
{\small

	\begin{equation}\label{eq:s-ITCA}
	\text{ITCA}(\pi_K; \mathcal{D}_t, \mathcal{D}_v, \mathcal{C})
		:= \sum_{k=1}^K \left[ -p_{\pi_K}^{\mathcal{D}_v}(k) \cdot \log p_{\pi_K}^{\mathcal{D}_v}(k)\right]
		\cdot
		\frac{\sum\limits_{(\bX_i, Y_i)\in \mathcal{D}_v} \1(\phi_{\pi_K}^{\mathcal C, \mathcal{D}_t}(\bX_i)=k,\, \pi_K(Y_i)=k)}{1 \bigvee \sum\limits_{(\bX_i, Y_i) \in \mathcal{D}_v} \1 (\pi_K(Y_i)=k)}\,.
	\end{equation}

}%
In this definition, $\pi_K$'s $k$-th combined class has weight $-p_{\pi_K}^{\mathcal{D}_v}(k) \cdot \log p_{\pi_K}^{\mathcal{D}_v}(k)$, i.e., its contribution to the classification resolution $\sum_{k=1}^K -p_{\pi_K}^{\mathcal{D}_v}(k) \cdot \log p_{\pi_K}^{\mathcal{D}_v}(k)$. Figure~\ref{fig:itca_acc_weights} illustrates the different class weights used in ACC and ITCA under four scenarios of class proportions. Intuitively, ITCA overweighs minor classes so it would be less dominated by major classes than ACC is.
\par
Note that the ITCA definition in (\ref{eq:s-ITCA}) is equivalent to 
\begin{equation}\label{eq:s-ITCA2}
		\text{ITCA}(\pi_K; \mathcal{D}_t, \mathcal{D}_v, \mathcal{C})  = \frac{1}{|\mathcal{D}_v|}\sum\limits_{(\bX_i, Y_i)\in \mathcal{D}_v}-\log p_{\pi_K}^{\mathcal{D}_v}(\pi_K(Y_i)) \cdot \1\left(\phi_{\pi_K}^{\mathcal C, \mathcal{D}_t}(\bX_i)=\pi_K(Y_i)\right)\,,
	\end{equation}
where every validation data point is weighted by the negative logarithm of the proportion of its (combined) class, unlike in ACC, where all validation data points have equal weights.
This alternative definition (\ref{eq:s-ITCA2}) has an intuitive interpretation. 
Imagine that we represent the $K_0$ original classes in $\mathcal{D}_v$ as $K_0$ non-overlapping intervals in $[0, 1]$ such that class $k_0$ is represented by an interval $I_{k_0}$ of length  equal to its proportion in $\mathcal{D}_v$; then $\cup_{k_0 =1}^{K_0} I_{k_0} = [0,1]$. With this representation, we can introduce a random variable $U \sim \mathrm{Uniform}([0,1])$ such that $Y = \sum_{k_0=1}^{K_0} k_0 \1(U \in I_{k_0})$; that is, the event $U \in I_{k_0}$ is equivalent to the event $Y=k_0$. For a data point $\bX_i \in \mathcal{D}_v$, when $\phi_{\pi_K}^{\mathcal C, \mathcal{D}_t}$ correctly predicts its combined class label $\pi_K(Y_i)$, i.e., $\phi_{\pi_K}^{\mathcal C, \mathcal{D}_t}(\bX_i)=\pi_K(Y_i)$, we weight this correct prediction in (\ref{eq:s-ITCA2}) by difficulty, which should be higher if the combined class $\pi_K(Y_i)$ is smaller, i.e., the total length of $\cup_{k_0 \in \pi_K^{-1}(\pi_K(Y_i))}I_{k_0}$ is shorter. Hence, we define the weight as the difference between the entropy of $\mathrm{Uniform}([0,1])$ (i.e., the distribution of $U_i$ without knowledge of $Y_i$) and the entropy of $\mathrm{Uniform}(\cup_{k_0\in \pi_K^{-1}(\pi_K(Y_i)) } I_{k_0})$ (i.e., the distribution of $U_i$ conditional on $\pi_K(Y_i)$); this difference is bigger when the total length of $\cup_{k_0 \in \pi_K^{-1}(\pi_K(Y_i))}I_{k_0}$ is shorter, satisfying our requirement. It can be derived that the entropy of $\mathrm{Uniform}([0,1])$ is $0$, and the entropy of $\mathrm{Uniform}(\cup_{k_0\in \pi_K^{-1}(\pi_K(Y_i)) } I_{k_0})$ is $\log p_{\pi_K}^{\mathcal{D}_v}(\pi_K(Y_i))$. Hence, the correct prediction of $\pi_K(Y_i)$ has the weight $-\log p_{\pi_K}^{\mathcal{D}_v}(\pi_K(Y_i))$ as in (\ref{eq:s-ITCA2}).
We summarize the the attributed weights of ACC and ITCA in Tabel~\ref{table:attributed weights}.
\begin{table}[htp]
    \centering
    \caption{The attributed weights in $\text{ACC}(\pi_K; \mathcal{D}_t, \mathcal{D}_v, \mathcal{C})$ and $\text{ITCA}(\pi_K; \mathcal{D}_t, \mathcal{D}_v, \mathcal{C})$}
    \label{table:attributed weights}
    \begin{tabular}{ccc}
    \toprule
            & Weight attributed to data point $Y_i$ & Weight attributed to $\pi_K$'s  $k$-th class  \\
    \midrule
         ACC  &  1 & $p_{\pi_K}^{\mathcal{D}_v}(k)$\\
         ITCA &  $-\log p_{\pi_K}^{\mathcal{D}_v}(\pi_K(Y_i))$&$-p_{\pi_K}^{\mathcal{D}_v}(k)\cdot \log p_{\pi_K}^{\mathcal{D}_v}(\pi_K(k))$\\  
    \bottomrule
    \end{tabular}
\end{table}
\par
To address the issue that ITCA is defined based on one random split on $\mathcal{D}$ (each data point is used only once for either training or validation) in (\ref{eq:s-ITCA}) and (\ref{eq:s-ITCA2}), we further define the \textit{$R$-fold cross-validated (CV) ITCA} as 
\begin{equation}\label{eq:ITCA_CV}
	\text{ITCA}^{\text{CV}}(\pi_K; \mathcal{D}, \mathcal{C}) := \frac{1}{R} \sum_{r=1}^R\text{ITCA}(\pi_K; \mathcal{D}_t^r, \mathcal{D}_v^r, \mathcal{C})\,,
\end{equation}
where the dataset $\mathcal{D}$ is randomly split into $R$ equal-sized folds, with the $r$-th fold $\mathcal{D}_v^r$ serving as the validation data and the union of the remaining $R-1$ folds $\mathcal{D}_t^r$ serving as the training data. In the following text, we will refer to $\text{ITCA}^{\text{CV}}$ as the ITCA criterion if without specification.
\par
By definition, ITCA is non-negative and becomes equal to zero when $K=1$, i.e., the degenerate case when all classes are combined as one and classification becomes meaningless. This gives ITCA a nice property: unless all predictions are wrong for all $K \ge 2$---an unrealistic scenario, the class combination that maximizes ITCA would not be the degenerate $\pi_1$.
\par
An advantage of ITCA is that it is adaptive to all machine-learning classification algorithms and its values are comparable for different algorithms.
Hence, ITCA allows users to choose the most suitable algorithm for a specific classification task.
In a task where prediction accuracy is of primary interest, users may compare algorithms by their optimal ITCA values (whose corresponding class combinations may differ for different algorithms) and choose the algorithm (along with the class combination) that gives the largest optimal ITCA value.
Granted, if a classification algorithm has a sufficiently high accuracy for predicting the original $K_0$ classes, ITCA would not suggest any classes to be combined.
Hence, in exploratory data analysis where the goal is to find similar classes, users may use ITCA with a weak classification algorithm (e.g., LDA) so that the class combination found by ITCA can reveal similar classes.
\subsection{Search strategies}\label{sec:search}
\par
Given the dataset $\mathcal{D}=\{(\bX_i, Y_i) \}_{i=1}^n$, we aim at finding the optimal class combination that maximizes the ITCA. 
A na\"ive strategy is the \textit{exhaustive search}, i.e., computing the ITCA of all allowed class combinations $\pi_K$'s for $2 \le K \le K_0$, whose set is denoted by $\mathcal{A}$; however, $|\mathcal{A}|$ can be huge even when $K_0$, the number of original classes, is moderate (Table \ref{table:mapping_numbers}). 
Specifically, if the class labels are nominal, the number of allowed combinations, $|\mathcal{A}|$, is known as the Bell number minus one \citep{Bell1938}.  
If the class labels are ordinal, only adjacent classes should be combined; then, the number of allowed combinations is $2^{K_0-1} - 1$ \citep{Feller2008}. Since a classification algorithm $\mathcal{C}$ needs to be trained for every $\pi_K$ to calculate the ITCA of $\pi_K$, a large $|\mathcal{A}|$ would make the exhaustive search strategy computationally infeasible.
\begin{table}[hpbt]
	\centering 
	\caption{The number of allowed class combinations given $K_0$}
	\label{table:mapping_numbers}
	\begin{tabular}{ccrrrrrr}
		\toprule
		& \multirow{2}{*}{Label type} & \multicolumn{6}{c}{$K_0$}  \\ \cline{3-8}
		& & 2 & 4 & 6 & 8 & 12&16\\
		\midrule
		\multirow{2}{*}{$|\mathcal{A}|$} & Nominal & 1&14 &202 &4{,}139&4{,}213{,}596&\textappr $10^{10}$  \\
		& Ordinal &  1 & 7&31 & 127& 2{,}047&  32{,}767\\
		\bottomrule
	\end{tabular}
\end{table} 
\par
This combinatorial optimization problem resembles the \textit{multiway partition problem}.
A typical multiway partition problem consists of a finite ground set $V$ and a nonnegative \textit{submodular} set function $f:2^V\to\R_+$, which maps a subset of $V$ to a nonnegative real value; $f$ is \textit{submodular} if and only if $f(A) + f(B) \geq f(A\cup B) + f(A\cap B)$, $\forall A, B \subset V$. 
The multiway partition problem aims to partition $V$ into $K$ disjoint sets $A_1, \dots, A_K$ (with $\cup_{k=1}^K A_k = V$) to minimize $\sum_{k=1}^{K} f(A_k)$.
As a well-studied problem, 
the multiway partition is known as NP-hard when $K$ is not fixed \citep{Queyranne1999}.
There are two types of approximation algorithms for solving the multiway partition problem.
The first type includes local search strategies such as the greedy search strategy \citep{Zhao2005, Lee2010}. 
The second type utilizes the submodularity of $f$ to relax the combinatorial optimization problem as a continuous optimization problem \citep{Chekuri2011, Feldman2017}.\\\\
In our setting, a class combination $\pi_K$ induces a partition of the $K_0$ observed classes: $A_1:=\pi_K^{-1}(1), \ldots, A_K:=\pi_K^{-1}(K)$ such that $\cup_{k=1}^K A_k = V := [K_0]$. Note that $A_k$ indicates the composition of the $k$-th combined class. Hence, maximizing the ITCA defined in (3) can be re-expressed in the form of a multiway partition problem if we define 
\[ f(A_k) = - \text{ resolution}(A_k) \cdot \text{accuracy}(A_k) + c\,, 
\]
where $\text{resolution}(A_k) := - p^{\mathcal{D}_v}(A_k) \cdot \log p^{\mathcal{D}_v}(A_k)$, with $p^{\mathcal{D}_v}(A_k)$ indicating the proportion of $A_k$ (the $k$-th combined class) in the validation dataset $\mathcal{D}_v$; $\text{accuracy}(A_k)$ means the prediction accuracy of the trained classifier $\phi_{\pi_K}^{\mathcal C, \mathcal{D}_t}$ for $A_k$ on $\mathcal{D}_v$ (i.e., prediction accuracy conditional on $A_k$); the constant  
$c:=-\min_{k_0\in[K_0]}\log p^{\mathcal{D}_v}(\{k_0\})$, where $p^{\mathcal{D}_v}(\{k_0\})$ is the proportion of the $k_0$-th observed class in $\mathcal{D}_v$, ensures that $f$ is nonnegative. Hence, maximizing $\text{ITCA}(\pi_K; \mathcal{D}_t, \mathcal{D}_v, \mathcal{C})$ in (3) is equivalent to minimizing $\sum_{k=1}^{K} f(A_k)$. In the ITCA maximization problem, since $K$ is not fixed, the problem is NP-hard; also, since $f$ is not submodular, the relaxation strategies used in the second type of approximation algorithms do not apply. Hence, from the first type of approximation algorithms, we adopt two heuristic local search strategies---the \textit{greedy search} and \textit{breadth-first search (BFS)} strategies---to maximize ITCA.
\par 
For greedy search, we start from $\pi_{K_0}^*=\pi_{K_0}$, which does not combine any classes.
In the $k$-th round ($1\le k \le K_0-2$), we find the best combination that maximizes the ITCA among the allowed combinations $\pi_{K-k}$'s, which are defined based on the chosen $\pi_{K-k+1}^*$. Next, we start from the chosen $\pi_{K-k}^*$ and repeat this procedure until ITCA cannot be improved or $K_0-2$ rounds are finished.
This greedy search strategy reduces the search space significantly and is summarized in Algorithm \ref{algo:greedy}.
\par 
Here we analyze the greedy search's computational cost for computing the $R$-fold CV ITCA.
	Given a size-$n$ dataset $\mathcal{D}$ and a classification algorithm $\mathcal{C}$, suppose that the computational cost is $\theta(n, \mathcal C)$ for training a classifier on every $R-1$ folds of training data and evaluating the classifier on the remaining fold of training data. Hence, the computational cost for evaluating each class combination is $R\theta(n, \mathcal C)$.
	For ordinal class labels, in the $1$-st round, the greedy search begins with $K_0$ classes and has $K_0 - 1$ feasible combinations; 
	hence, the computational cost is $(K_0 - 1)R\theta(n, \mathcal C)$.
	In the $i$-th round, there are $K_0 - i + 1$ classes and $K_0 - i$ feasible combinations; hence, the computational cost is $(K_0 - i)R\theta(n, \mathcal C)$.
	In the worst case, the greedy search terminates after $K_0 - 2$ rounds, leaving $2$ classes.
	Hence, the greedy search's worst-case complexity is $\sum_{i=1}^{K_0-2} (K_0 - i) R\theta(n, \mathcal C) = (K_0 - 2)(K_0 + 1)R\theta(n, \mathcal C)/2$.
	For nominal class labels, there are $\binom{K_0}{2}$ possible combinations in the $1$-st round, $\binom{K_0-1}{2}$ possible combinations in the $2$-nd round, and $\binom{3}{2}$ possible combinations in the $(K_0-2)$-th round in the worst case.
	Hence, the greedy search's worst-case complexity is $\sum_{K = 3}^{K_0} \binom{K}{2} R\theta(n, \mathcal C)=\left[\binom{K_0+1}{3} - 1\right]  R\theta(n, \mathcal C)$.
	We summarize the greedy search's worst-case complexities for ordinal and nominal class labels in Table~\ref{table:complexity}.
\begin{algorithm}[hbpt]
\caption{Greedy search algorithm}\label{algo:greedy}
	\begin{algorithmic}[1]
		\State Input data $\mathcal{D} = \{(\bX_i, Y_i) \}_{i=1}^n$ and a classification algorithm $\mathcal{C}$. 
		\State Set $K \leftarrow K_0$ and $\pi_K^* \leftarrow \pi_{K_0}$. 
		\State Compute $\text{ITCA}^{\text{CV}}(\pi_{K}^*; \mathcal{D}, \mathcal{C})$.
		\While {$K > 2$}
		\State Determine the set of allowed class combinations $\mathcal{A}_{K-1}$ based on $\pi_K^*$.
		\For {each allowed class combination $\pi_{K-1} \in \mathcal{A}_{K-1}$}
		\State Compute $\text{ITCA}^{\text{CV}}(\pi_{K-1}; \mathcal{D}, \mathcal{C})$. 
		\EndFor   
		\If  {there exists no $\pi_{K-1}$ that achieves $\text{ITCA}^{\text{CV}}(\pi_{K-1}; \mathcal{D}, \mathcal{C}) > \text{ITCA}^{\text{CV}}(\pi_{K}^*; \mathcal{D}, \mathcal{C})$}
		\State Break.
		\Else
		\State $K \leftarrow K-1$; 
		\State $\pi_K^* \leftarrow  \argmax\limits_{\pi_{K-1} \in \mathcal{A}_{K-1}}\text{ITCA}^{\text{CV}}(\pi_{K-1}; \mathcal{D}, \mathcal{C})$.
		\EndIf
		\EndWhile
		\State Return $\pi^*_{K}$.
	\end{algorithmic}
\end{algorithm}
\par 
It is well known that the greedy search strategy may not lead to the globally optimal class combination.
Besides the greedy search, another commonly used search strategy is the \textit{breadth-first search (BFS)} summarized in Algorithm \ref{alg:BFS}, which uses a queue to store class combinations that may be further combined.
Specifically, given $\pi_K$ (with $K \ge 3$) removed from the front of queue, we use $\mathcal{N}(\pi_K)$ to denote the set of allowed combinations $\pi_{K-1}$'s that combine any two class defined by $\pi_K$. For $\pi_{K-1} \in \mathcal{N}(\pi_K)$ that has not been visited and improves the ITCA of $\pi_K$, the BFS strategy adds $\pi_{K-1}$ to the end of the queue and considers $\pi_{K-1}$ as a candidate optimal class combination. The BFS strategy stops when the queue is empty.
\begin{table}[hpbt]\caption{Worst-case complexity of greedy search and BFS}\label{table:complexity}
	\centering
	\begin{threeparttable}
	\begin{tabular}{lcc}
		\toprule
		Label type& Greedy search & BFS \\
		\midrule
		Ordinal & $(K_0 - 2)(K_0 + 1)R\theta(n, \mathcal C)/2$& $\left(2^{K_0 - 1} - 1\right)R\theta(n, \mathcal C)$\\
		Nominal & $\left[\binom{K_0 + 1}{3} - 1\right]R\theta(n, \mathcal C)$ & $\left(B_{K_0} - 1\right)R\theta(n, \mathcal C)$* \\
		\bottomrule
	\end{tabular}
    \begin{tablenotes}
      \small
    \item *$B_{K_0}$ is the Bell number of $K_0$.
    \end{tablenotes}
    \end{threeparttable}
\end{table}
\par
The BFS has a search space much larger than the greedy search's but usually smaller than the exhaustive search's.
The BFS may enumerate all allowed class combinations in the worst case and become the exhaustive search.
Hence, the BFS's worst-case complexity is $\left(2^{K_0 - 1} - 1\right)R\theta(n, \mathcal C)$ for ordinal labels and $\left(B_{K_0} - 1\right)R\theta(n, \mathcal C)$ for nominal labels, where $B_{K_0}$ is the Bell number of $K_0$,
i.e., $B_{K_0} :=\sum_{K=0}^{K_0}\stirlingII{K_0}{K}$, where $\stirlingII{K_0}{K}:=\frac{1}{K!}\sum_{i=0}^{K} (-1)^i \binom{K}{i}(K-i)^{K_0}$ is the Stirling number of the second kind. 
We summarize the BFS's worst-case complexities in Table~\ref{table:complexity}.
\par
Despite being heuristic strategies, the greedy search and BFS perform equally well as the exhaustive search in our simulation results in Section 3. Moreover, our theoretical analysis shows that BFS is equivalent to the exhaustive search with the oracle classification algorithm (see Appendix~\ref{subsec:search property}).
\begin{algorithm}[hpt]	
	\begin{algorithmic}[1]
		\State Input data $\mathcal{D} = \{(\bX_i, Y_i) \}_{i=1}^n$ and a classification algorithm $\mathcal{C}$.  
		\State Initialize an empty set of combinations $\mathcal{A}$ and an empty queue $\mathcal{Q}$. 
		\State Compute $\text{ITCA}^{\text{CV}}(\pi_{K_0}; \mathcal{D}, \mathcal{C})$.
		\State $\mathcal{A} \leftarrow \mathcal{A} \cup \{\pi_{K_0}\}$. \Comment{Add $\pi_{K_0}$ to $\mathcal{A}$.}
		\State $\mathcal{Q}.\text{enqueue}(\pi_{K_0})$. \Comment{Add $\pi_{K_0}$ to the back of $\mathcal{Q}$.}
		\While {$\mathcal{Q}$ is not empty}
		\State $\pi_K \leftarrow \mathcal{Q}.\text{dequeue}()$. \Comment{Remove the front element of $\mathcal{Q}$ as $\pi_K$.}
		\State Determine the set of allowed class combinations $\mathcal{N}(\pi_K)$ based on $\pi_K$.
		\For {each allowed class combination $\pi_{K-1} \in \mathcal{N}(\pi_K)$}
		\If {$\pi_{K-1}$ is not visited}
		\State Compute $\text{ITCA}^{\text{CV}}(\pi_{K-1}; \mathcal{D}, \mathcal{C})$.
		\If {$\text{ITCA}^{\text{CV}}(\pi_{K-1}; \mathcal{D}, \mathcal{C})>\text{ITCA}^{\text{CV}}(\pi_K; \mathcal{D}, \mathcal{C})$}
		\State $\mathcal{A} \leftarrow \mathcal{A} \cup \{\pi_{K-1}\}$. \Comment{Add $\pi_{K-1}$ to $\mathcal{A}$.}
		\State $\mathcal{Q}.\text{enqueue}(\pi_{K-1})$. \Comment{Add $\pi_{K-1}$ to the back of $\mathcal{Q}$.} 
		\EndIf
		\State Mark $\pi_{K-1}$ as visited.
		\EndIf	
		\EndFor
		\EndWhile
		\State Return $\pi_K^*\leftarrow \underset{\pi \in \mathcal{A}}{\arg\max}~\text{ITCA}^{\text{CV}}(\pi; \mathcal{D}, \mathcal{C})$.
	\end{algorithmic}
	\caption{Breadth-first search algorithm}
	\label{alg:BFS}
\end{algorithm}   
\subsection{Some theoretical remarks}
\par Of note, the ability of ITCA to find the true class combination depends on the classification algorithm.
Using the oracle algorithm (see Definition \ref{def:oracle} in Appendix~\ref{sec:p-ITCA}) and the LDA algorithm as examples, we analyze the properties of ITCA at a population level.
As expected, when used with the oracle algorithm, ITCA has a much stronger ability to find the true class combination than when it is used with the LDA algorithm (see Appendix~\ref{sec:p-ITCA}).
We also find that, when the LDA is used as a soft classification algorithm, ITCA is more likely to find the true class combination.
In a special case where data are drawn from two well-separated Gaussian distributions, we can see that the soft LDA becomes the oracle classification algorithm (see Appendix~\ref{subsec:enhance}).
\par
Our theoretical analysis also reveals that ITCA is unsuitable for combining observed classes, even if ambiguous, into a large class that dominates in proportion.
The reason is that ITCA automatically balances the trade-off between prediction accuracy and classification resolution. As a result, if the combined class dominates in proportion, ITCA may decrease since the decrease in classification resolution outweighs the increase in prediction accuracy. In other words, ITCA refrains from outputting a combined class dominant in proportion. This property ensures that ITCA would never combine all classes into one, and it is reasonable for applications in which balanced classes are desired.
\par 
In Appendix~\ref{subsec:search property}, we show that BFS is equivalent to the exhaustive search with the oracle classification algorithm.
Moreover, we show that for both greedy search and BFS, the search space of possible class combinations can be further pruned if the classification algorithm satisfies a non-stringent property (see Appendix~\ref{sec:pruning}).
Please refer to Appendix for the detailed results. 
\section{Simulation studies}\label{sec:exp}
\subsection{ITCA outperforms alternative class combination criteria on simulated data}\label{subsec:sim}
\par
To verify the effectiveness of ITCA, we first compare ITCA with five alternative criteria---\textit{accuracy} (ACC), \textit{mutual information} (MI), \textit{adjusted accuracy} (AAC), \textit{combined Kullback-Leibler divergence} (CKL), and \textit{prediction entropy} (PE)---in simulations.
Among the five alternative criteria, ACC is commonly used to evaluate classification algorithms, and MI is often to evaluate clustering algorithms. 
The rest three alternative criteria, namely AAC,  CKL, and PE, are our newly proposed criteria to balance the trade-off between classification accuracy and classification  resolution from three other perspectives.
Please refer to Appendix \ref{sec:alternative_criteria} and Supplementary Material Section~2.4 for the definitions of the five alternative criteria.
\par
Given the number of observed classes $K_0$ and the true class combination $\pi^*_{K^*}: [K_0] \rightarrow [K^*]$, we first generate $K^*$ class centers by a random walk in $\R^d$.
The random walk starts from the center of class $1$: $\boldsymbol{\mu}_1 =  \boldsymbol{0} \in \R^d$. 
At time $k = 2, \ldots, K^*$, the walk chooses a direction at random and takes a step with a fixed length $l$; that is, the center of class $k$ is 
$
\boldsymbol{\mu}_{k} = \boldsymbol{\mu}_{k-1} + l\boldsymbol{v},
$  
where $\boldsymbol{v} \in \R^d$ is a random direction vector with the unit length, i.e., $||\boldsymbol{v}||=1$.
To make the true classes distinguishable, we ensure that the minimal pairwise Euclidean distance among $\boldsymbol{\mu}_{1}, \ldots, \boldsymbol{\mu}_{K^*}$ is greater than $\sigma$, i.e., the standard deviation of every feature in each class.
Given $\{\boldsymbol{\mu}_{k}\}_{k=1}^{K^*}$ and $\sigma$, we then define the distribution of $(\bX, Y)$ as follows.
\begin{enumerate}
	\item The true class $Y^* \sim \dunif([K^*])$, i.e., $Y^*$ randomly picks a value in $[K^*]$ with probability $1/K^*$.
	\item The observed class $Y \sim \dunif(\pi_{K^*}^{*-1}(Y^*))$, i.e., $Y$ randomly picks a value in $\pi_{K^*}^{*-1}(Y^*) = \{k_0 \in [K_0]: \pi_{K^*}^{*}(k_0) = Y^*\} \subset [K_0]$---the observed classes that belong to the true class $Y^*$---with probability $1/|\pi_{K^*}^{*-1}(Y^*)| = 1/(\sum_{k_0=1}^{K_0}\1\left(\pi_{K^*}^{*}(k_0) = Y^*)\right)$.
	\item The observed feature vector $\bX \sim \mathcal{N}(\boldsymbol{\mu}_{Y^*}, \sigma^2 \mathbf{I}_d)$, i.e., $\bX$ follows a $d$-dimensional Gaussian distribution with mean $\boldsymbol{\mu}_{Y^*}$ specified by the true class $Y^*$.
\end{enumerate}
Given a dataset $\{(\bX_i, Y_i)\}_{i=1}^n$, which contains independently and identically distributed (i.i.d.) observations from the above distribution, an ideal criterion of class combination is expected to be maximized at $\pi_{K^*}^*$.

\par 
We  generate a simulated dataset with $K_0=6$, $K^*=3$, $l=3$, $\sigma=1.5$, $n=2000$, and $d=5$. We assume the true class combination is $\pi^*_3=\{(1, 2), (3, 4), (5, 6)\}$, i.e., the $1$-st and $2$-nd observed classes belong to one true class, and so do the $3$-rd and $4$-th observed classes, as well as the $5$-th and $6$-th observed classes. We also assume that the classes are ordinal; that is, $\mathcal A$ contains $2^5 - 1=31$ class combinations (Table~\ref{table:mapping_numbers}).
Because of the moderate size of $|\mathcal{A}|$, we use the exhaustive search to enumerate all allowed class combinations.
\par
\begin{figure}[hbt]
	\centering
	\includegraphics[width=\linewidth]{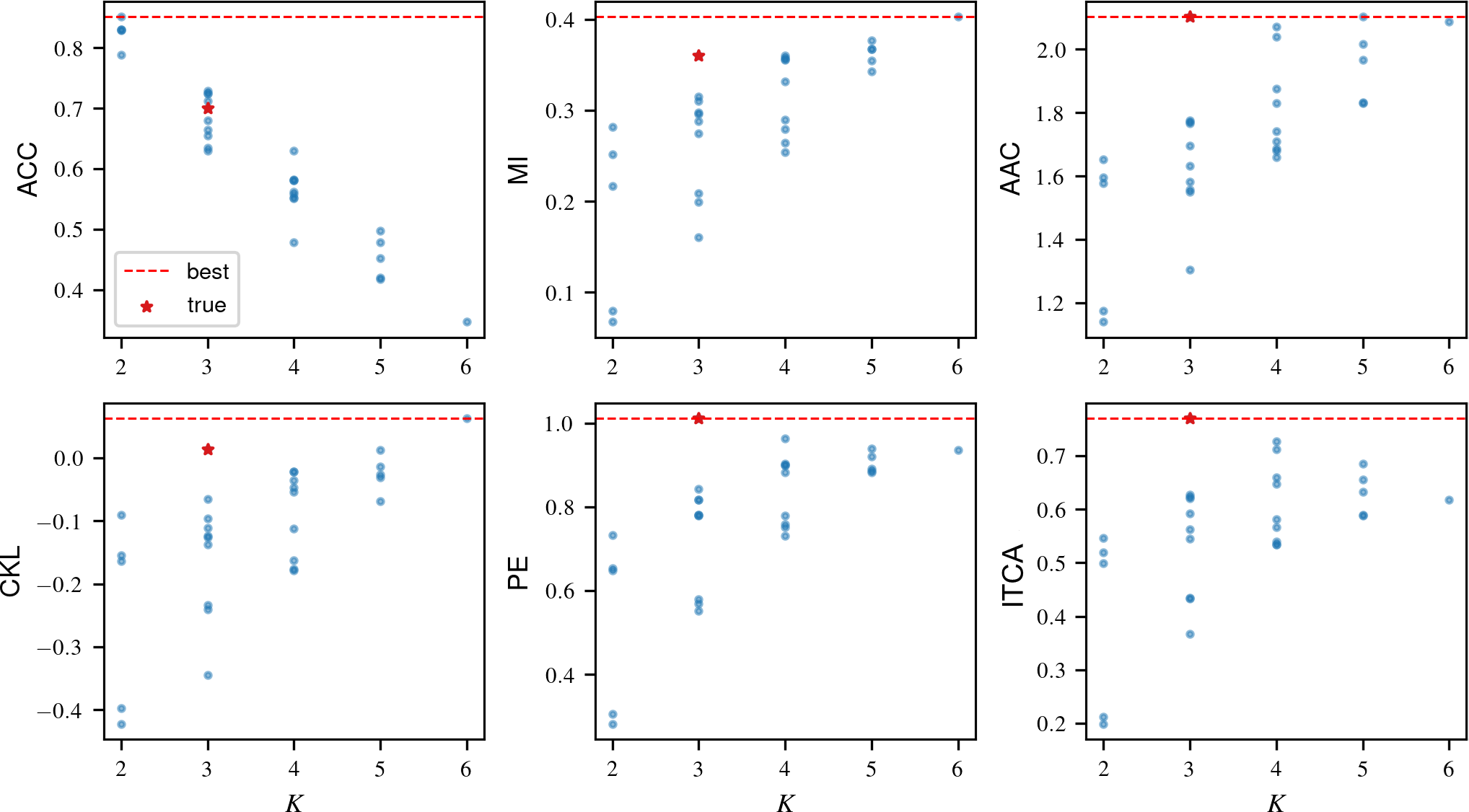}
	\caption{Comparison of ITCA and five alternative criteria using the LDA as the classification algorithm. The dataset is generated with $K_0=6$, $K^*=3$, $l=3$, $\sigma=1.5$, $n=2000$, and $d=5$. The true class combination is $\pi^*_3=\{(1, 2), (3, 4), (5, 6)\}$. For each criterion, the $31$ blue points correspond to the $31$ class combinations $\pi_K$'s with $K=2,\ldots,6$. The true class combination $\pi_{K^*}^*$ is marked with the red star, and the best value for each criterion is indicated by a horizontal dashed line. The true class combination is only found by PE and ITCA without close ties.}
	\label{fig:sim_LDA}
\end{figure}

Regarding the classification algorithm, we consider the LDA and random forest (RF) \citep{Breiman2001}. Figure~\ref{fig:sim_LDA} shows that ITCA using LDA successfully finds the true class combination $\pi^*_3$; that is, when evaluated on the $31$ $\pi_K$'s, ITCA is maximized at $\pi^*_3$.
In contrast, ACC is maximized when the $K_0=6$ observed classes are combined into $K=2$ classes; hence, it is not appropriate for guiding class combination.
Although better than ACC, MI and two alternative criteria we proposed (AAC and CKL) still fail to find $\pi^*_3$. Among the alternative criteria, only PE correctly identifies $\pi_3^*$ because its definition is similar to that of ITCA.  
The results using RF (Supplementary Material, Figure~S1) are consistent with Figure~\ref{fig:sim_LDA}.
We design another simulation with $K^*=5$ and the true class combination $\pi_5^* = \{(1,2),3,4,5,6\}$. The results in Figures~S2 and S3 (Supplementary Material) show that ITCA outperforms all five alternative criteria, including PE, by finding $\pi_5^*$ with the largest gap from other class combinations.
\par
To further evaluate the performance of ITCA and the five alternative criteria, we escalate this simulation design by setting $K_0 = 8$; then the number of allowed class combinations is $2^7 - 1 = 127$ (Table~\ref{table:mapping_numbers}).
We repeat the simulation for $127$ times, each time using one of the 127 allowed class combinations as the true class combination $\pi_{K^*}^*$; the other simulation parameters are kept the same. 
We use the exhaustive search with LDA or RF to find the best class combination guided by each criterion, denoted by $\pi_K^m$ for criterion $m \in \{\text{ITCA}, \text{ACC}, \text{MI}, \text{AAC}, \text{CKL}, \text{PE}\}$. To evaluate the performance of criterion $m$, we define the distance between $\pi_K^m$ and $\pi_{K^*}^*$ as follows. First, we encode each allowed class combination $\pi_K: [K_0] \rightarrow [K]$ for combining $K_0$ ordinal classes as a $(K_0-1)$-dimensional binary vector, as in the ``stars and bars'' used in combinatorics \citep{Feller2008}.
For example, the class combination $\{(1, 2), 3, 4, 5, 6, 7, 8\}$ can be represented by $1 2|3|4|5|6|7|8$ and thus encoded as the binary vector $(0,1,1,1,1,1,1)$, where the $0$ indicates that there is no bar between the original classes 1 and 2, the first $1$ indicates that there is a bar between the original classes $2$ and $3$, etc. Second, we define the distance between $\pi_K^m$ and $\pi_{K^*}^*$ as the Hamming distance between their binary encodings; hence, the distance takes an integer value ranging from $0$ to $K_0-1$, with $0$ indicating that $\pi_K^m = \pi_{K^*}^*$, i.e., the criterion $m$ finds the true class combination.
\par
\begin{table}[hpbt] 
	\caption{The performance of six criteria on the $127$ simulated datasets with $K_0=8$. The best result in each column is boldfaced.}\label{table:table-sim-8}
	\begin{tabular}{lrccrcc}
		\toprule
		\multirow{3}{*}{Criterion} & \multicolumn{1}{c}{\# successes} & Average & Max & \multicolumn{1}{c}{\# successes} & Average & Max\\
		\cmidrule(r){2-2} \cmidrule(r){5-5}
		& \multicolumn{1}{c}{\# datasets} & Hamming & Hamming & \multicolumn{1}{c}{\# datasets} & Hamming & Hamming\\
		\cmidrule(r){2-4} \cmidrule(r){5-7}
		& \multicolumn{3}{c}{LDA} &  \multicolumn{3}{c}{RF} \\
		\midrule
		ACC&6/127&2.54&6&7/127&2.53&6\\
		MI&7/127&2.51&6&11/127&2.33&6\\
		AAC&15/127&2.02&6&15/127&1.98&6\\
		CKL&3/127&3.68&6&5/127&2.87&5\\
		PE&101/127&0.47&4&94/127&0.46&3\\
		ITCA&\textbf{120/127}&\textbf{0.12}&\textbf{3}&\textbf{120/127}&\textbf{0.08}&\textbf{2}\\
		\bottomrule
	\end{tabular}
\end{table} 
We evaluate the performance of each criterion using three criteria: (1) the number of datasets (the larger the better) on which the criterion identifies the true class combination; (2) the average and (3) the maximum Hamming distances (the smaller the better) between the criterion's best class combination and the true class combination across the $127$ datasets. Table \ref{table:table-sim-8} shows that, among the six criteria, 
ITCA has the best performance under all three criteria;
PE has the second best performance after ITCA;
the other criteria fail to find the true class combination on at least $80\%$ of the datasets.
ITCA only misses the true class combination when $K^*=2$, which corresponds to $\binom{7}{6}=7$ true class combinations (see Supplementary Material Table~S5).
This result is confirmed in a similar analysis with $K_0=6$ (in Supplementary Material Tables~S1 and S4), and it can be explained by the theoretic analysis in Appendix~\ref{sec:p-ITCA}, which shows that ITCA would not combine two same-distributed classes when the combined class' proportion is too large (e.g., larger than $0.5$).    

\par
\begin{table}[hbpt]
	\caption{Performance of ITCA using five search strategies and LDA on the $127$ simulated datasets with $K_0=8$.}
	\label{table:search-8}
	\begin{tabular}{lcccr}
		\toprule
		\multirow{2}{*}{Strategy} &  \# successes & Average & Max & Average \# class \\
		\cmidrule(r){2-2}
		& \# datasets & Hamming & Hamming & combinations examined\\
		\midrule
		Exhaustive  &120/127 & 0.13 &3&127.00 \\
		Greedy search&120/127&0.12 &3 &22.52 \\
		BFS & 120/127&0.10 &2 &53.61  \\
		Greedy (pruned)& 120/127&0.09 &2 &11.91\\
		BFS (pruned) &120/127&0.09 &3 &27.20  \\
		\bottomrule		
	\end{tabular}
\end{table}
The above results verify the effectiveness of ITCA in finding the true class combination. 
In the following, we compare the two proposed search strategies, the greedy search and BFS, with the exhaustive search.
Specifically, we use the aforementioned $127$ simulated datasets corresponding to $K_0=8$, and we apply ITCA under the three search strategies, with LDA as the classification algorithm.
The top three rows of Table \ref{table:search-8} show that the greedy search and BFS are as effective as the exhaustive search in finding the true class combinations. Supplementary Material Table~S2 shows similar results for $K_0=6$. Compared with the exhaustive search, the greedy search and BFS examine fewer class combinations and thus greatly reduce the computational time because each class combination, if examined, needs a separate classifier training.

We note that the search space of the greedy search and BFS can be further pruned if the classification algorithm satisfies a non-stringent property, and we will discuss this pruning procedure in Appendix~\ref{sec:pruning}.
As a preview, the bottom two rows of Table \ref{table:search-8} show that pruning reduces the search spaces of the greedy search and BFS while maintaining the performance.

When $K_0$ is large, it is unrealistic to use the exhaustive search. Here we use $K_0=20$ ordinal classes as an example.
Out of the $2^{19} - 1\approx 5.24 \times 10^5$ allowed class combinations (Table~\ref{table:mapping_numbers}), we randomly select $50$ class combinations as the true class combination $\pi_{K^*}^*$, whose $K^*$ ranges from 7 to 16. From each $\pi_{K^*}^*$, we generate a dataset with $n=10{,}000$ data points (the other parameters are the same as in the aforementioned simulations).
The results show that the greedy search works as well as the BFS (Supplementary Material Table~S3): both successfully find the $\pi_{K^*}^*$ of each dataset. 
On average, the greedy search only needs to evaluate ITCA on $150.08$ class combinations (87.70 combinations with the pruned search space)  out of the $\sim 5.24 \times 10^5$ allowed class combinations. In contrast, the BFS has a much larger search space ($\sim 10^4$ class combinations).
Notably, ITCA has a higher probability of success when $K^*$ is larger.
We discuss this phenomenon in Appendix~\ref{sec:p-ITCA}.
\subsection{ITCA outperforms alternative class combination criteria on the Iris data}
\par
We also compare ITCA with the five alternative class combination criteria on the famous Iris dataset\footnote{http://archive.ics.uci.edu/ml/datasets/Iris/},
which contains $K^* = 3$ classes (corresponding to three types of irises: \textit{setosa}, \textit{versicolor}, and \textit{virginica}) with $50$ data points in each class. 
The \textit{setosa} class is linearly separable from the \textit{versicolor} and \textit{virginica} classes, while \textit{versicolor} and \textit{virginica} are not linearly separable from each other.
To prepare the dataset for class combination, we randomly split the \textit{setosa} class into two equal-sized classes, making the number of observed classes $K_0=4$. Since the four classes are nominal, there are 14 allowed class combinations (Table \ref{table:mapping_numbers}).
\begin{figure}[H]
	\centering
	\includegraphics[width=0.95\linewidth]{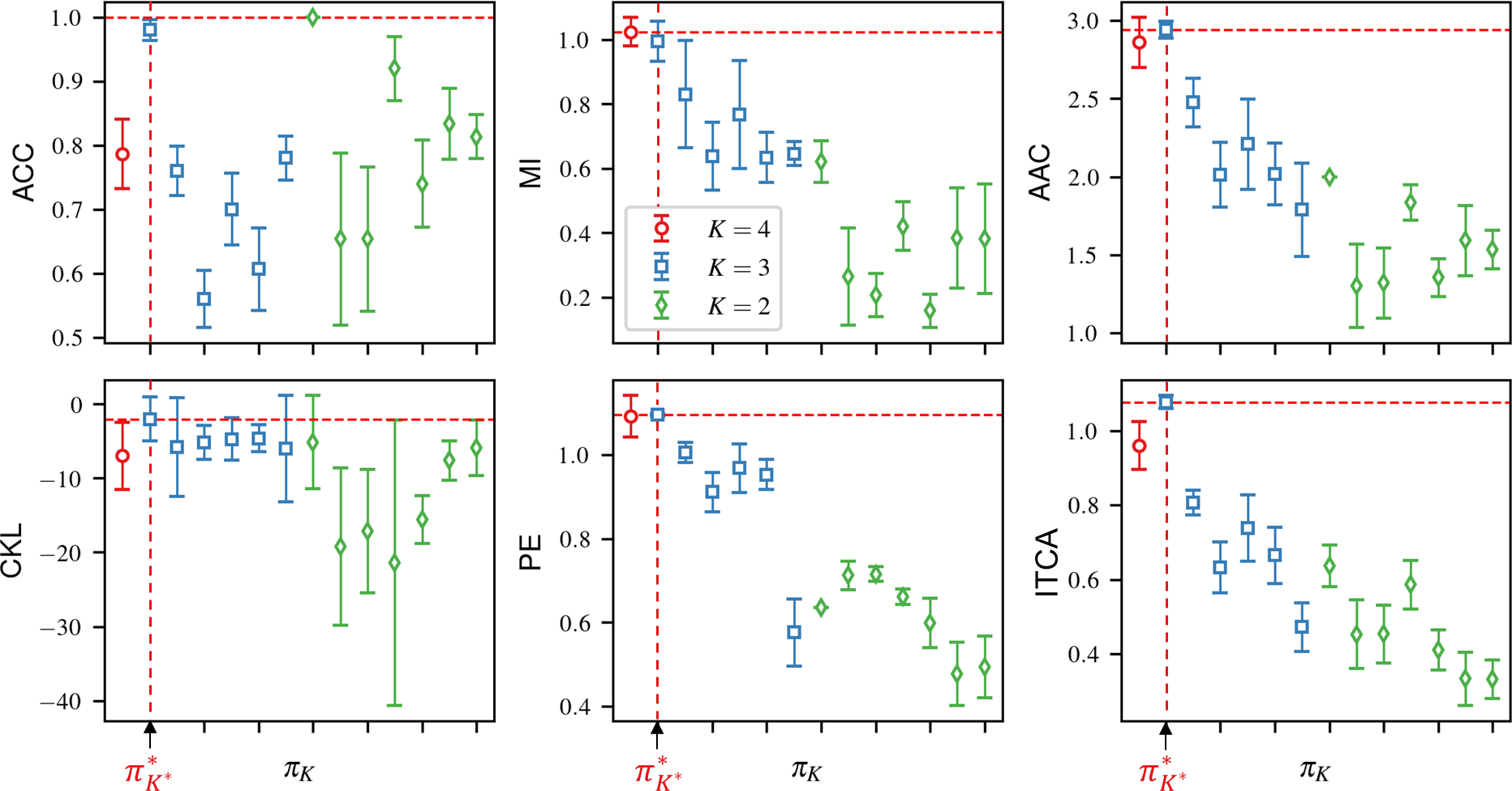}
	\caption{Comparison of ITCA and five alternative criteria using LDA as the classification algorithm on the Iris data. 
		$K_0=4$ and $K^*=3$ (with the true class combination $\pi_{K^*}^*$ marked by the arrow and the red vertical dashed line in every panel). For every allowed class combination $\pi_K$, each criterion has its value (calculated by $5$-fold CV) marked by a red circle for $K=4$, a blue square for $K=3$, and a green diamond for $K=2$); each error bar has half its length corresponding to the standard error of the criterion value (i.e., the standard deviation of the $5$ criterion values in the 5-fold CV, divided by $\sqrt{5}$). The horizontal line marks each criterion's best value. Among the six criteria, only AAC, CKL, PE, and ITCA are maximized at $\pi_{K^*}^*$, and only ITCA has a clear gap between $\pi_{K^*}^*$ and all other class combinations.}
	\label{fig:iris}
\end{figure}
\par For each allowed class combination, we compute the six class combination criteria with LDA as the classification algorithm (Figure \ref{fig:iris}).
Among the six criteria, AAC, CKL, PE, and ITCA successfully find the true class combination $\pi^*_{K^*}$. However, only ITCA leads to a clear gap between $\pi^*_{K^*}$ and the other $13$ allowed class combinations. Note that CKL has large error bars because its computation involves the inverses and determinants of sample covariance matrices, whose accurate estimation requires a large sample size.
Particularly, ACC has an undesirable result: its maximal value $1$ is obtained at the class combination $\pi_2$ where the \textit{versicolor} and \textit{virginica} classes are combined. Again, these results confirm the unsuitability of ACC for guiding class combination, and they demonstrate the advantage ITCA has over the five alternative criteria. 

\subsection{ITCA outperforms clustering-based class combination}\label{sec:clustering}
\par While ITCA provides a powerful data-driven approach for combining ambiguous classes, one may intuitively consider using a clustering algorithm to achieve the same goal. 
We consider three clustering-based class combination approaches, which are summarized below, and we compare them with ITCA on simulated data under four settings.
\begin{figure}[H]
	\centering
	\includegraphics[width=\linewidth]{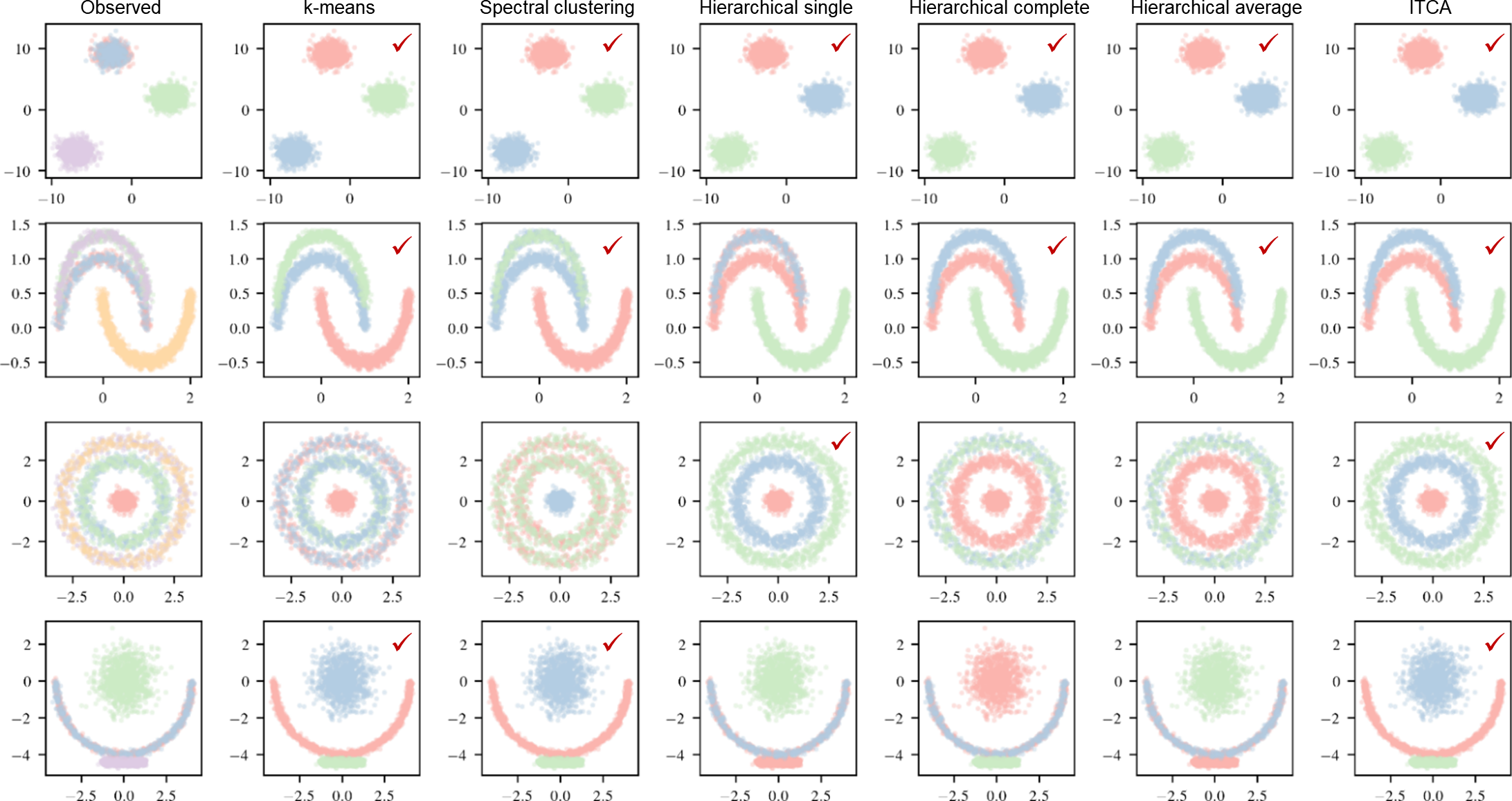}
	\caption{Comparison of clustering-based class combination approaches and ITCA (using Gaussian kernel SVM as the classification algorithm). Each row corresponds to one simulated dataset. From top to bottom, the number of observed classes is $K_0 = 4$, $5$, $5$, and $4$, and the number of true classes is $K^* = 3$, $3$, $3$, and $3$.
		In the leftmost column, colors mark the observed classes; in the other columns, the three colors indicate the three combined classes found by each combination approach. Check marks indicate the cases where the true class combinations are found. Only ITCA finds the true class combination on every dataset.
	}
	\label{fig:sim_nonlinear}
\end{figure}
\par \textbf{$K$-means-based class combination}. For the $k_0$-th class ($k_0 = 1, \ldots, K_0$), we first compute the  $k_0$-th class center as $\left(\sum_{i=1}^n \1(Y_i = k_0)\bX_i\right) \left/ \left( \sum_{i=1}^n \1(Y_i = k_0) \right) \right.$.
We then use the $K$-means clustering to cluster the $K_0$ class centers into $K^*$ clusters so that the $K_0$ observed classes are correspondingly combined into $K^*$ classes.
\par \textbf{Spectral-clustering-based class combination}. We first compute the $K^*$-dimensional spectral embeddings of $\bX_1, \ldots, \bX_n$ \citep{Ng2001} . 
Then we apply the above $K$-means-based class combination approach to the $n$ spectral embeddings to combine the $K_0$ observed classes into $K^*$ combined classes.
\par \textbf{Hierarchical-clustering-based class combination}. We first compute the $K_0$ class centers as in the $K$-means-based class combination approach. Then we use the hierarchical clustering with the single, complete, or average linkage to cluster the $K_0$ class centers into $K^*$ clusters so that the $K_0$ observed classes are correspondingly combined into $K^*$ classes.

\par Note that all these clustering-based class combination approaches require that $K^*$ (the true number of classes) is known or estimated by an external approach (e.g., an approach for determining the number of clusters \citep{Tibshirani2001,Sugar2003,Pham2005}), which is by itself a difficult problem in real-world applications. In contrast, ITCA does not require $K^*$ to be known beforehand; instead, its optimal class combination determines $K^*$ in a data-driven way. 
\par To benchmark ITCA against the clustering-based class combination approaches, we generate four datasets with two-dimensional features ($\bX_1, \ldots, \bX_n \in \R^2$), which are shown in the four rows of Figure \ref{fig:sim_nonlinear}.
Then we apply the above five clustering-based class combination approaches (three of which are hierarchical clustering with three linkages) and ITCA to the synthetic data; for ITCA, we use the Gaussian kernel support vector machine (SVM) as the classification algorithm.
\par Figure \ref{fig:sim_nonlinear} shows the results of ITCA and the clustering-based class combination approaches with $K^* = 3$ known:
only ITCA successfully finds the true class combination on every dataset. We conclude that ITCA is advantageous over the clustering-based approaches even with a known $K^*$. The major reason is that the clustering-based approaches only use the $K_0$ class centers (whose definition depends on a distance metric) and do not fully use the information in individual data points, which play a central role in the definition of ITCA.

\section{Applications}\label{sec:applications}
\subsection{Prognosis of rehabilitation outcomes of traumatic brain injury patients}\label{sec:TBI}
\par
According to the Centers for Disease Control and Prevention, traumatic brain injury (TBI) affects an estimated 1.5 million Americans every year\footnote{\url{https://www.cdc.gov/traumaticbraininjury/get_the_facts.html}}. 
The inpatient TBI rehabilitation care alone costs each patient tens of thousands of dollars per month. However, TBI rehabilitation is proven to be effective for only some but not all patients \citep{TurnerStokes2008}. 
Therefore, there is a great demand to have an automatic prognosis algorithm that can accurately predict rehabilitation outcomes for individual patients and assist patients’ decisions in seeking rehabilitation care.
We have access to the Casa Colina dataset of $n=3078$ TBI patients who received inpatient rehabilitation care. 
Patients’ disability severity was evaluated and recorded by physical therapists in the form of the Functional Independence Measure (FIM) of 17 activities at admission and discharge (Supplementary Material Table~S7). 
The FIM has seven scales ranging from 1 (patient requires total assistance to perform an activity) to 7 (patient can perform the activity with complete independence) \citep{Linacre1994}. 
In addition, patients' characteristics are recorded, including demographics (gender and age) and admission status. 
This dataset allows the development of an algorithm to predict the efficacy of rehabilitation care for individual patients. 
\par
We formulate the task of predicting patients’ rehabilitation outcomes as a multi-class classification problem, where the discharge FIM of each activity is a seven-level ($K_0=7$) ordinal outcome and the features are patients' characteristics and admission FIM. 
We are motivated to combine outcome levels because the RF algorithm, though being a best-performing algorithm in our study, has low accuracy for predicting the $K_0$ levels of many activities. 
Hence, we consult the physical therapists who graded the activities and obtain their suggested combination $\{1, (2,3,4), 5, 6, 7\}$, which has levels 2--4 combined. However, this expert-suggested class combination is subjective and not activity-specific. Intuitively, we reason that different activities may have different classification resolutions and thus different outcome level combinations. Hence, we apply ITCA as a data-driven approach to guide the combination of outcome levels for each activity.

\begin{figure}[hbpt]
	\centering
	\includegraphics[width=\linewidth]{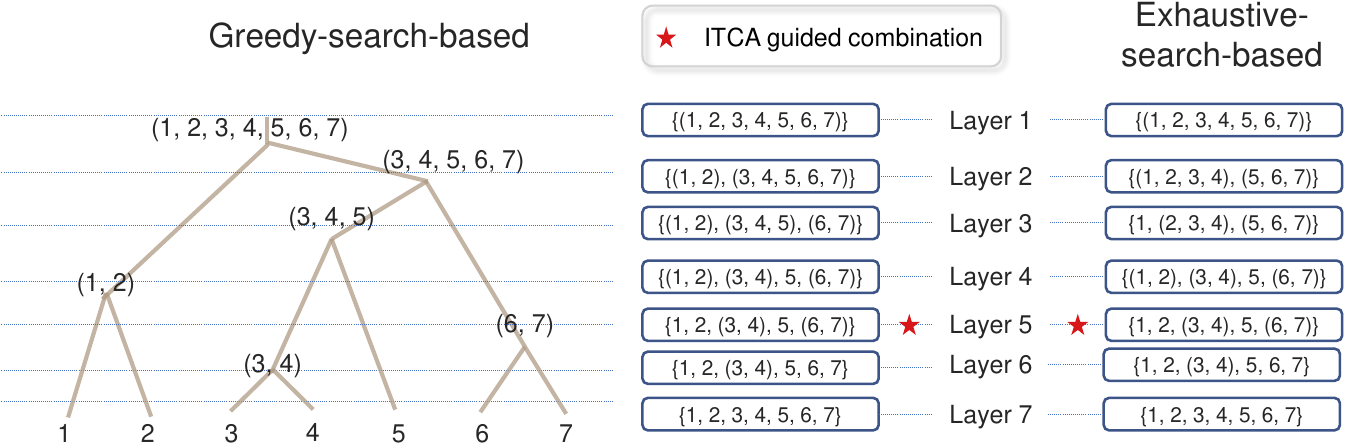} 
	\caption{Greedy-search-based and exhaustive-search-based multilayer frameworks for predicting the Toileting activity in the Casa Colina dataset. In either framework, layer $K$ has a class combination $\pi_K$ chosen by ITCA. In the greedy-search-based framework, the layers have a tree structure. Layer 5 has the same class combination in both frameworks and is found optimal by ITCA.} 
	\label{fig:casa_multilayers}
\end{figure}

Powered with ITCA, we can construct a multilayer prediction framework (Figure \ref{fig:casa_multilayers}), whose $K_0$ layers (from top to bottom) correspond to the numbers of combined classes $K=1,\ldots,K_0$, with the bottommost layer indicating no class combination.
There are two ways to construct a framework: greedy-search-based and exhaustive-search-based. 
In a greedy-search-based framework, layers are constructed by the greedy search (Section \ref{sec:search}) in a sequential way from bottom up: classes in each layer except the bottommost layer are combined by ITCA from the classes in the layer right below, so the layers would follow a nested structure. 
In contrast, in an exhaustive-search-based framework, layers are constructed separately from the bottommost layer: each layer has its optimal class combination defined by ITCA and found by the exhaustive search (Section \ref{sec:search}) given its $K$; thus, there is no nested constraint. Note that $K_0$ is often not too large in medical diagnosis and prognosis, making the exhaustive search computationally feasible.
\par
Greedy-search-based and exhaustive-search-based frameworks have complementary advantages. 
On the one hand, the former outputs a data-driven hierarchy of class combinations and is thus more interpretable if the prediction would be conducted for all layers. 
On the other hand, the latter outputs the optimal class combination for each layer and allows choosing the optimal layer (that maximizes ITCA); hence, it is more desirable if prediction would only be conducted for the optimal layer. 
For example, if healthcare providers would like to predict outcomes at a multilayer resolution, the greedy-search-based framework is more suitable. In contrast, if the priority is to combine outcome levels into the optimal classification resolution for prediction, the exhaustive-search-based framework would be a better fit. 

\par 
We use RF as the classification algorithm for its better accuracy than other popular algorithms' on the Casa Colina dataset.
Leveraging the RF algorithm, we apply the greedy-search-based and exhaustive-search-based multilayer frameworks to predict the rehabilitation outcome of each of the $17$ activities. 
For example, Figure~\ref{fig:casa_multilayers} shows the results for predicting the Toileting activity:
ITCA indicates the same optimal class combination in both frameworks: $\{1,2,(3,4),5,(6,7)\}$ in layer $K=5$, where levels 3 and 4 are combined, and so are levels 6 and 7.
\begin{figure}[H]
	\centering
	\includegraphics[width=\linewidth]{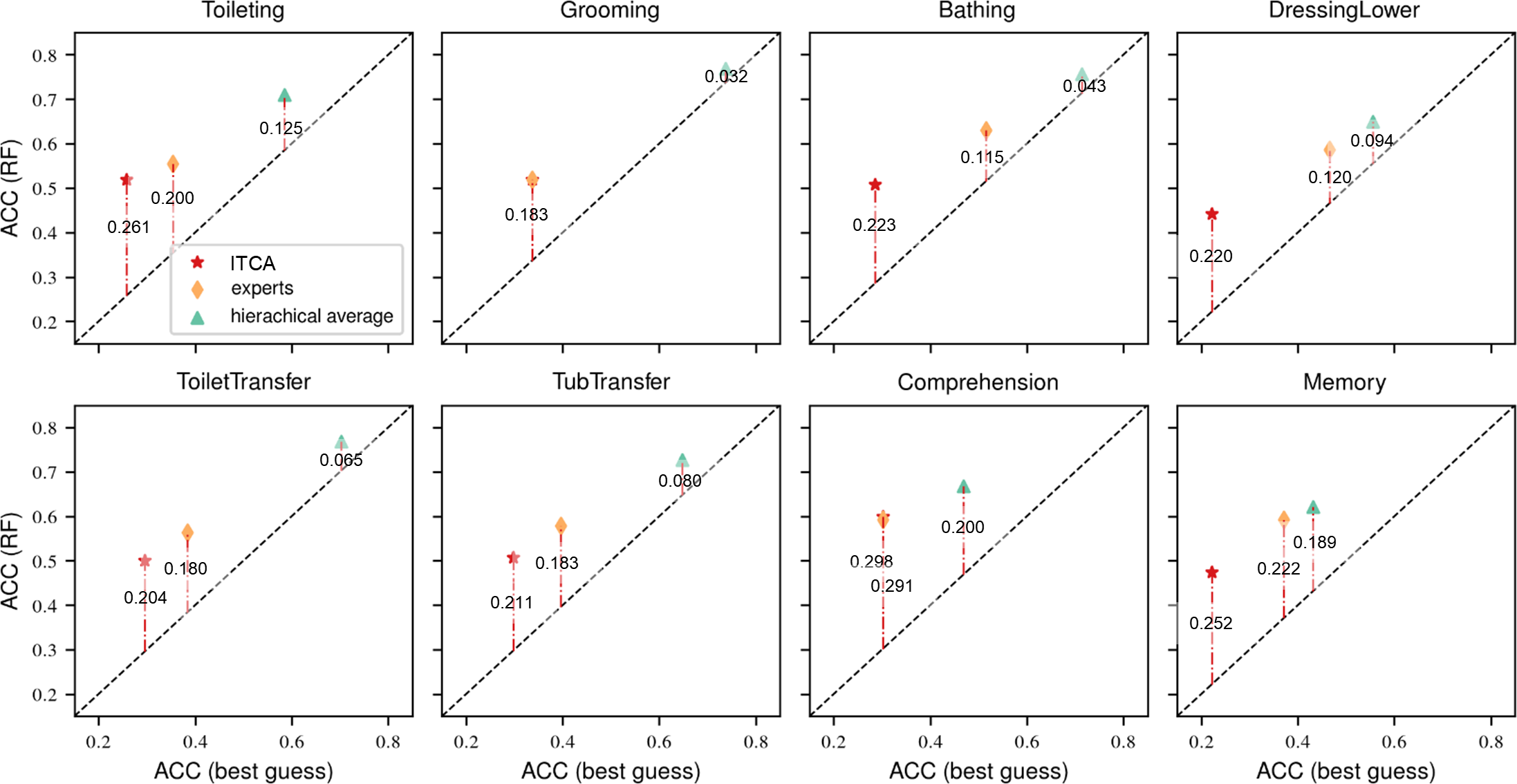}
	\caption{Prediction accuracy for the optimal level combination indicated by ITCA (using the exhaustive-search-based multilayer framework), the expert-suggested combination, and the hierarchical-clustering-based level combinations for the rehabilitation outcomes of eight activities in the Casa Colina dataset. 
		The horizontal axis shows the 5-fold CV prediction accuracy of the best guess algorithm, and the vertical axis shows the accuracy of the RF algorithm. 
		The accuracy improvement of the RF from the best guess is marked on vertical dashed lines. Note that in the ``Comprehension'' panel, ITCA finds the expert-suggested combination; the two accuracy improvement values ($0.298$ and $0.291$) should be equal in theory, but they are different due to the randomness of data splitting in the 5-fold CV.}
	\label{fig:casa_comparison}
\end{figure}
\par 
To evaluate the results of ITCA for the $17$ activities, we calculate the prediction accuracy by RF for our ITCA-guided combinations (using the exhaustive-search-based multilayer framework), the expert-suggested combination, and the hierarchical-clustering-based combinations (defined in Section~\ref{sec:clustering}; the results are based on the average linkage).
Note that the prediction accuracy for different class combinations may differ even if the best guess algorithm (i.e., the na\"ive algorithm that assigns data points to the largest class---a baseline control) is used. Hence, for a fair comparison, we evaluate each class combination by the 5-fold CV prediction accuracy improvement of the RF algorithm from the best guess algorithm.

Figure~\ref{fig:casa_comparison} shows the comparison results for eight activities, for which ITCA selects class combinations with $K$'s closest to $5$ (i.e., the number of combined levels suggested by experts). We find that, compared with the expert-suggested combination and the hierarchical-clustering-based combinations, the ITCA-guided combinations consistently lead to more balanced classes (which are more difficult to predict, as indicated by the lower prediction accuracy of the best guess algorithm) and more significant improvement in prediction accuracy. 
Figure~S5 and Tables~S8--S9 (in Supplementary Material) summarize for all 17 activities the ITCA-guided class combinations and their corresponding prediction accuracy (i.e., the ACC criterion defined in Appendix~\ref{sec:alternative_criteria}), as well as the ITCA values.

As a side note, we argue that it is inappropriate to use the prediction accuracy improvement (from the best guess) as a class combination criterion.  The reason is that the prediction accuracy of the best guess, though serving as a baseline accuracy, does not necessarily reflect the classification  resolution.
For example, an outcome encoded as two classes with equal probabilities has a lower classification resolution than another outcome encoded as three classes with probabilities 0.5, 0.25, and 0.25; however, the best guess algorithm has the same prediction accuracy $0.5$ for the two outcomes.
\subsection{Prediction of glioblastoma cancer patients' survival time}
\par
Glioblastoma cancer, also known as glioblastoma multiforme (GBM),  is the most aggressive type of cancer that begins within the brain. Hence, it is of critical importance to predict GBM patients' survival time so that appropriate treatments can be provided. In this prediction task, patients' survival time would be predicted from their clinical measurements.

We download the TCGA GBM dataset from the cBio cancer genomics portal \citep{Cerami2012}\footnote{The dataset is available at \url{https://www.cbioportal.org/}}.
The dataset contains $n=541$ patients' demographics, gene expression subtypes, therapy, and other clinical information.
We remove nine features irrelevant to survival prediction and keep $d=36$ features (see Supplementary Material Section~3.2 for the data processing details).
\par 
We formulate this survival prediction task as a classification problem instead of a regression problem to demonstrate the use of ITCA, and we compare the performance of the classifier guided by ITCA with that of the Cox regression model.
Concretely, we first discretize patients' survival time into $K_0=12$ intervals: $[0, 3)$, $[3, 6)$, $[6, 9)$, $[9, 12)$, $[12, 15)$, $[15, 18)$, $[18, 21)$, $[21, 24)$, $[24, 27)$, $[27, 30)$, $[30, 33)$, $[33, +\infty)$, where $[0,3)$ and $[33, +\infty)$ indicate that the survival time is less than 3 months and at least 33 months, respectively.

\par
\begin{figure}[H]
\centering
	\includegraphics[width=0.65\linewidth]{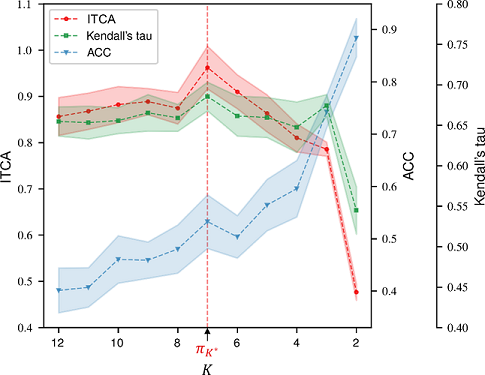}
	\caption{Results of GBM survival prediction. Three criteria, ACC, ITCA and Kendall's tau coefficient, are shown for $K$ from $K_0$ (=12) to $2$ (for each $K$, a class combination is found by ITCA, and an NN classifier is trained). Each criterion is calculated by the 5-fold CV, and its mean and standard error (i.e., standard deviation of the $5$ criterion values in the 5-fold CV, divided by $\sqrt{5}$) are shown.
	The best class combination $\pi_{K}^*$ found by ITCA is indicated by the red vertical dashed line ($K=7$), where the Kendall's tau is also maximized.
}\label{fig:survival_pred}
\end{figure}

 
We then use ITCA to optimize the survival time intervals for prediction.
For the classification algorithm, we use a three-layer neural network (NN) with the ReLU activation function and a modified cross entropy as the loss function for handling censored survival time (see Supplementary Material Section~3.2 and Figure~S6).
Starting from the observed $K_0=12$ classes, we use ITCA with greedy search to select a class combination for each $K = 11, \ldots, 2$.
For each $K$ and its selected class combination, we train a NN classifier and show the ITCA and ACC in Figure \ref{fig:survival_pred}.
As expected, as $K$ decreases, ACC increases, confirming our motivation that ACC cannot be used to guide class combination. In contrast, as $K$ decreases, ITCA first increases until $K=7$ and then decreases, confirming that ITCA balances the trade-off between prediction accuracy and classification  resolution.
\par
Following the tradition in survival analysis, we use the Kendall's tau coefficient (calculated between the predicted outcome and the observed survival time, not the discretized survival time interval, in 5-fold CV) to evaluate the prediction accuracy. Note that the Kendall's tau is a reasonable accuracy measure for survival prediction because it is an ordinal association measure that allows the predicted outcome to be either discrete (as in our classification formulation) or continuous (as in a regression formulation). Figure \ref{fig:survival_pred} shows that the optimal class combination found by ITCA leads to the best Kendall's tau, verifying that ITCA optimizes the survival time intervals for classifier construction.
\par 
We also compare the NN algorithm with two commonly used survival prediction algorithms: the Cox regression \citep{Cox1972}, a regression algorithm that predicts patients' risk scores, and the logistic regression (LR), a multi-class classification algorithm that uses the same modified cross entropy loss to predict survival time intervals as the NN algorithm does. We use the Kendall's tau to evaluate five prediction models: NN and LR classifiers for predicting the original $K_0$ survival time intervals, NN and LR classifiers for predicting their respective combined intervals guided by ITCA, and a Cox regression model for predicting risk scores (the Kendall's tau is calculated between negative predicted risk scores and observed survival time). 
Table \ref{table:survival} shows that the NN classifier trained for ITCA-guided combined intervals has the best prediction accuracy in terms of the Kendall's tau; moreover, it has the highest ITCA value among the four classifiers. This result again verifies that ITCA is a meaningful accuracy measure.  
\par
Note that the Kendall's tau is not an appropriate measure to replace ITCA because it requires the response to be a numerical or ordinal variable. Hence, the Kendall's tau cannot guide the combination of nominal class labels. 
\begin{table}[hpt]
	\centering
	\caption{Performance of survival prediction algorithms on the GBM dataset.}\label{table:survival}
	\begin{threeparttable}
		\begin{tabular}{lccc}
			\toprule
			Model &ITCA &Kendall's tau & Average p-value \\
			\midrule
			NN ($K_0$ survival time intervals)& $0.8565 \pm 0.0410$ & $0.6547 \pm 0.0181$ &2.11e-14\\
			LR ($K_0$ survival time intervals)&$0.6354 \pm 0.0620$ & $0.6024 \pm 0.0244$ &1.64e-11\\
			NN (ITCA-guided combined intervals)& $\mathbf{0.9623 \pm 0.0464}$ & $\mathbf{0.6855 \pm 0.0178}$ &\textbf{1.27e-15}\\
			LR (ITCA-guided combined intervals)& $0.8196 \pm 0.0222$ & $0.6236 \pm 0.0240$ &5.34e-10\\
			Cox regression (risk scores) & - & $0.6303\pm 0.0542$ & 2.04e-13\\
			\bottomrule
		\end{tabular}
		\begin{tablenotes}
			\item 
			Each criterion is computed by $5$-fold cross validation; its mean and standard error (i.e., standard deviation of the $5$ criterion values in the 5-fold CV, divided by $\sqrt{5}$) are listed; the average of the $5$ p-values corresponding to the Kendall's tau coefficients in the $5$-fold CV is also listed. The NN algorithm trained with ITCA-guided $K=7$ combined intervals achieves the best ITCA value, the best Kendall's tau, and the most significant average p-value.  
		\end{tablenotes}
	\end{threeparttable}
\end{table}
\subsection{Prediction of user demographics using mobile phone behavioral data}
\par 
One of the essential tasks in personalized advertising is to predict users' demographics (gender and age) using behavioral data.
A good predictive model is necessary for data-driven marketing decisions. 
To simplify the prediction of user ages, data scientists often first discretize ages into groups and then construct a multi-class classifier to predict age groups instead of exact ages \citep{talkingdata2016}.
However, the discretization step is heuristic and unjustified.
Here we use ITCA to determine age groups in a principled, data-driven way.
\par
We apply ITCA to the TalkingData mobile user demographics\footnote{The dataset is available at \url{https://www.kaggle.com/c/talkingdata-mobile-user-demographics}}, a public dataset of mobile phone users' behavioral data in China. 
Specifically, the dataset contains users' app usage, mobile device properties, genders, and ages.
Our goal is to predict a user's gender and age from mobile device and app usage.
In detail, we discretize male users' ages into $17$ ordinal groups encoded as M20$-$, M20--21, \dots, M48--49, and M50$+$ (where M20$-$, M20--21, and M50$+$ indicate male users whose ages are $<20$, $\ge 20$ \& $< 21$, and $\ge 50$, respectively). 
Similarly, we divide female users into $17$ ordinal age groups: F20$-$, F20--21, \dots, F48--49, and F50$+$. 
Together, we have $K_0=34$ classes to start with.
We use one-hot encoding to convert users' mobile devices and app usage data into $818$ features; after deleting the users with zero values in all features, we retain $n=23{,}556$ users.
Then we use  XGBoost \citep{Chen2016a} as the classification algorithm for its successes on similar prediction tasks in Kaggle competitions (see Supplementary Material Section~3.3 for details of data processing and algorithm training procedures).
\begin{figure}[hbt]
	\centering
	\includegraphics[width=\linewidth]{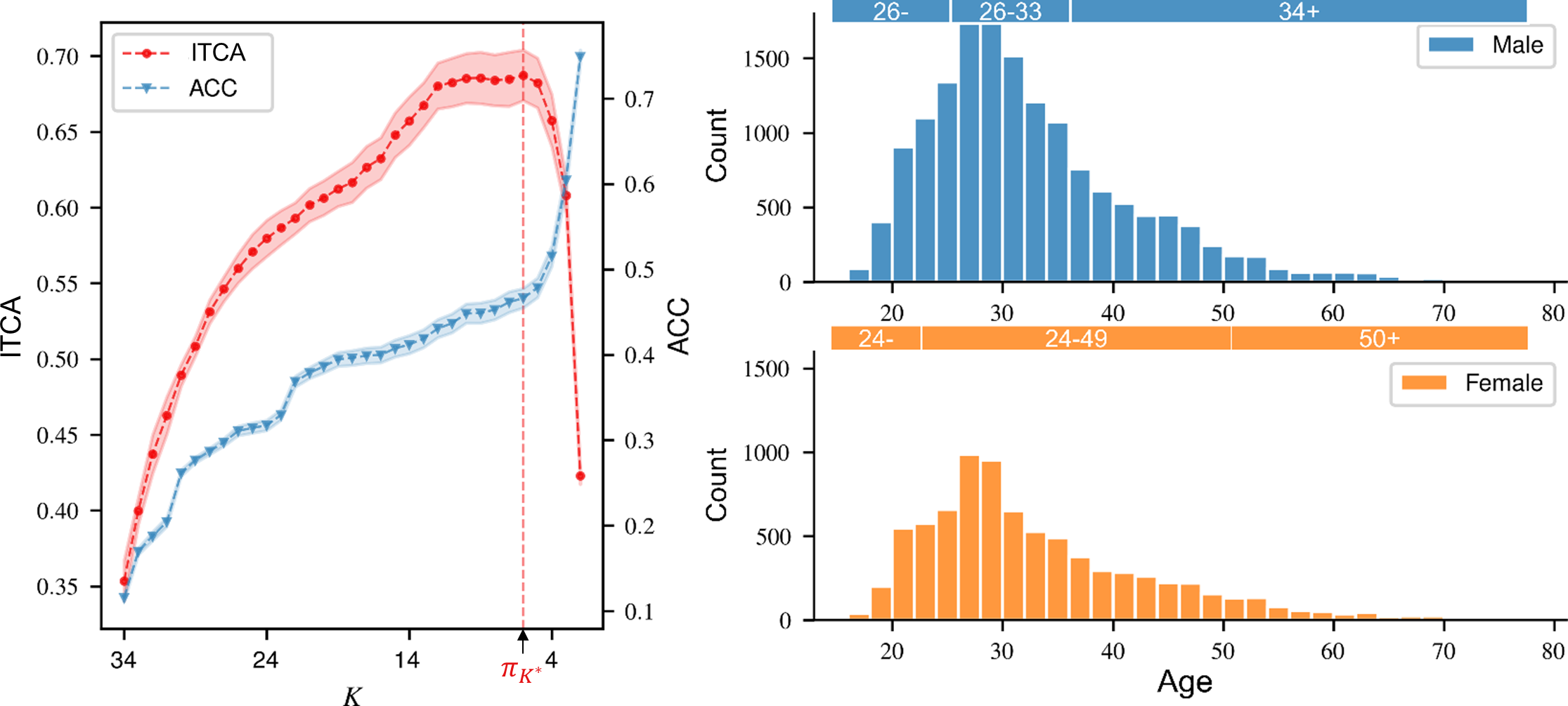}
	\caption{Results on TalkingData mobile user demographics dataset using XGBoost. Left: ITCA and ACC versus the number of combined classes 
		$K$, which ranges from $K_0=34$ to 2. 
		The criteria are estimated by $5$-fold CV, and their standard errors (marked by the shades) are calculated by the standard deviations in the $5$-fold CV divided by $\sqrt{5}$. 
		The best class combination $\pi_{K}^*$ (with $K=6$) is indicated by the vertical dashed line. 
		Right upper panel: the histogram of the ages of male users; $\pi_{K}^*$ suggests three age groups: M26$-$, M26--33, and M34$+$.
		Right lower panel: the histogram of the ages of female users; $\pi_{K}^*$ suggests three age groups: F24$-$, F24--49, and F50$+$. 
	}
	\label{fig:talking_data}
\end{figure}
\par
Since $30$ out of the $34$ classes are exact ages, it is intuitively too challenging to accurately predict the $34$ classes simply from users' phone devices and app usage. This is indeed the case, reflected by the low accuracy ($<0.35$) of XGBoost (Figure \ref{fig:talking_data}).
Hence, we use ITCA with the greedy search strategy to combine the $34$ ordinal classes into coarser, more meaningful age groups from the prediction perspective.
Note that we add the constraint for not combining a male class with a female class .
\par
Interestingly, ITCA suggests three different age groups for the male and female users (male: M26$-$, M26--33, and M34$+$; female: F24$-$, F24--49, and F50$+$) (Figure \ref{fig:talking_data}, right panel).
This result reveals a gender difference in the mobile phone behavioral data: female users have a wider middle age group (24--49 vs. male users' 26--33). 
A possible explanation of this gender difference is the well-known gender disparity in career development in China: more males undergo promotion into senior positions in middle 30s compared with females, who tend to slow down career development at young ages for reasons such as marriage and child birth \citep{Wei2011}.
This explanation is reasonable in that users' career development and mobile phone use behaviors are likely correlated. 
Hence, if we interpret the male and female age groups from the perspective of career development, 
we find a possible explanation of why the age of 34 is a change point for males but not for females, whose middle-to-senior change point is the age of 49, close to the retirement age of most females in China.
In summary, ITCA provides a data-driven approach to defining user age groups based on mobile phone behaviors, making it a potentially useful tool for social science research.     

\subsection{Detection of biologically similar cell types inferred from single-cell RNA-seq data}\label{subsec:single-cell}
\par
Recent advances in single-cell sequencing technologies provide unprecedented opportunities for scientists to decipher the mysteries of cell biology \citep{Wang2015,Stuart2019}. 
An important topic is to discern cell types from single-cell RNA-seq data, which profile transcriptome-wide gene expression levels in individual cells.
\par
Concretely, a single-cell RNA-seq dataset is processed into a data matrix of $n$ cells and $d_0$ genes, with the $(i, j)$-th entry as the expression level (i.e., log-transformed count of reads or unique molecular identifiers) of the $j$-th gene in the $i$-th cell.  
Starting from the matrix, a standard analysis pipeline involves the following steps \citep{Stuart2019}. 
First, principal component analysis is performed on the data matrix to reduce the column dimension from $d_0$ to $d \ll d_0$, resulting in a principal component matrix of $n$ cells and $d$ principal components.
Second, a clustering algorithm (e.g., the graph-based Louvain algorithm \citep{Blondel2008}) is applied to the principal component matrix to cluster the $n$ cells.
Finally, experts use knowledge to manually annotate the cell clusters with cell type labels. 
\par However, the annotated cell types might be ambiguous due to the subjectivity of setting parameter values in the above pipeline (e.g., the number of principal components $d$ and the clustering algorithm's parameters) and the uncertainty of the clustering step. As a result, if cells are overclustered, some annotated cell types might be biologically similar. 
\begin{figure}[hpbt]
	\centering
	\includegraphics[width=0.95\linewidth]{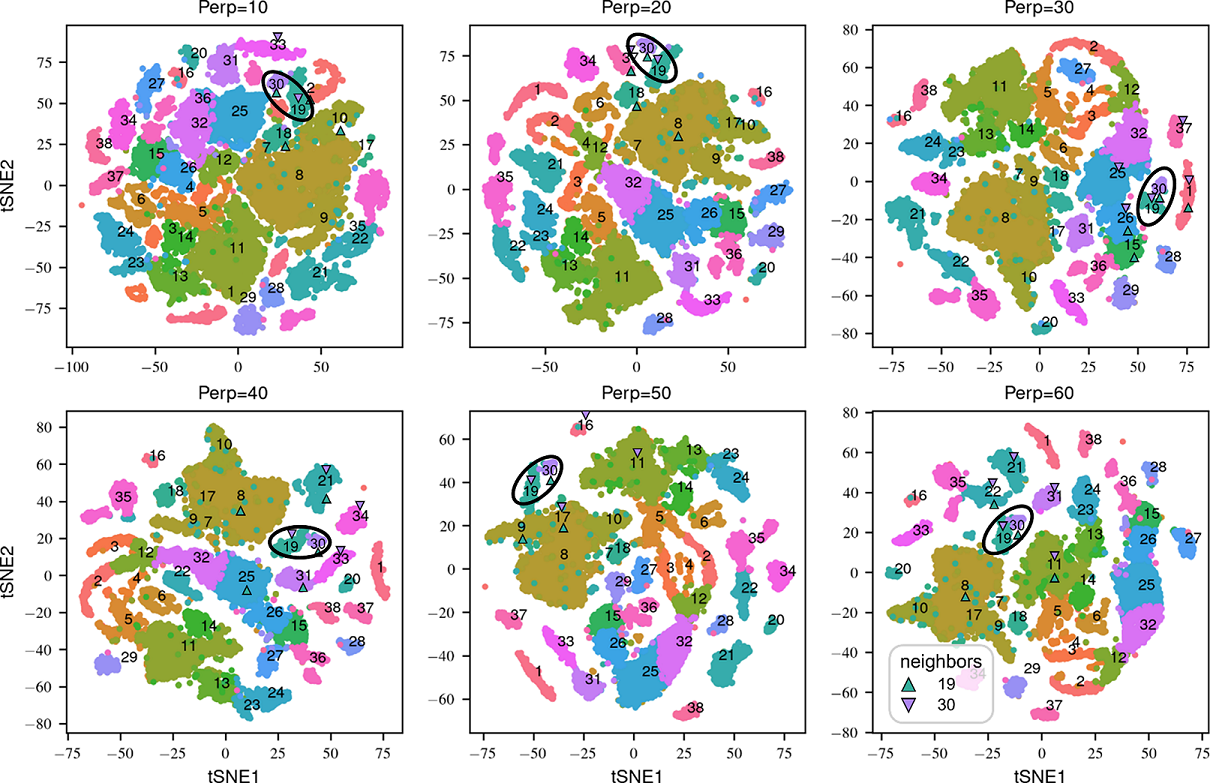}
	\caption{Visualizations of the cells in the hydra single-cell RNA-seq dataset using t-SNE under six perplexity values from 10 to 60. 
		The cell types 19 and 30 are marked by triangles and black circles.}\label{figure:tSNE-hydra}
\end{figure}
\begin{figure}[hpbt]
	\centering
	\includegraphics[width=0.75\linewidth]{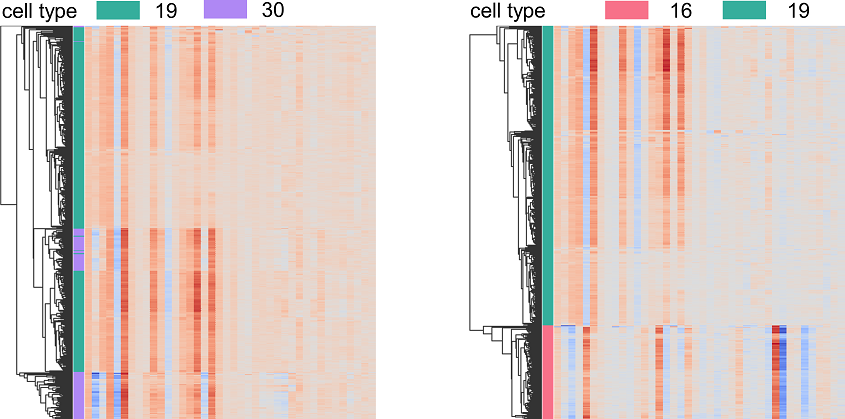}
	\caption{Heatmaps of the first 40 principal components of the hydra single-cell RNA-seq data. Row colors indicate the cell types: green for cell type 19 (458 cells; left and right panels), purple for cell type 30 (134 cells; left panel) and pink for cell type 16 (143 cells; right panel). 
		Hierarchical clustering can hardly distinguish cell types 19 and 30 (left panel) but can well separate cell types 16 and 19 (right panel).}
	\label{fig:hydra_heatmap}
\end{figure}
\par
This problem, the detection of biologically similar cell types, can be formulated as an application of ITCA.
Here, we use a single-cell RNA-seq dataset of hydra \citep{Siebert2019} as an example\footnote{The dataset is available at Broad Institute’s Single Cell Portal \url{https://singlecell.broadinstitute.org/single_cell/study/SCP260/}.}.
The processed dataset contains $n=25{,}052$ cells annotated into $K_0=38$ cells types (nominal class labels).
Following the procedure in  \citet{Siebert2019}, we use the first $d=40$ principal components as features. We choose the LDA as the classification algorithm for two reasons. First, our goal is to discover ambiguous cell types instead of achieving high prediction accuracy, so it is reasonable to choose a weak classification algorithm.
Second, the features are the principal components obtained from the log-transformed counts, and they are found to approximately follow a multivariate Gaussian distribution.
Given the large $K_0$, we apply ITCA using the greedy search algorithm. The result suggests the combination of cell type $19$ (``enEp tentacle": endodermal epithelial cells in tentacles) and cell type $30$ (``enEp tentnem(pd)": endodermal epithelial cells in tentacle nematocytes---suspected phagocytosis doublets), which indeed have similar cell type labels. 

\par
To evaluate this result, we examine cell types $19$ and $30$ using two-dimensional t-SNE visualization across a wide range of perplexity values (perplexity is the key hyperparameter of t-SNE). The t-SNE plots (Figure~\ref{figure:tSNE-hydra}) show that the two cell types are always direct neighbors of each other.
\par
We confirm this result by hierarchical clustering: the two cell types share similar gene expression patterns and are barely distinguishable in the first 40 principal components (Figure~\ref{fig:hydra_heatmap}, left panel).
As a control, we apply hierarchical clustering to distinguishing cell type 16 (``i\_neuron\_en3": neuronal cells of the interstitial lineage), whose number of cells is closest to that of cell type 19, from cell type 30. The result shows that, unlike cell types 19 and 30, cell types 16 and 30 are well separated (Figure~\ref{fig:hydra_heatmap}, right panel). Together, these evidence verifies the similarity of cell types 19 and 30, suggesting that ITCA can serve as a useful tool for identifying similar cell types and refining cell type annotations.


\section{Conclusion}\label{sec:conlustion}
We introduce ITCA, an information-theoretic criterion for combining ambiguous outcome labels in classification tasks; typical examples are in medical and social sciences where class labels are often defined subjectively.
ITCA automatically balances the trade-off between the increase in classification accuracy and the loss of classification  resolution, providing a data-driven criterion to guide class combination.
The simulation studies validate the effectiveness of ITCA and the proposed search strategies.
The four real-world applications demonstrate the wide application potential of ITCA. 
The theoretical analysis in Appendices~\ref{sec:remarks}--\ref{sec:pruning} characterizes several properties of ITCA and the search strategies, and it introduces a way to enhance the LDA algorithm for class combination.
\par 
One merit of ITCA is its universality: it can be applied with any classification algorithm without modification. While ITCA has an implicit trade-off between classification accuracy and classification  resolution, it is worth extending ITCA to incorporate user-specified weights for classification accuracy and classification  resolution. Another open question is how to incorporate users' predefined classes' importance to the ITCA definition; that is, users may prefer some important classes to stay as uncombined.
\par  
In addition to classification problems, ITCA may potentially serve as a model-free criterion for determining the number of clusters in clustering problems. We will investigate this direction in future work.

\section*{Acknowledgements}
This work was supported by the National Science Foundation DMS 2113754, Johnson \& Johnson WiSTEM2D Award, Alfred Sloan Foundation (Sloan Research Fellowship), and W.M. Keck Foundation (UCLA David Geffen School of Medicine W.M. Keck Foundation Junior Faculty Award) to Jingyi Jessica Li. This work was also supported by the National Key R\&D Program of China [2019YFA0709501], the Strategic Priority Research Program of the Chinese Academy of Sciences (CAS) [XDPB17], the Key-Area Research and Development of Guangdong Province [2020B1111190001], the National Natural Science Foundation of China [61621003], the National Ten Thousand Talent Program for Young Top-notch Talents, and the CAS Frontier Science Research Key Project for Top Young Scientist [QYZDB-SSW-SYS008] to Shihua Zhang.
Jingyi Jessica Li and Shihua Zhang are the corresponding authors.
\clearpage
\bibliographystyle{unsrtnat}
\bibliography{reference.bib}
\clearpage
\startcontents
\begin{appendices}
\renewcommand\thefigure{\thesection.\arabic{figure}}    
\setcounter{figure}{0}    
\renewcommand\theequation{\thesection.\arabic{equation}}    
\setcounter{equation}{0} 
	\section{Population-level ITCA and alternative criteria that may guide class combination}\label{sec:alternative_criteria}
To investigate the theoretical properties of ITCA, we define it at the population level as
{\small
	\begin{equation}\label{eq:p-ITCA}
		\text{p-ITCA}(\pi_K; \mathcal{D}_t, \mathcal{C}) := \sum_{k=1}^{K} [-\p(\pi_K(Y) = k) \log \p(\pi_K(Y) = k)] \cdot \p(\phi_{\pi_K}^{\mathcal C, \mathcal{D}_t}(\bX) = \pi_K(Y)| \pi_K(Y) = k) ,
	\end{equation}
}%
where $\phi_{\pi_K}^{\mathcal C, \mathcal{D}_t}$ is the classifier trained by the algorithm $\mathcal{C}$ on a finite training dataset $\mathcal{D}_t$. The population is used to evaluate the entropy contributions of $\pi_K$'s $K$ combined classes and the class-conditional prediction accuracies of $\phi_{\pi_K}^{\mathcal C, \mathcal{D}_t}$. The \textit{population-level ITCA (p-ITCA)} provides the basis of our theoretical analysis in in Appendix~\ref{sec:p-ITCA}.
\par
ITCA is aligned with the principle of maximum entropy \citep{Pal2003}, an application of the Occam’s razor.
For a fixed $K$, if the classifier $\phi_{\pi_K}^{\mathcal C, \mathcal{D}_t}$ has the same prediction accuracy for each class, i.e., $\p(\phi_{\pi_K}^{\mathcal C, \mathcal{D}_t}(\bX) =\pi_K(Y) |\pi_K(Y) = k)$ is a constant for all $k\in[K]$, then $\text{ITCA}$ in \eqref{eq:s-ITCA} is proportional to the entropy of $\pi_K(Y)$ and is maximized by the $\pi_K$ that results in the $K$ combined classes with the most balanced class probabilities. In the special case where the classifier performs perfect prediction for all $K$'s, i.e., $\p(\phi_{\pi_K}^{\mathcal C, \mathcal{D}_t}(\bX) =\pi_K(Y) |\pi_K(Y) = k) = 1$ for all $k\in[K]$ and all $K \in [K_0]$, ITCA becomes monotone increasing in $K$, i.e., the higher the classification  resolution, the larger the entropy.
	\par
	In addition to ITCA, we consider five alternative criteria that may guide class combination. The first two are commonly used criteria: classification accuracy and mutual information. The last three are our newly proposed criteria to balance the trade-off between classification accuracy and classification  resolution from three other perspectives.
	\par
	\textbf{Accuracy}. 
	It is the most commonly used criterion to evaluate the performance of a classification algorithm. For a class combination $\pi_K$, given a classification algorithm $\mathcal{C}$ and a size-$n$ dataset $\mathcal{D}$, the \textit{$R$-fold CV accuracy (ACC)} is
	\begin{equation}
		\text{ACC}^{\text{CV}}(\pi_K; \mathcal{D}, \mathcal{C}) := \frac{1}{R} \sum_{r=1}^R \frac{1}{|\mathcal{D}_v^r|} \sum\limits_{(\bX_i, Y_i)\in \mathcal{D}_v^r} \1\left(\phi_{\pi_K}^{\mathcal C, \mathcal{D}_t^r}(\bX_i)=\pi_K(Y_i)\right)\,,
	\end{equation}
	where the dataset $\mathcal{D}$ is randomly split into $R$ equal-sized folds, with the $r$-th fold $\mathcal{D}_v^r$ serving as the validation data and the union of the remaining $R-1$ folds $\mathcal{D}_t^r$ serving as the training data, and $\phi_{\pi_K}^{\mathcal C, \mathcal{D}_t^r}$ is the classifier trained by the algorithm $\mathcal{C}$ on $\mathcal{D}_t^r$. 
	Typically, $\text{ACC}^{\text{CV}}$ is used without class combination because it is maximized as $1$ when all classes are combined into one.
	Hence, intuitively, it is not an appropriate criterion for guiding class combination. In the following text, we refer to $\text{ACC}^{\text{CV}}$ as the ACC criterion.
	\par
	\textbf{Mutual information}. 
	It measures the dependence between two random variables, which, in the context of classification, can be the observed class label and the predicted class label. In this sense, the mutual information can be used as a criterion of classification accuracy.
	Following the definition of the mutual information of two jointly discrete random variables \citep{Cover1999}, we define the \textit{$R$-fold CV mutual information (MI)} of $\pi_K$ given $\mathcal{C}$ and $\mathcal{D}$ as
	\begin{align}\label{eq:MI}
		\small
		\text{MI}^{\text{CV}}(\pi_K; \mathcal{D}, \mathcal{C}) := \frac{1}{R} \sum_{r=1}^R \sum_{k_0=1}^{K_0} \sum_{k=1}^{K} & \left\{ \frac{\sum\limits_{(\bX_i, Y_i)\in \mathcal{D}_v^r} \1\left(Y_i=k_0,\, \phi_{\pi_K}^{\mathcal C, \mathcal{D}_t^r}(\bX_i)=k\right)}{|\mathcal{D}_v^r|} \right.\\
		& \left. \cdot \log \left( \frac{|\mathcal{D}_v^r| \sum\limits_{(\bX_i, Y_i)\in \mathcal{D}_v^r} \1\left(Y_i=k_0,\, \phi_{\pi_K}^{\mathcal C, \mathcal{D}_t^r}(\bX_i)=k\right)}{\left( \sum\limits_{(\bX_i, Y_i)\in \mathcal{D}_v^r} \1\left(Y_i=k_0\right) \right) \left( \sum\limits_{(\bX_i, Y_i)\in \mathcal{D}_v^r} \1\left( \phi_{\pi_K}^{\mathcal C, \mathcal{D}_t^r}(\bX_i)=k\right) \right) } \right) \right\} \,, \notag
	\end{align}
	where in the $r$-th fold, the mutual information is calculated for $\left\{\left(\phi_{\pi_K}^{\mathcal C, \mathcal{D}_t^r}(\bX_i),\, Y_i \right): (\bX_i, Y_i)\in \mathcal{D}_v^r\right\}$, i.e., between the predicted labels after class combination $\pi_K$ and the original labels. Note that the mutual information does not require $K_0 = K$ in \eqref{eq:MI}. The reason why we do not use the mutual information of $\left\{\left(\phi_{\pi_K}^{\mathcal C, \mathcal{D}_t^r}(\bX_i),\, \pi_K(Y_i) \right): (\bX_i, Y_i)\in \mathcal{D}_v^r\right\}$ (i.e., between the predicted labels and observed labels, both after class combination $\pi_K$) is that it increases as more classes are combined---an undesirable phenomenon. 
	In the following text, we refer to $\text{MI}^{\text{CV}}$ as the MI criterion.
	\par 
	\textbf{Adjusted accuracy}. 
	Neither the ACC criterion nor the MI criterion directly uses the class proportions.
	However, intuitively, it is easier to predict a data point from a larger class.
	To address this issue, we propose the \textit{$R$-fold CV adjusted accuracy (AAC)} to weigh each correctly predicted data point by the inverse proportion of the combined class to which the data point belongs:
	\begin{equation}\label{eq:AAC}
		\text{AAC}^{\text{CV}}(\pi_K; \mathcal{D}, \mathcal{C}) := \frac{1}{R} \sum_{r=1}^R \frac{1}{|\mathcal{D}_v^r|} \sum\limits_{(\bX_i, Y_i)\in \mathcal{D}_v^r} \frac{\1\left(\phi_{\pi_K}^{\mathcal C, \mathcal{D}_t^r}(\bX_i)=\pi_K(Y_i)\right)}{ p_{\pi_K}^{\mathcal{D}_v^r}(\pi_K(Y_i))}\,,
	\end{equation} 
	where $p_{\pi_K}^{\mathcal{D}_v}(\pi_K(Y_i))$ is the proportion of the combined class $\pi_K(Y_i)$ in $\mathcal{D}_v^r$, same as in the main text \eqref{eq:s-ITCA2}.
	The idea is straightforward: assigning smaller weights to the classes that take up larger proportions. 
	In the following text, we refer to $\text{AAC}^{\text{CV}}$ in (\ref{eq:AAC}) as the AAC criterion.
	An alternative proposal of the adjusted accuracy is to assign smaller weights to the classes that are combined from more original classes; however, this proposal does not work as well as the AAC criterion (see Supplementary Material Section~2.4).  
	\par
	\textbf{Combined Kullback–Leibler divergence}. 
	To balance the trade-off between the prediction accuracy and classification resolution, we also propose a combined Kullback–Leibler (CKL) divergence criterion that adds up (1) the divergence of the joint feature distribution estimated using combined class labels on the validation data $\mathcal{D}_v$ (denoted by $\widehat{F}_{\pi_K, \mathcal{D}_v}: \mathcal{X} \rightarrow [0,1]$) from that estimated using original class labels (denoted by $\widehat{F}_{\pi_{K_0}, \mathcal{D}_v}: \mathcal{X} \rightarrow [0,1]$) and (2) the divergence of the joint feature distribution estimated using predicted combined class labels (denoted by $\widehat{F}_{\phi_{\pi_K}^{\mathcal{C}, \mathcal{D}_t}, \mathcal{D}_v}: \mathcal{X} \rightarrow [0,1]$, where the classifier $\phi_{\pi_K}^{\mathcal{C}, \mathcal{D}_t}$ is trained on training data $\mathcal{D}_t$) from that estimated using combined class labels, i.e., $\widehat{F}_{\pi_K, \mathcal{D}_v}$. Accordingly, the \textit{$R$-fold CV CKL} is defined as 
	\begin{equation}\label{eq:CKL}
		\text{CKL}^{\text{CV}}(\pi_K; \mathcal{D}, \mathcal{C}) := \frac{1}{R} \sum_{r=1}^R \left[ D_{\text{KL}}\left(\widehat{F}_{\pi_K, \mathcal{D}_v^r}~||~ \widehat{F}_{\pi_{K_0}, \mathcal{D}_v^r}\right) + D_{\text{KL}}\left(\widehat{F}_{\phi_{\pi_K}^{\mathcal{C}, \mathcal{D}_t^r}, \mathcal{D}_v^r}~||~ \widehat{F}_{\pi_K, \mathcal{D}_v^r}\right) \right]\,.
	\end{equation} 
	In the following text, we refer to $\text{CKL}^{\text{CV}}$ as the CKL criterion. A challenge in calculating CKL is the estimation of $d$-dimensional joint feature distributions. To circumvent this challenge, we only calculate CKL when all class-conditional feature distributions are approximately Gaussian; that is, $\widehat{F}_{\pi_{K_0}, \mathcal{D}_v}$, $\widehat{F}_{\pi_K, \mathcal{D}_v}$, and $\widehat{F}_{\phi_{\pi_K}^{\mathcal{C}, \mathcal{D}_t}, \mathcal{D}_v}$ can all be approximated by Gaussian mixture models (see Supplementary Material Section~1for the computational detail of the KL divergence of two Gaussian mixture models). While this is an overly restrictive assumption, we use CKL as an alternative criterion to benchmark ITCA in simulation studies where this assumption holds (main text Section~\ref{subsec:sim}).
	\par
	\textbf{Prediction entropy}. The p-ITCA definition in \eqref{eq:p-ITCA} does not equate to but reminds us of the entropy of the distribution of $\left(\phi_{\pi_K}^{\mathcal C, \mathcal{D}_t}(\bX),\, \pi_K(Y)\right)$ conditional on $\phi_{\pi_K}^{\mathcal C, \mathcal{D}_t}(\bX) = \pi_K(Y)$, which we refer to as the \textit{population-level prediction entropy (p-PE)}:
	{\small%
		\begin{align}\label{eq:PE}
			&\text{p-PE}(\pi_K; \mathcal{D}_t, \mathcal{C}) =  \sum_{k=1}^{K} - \p\left( \phi_{\pi_K}^{\mathcal C, \mathcal{D}_t}(\bX) = \pi_K(Y) = k\right) \cdot \log \p\left( \phi_{\pi_K}^{\mathcal C, \mathcal{D}_t}(\bX) = \pi_K(Y) = k\right)\\
			= &\sum_{k=1}^{K} [-\p(\pi_K(Y) = k) \log \p(\pi_K(Y) = k)] \cdot \p(\phi_{\pi_K}^{\mathcal C, \mathcal{D}_t}(\bX) = \pi_K(Y)| \pi_K(Y) = k)  \notag \\
			& + \sum_{k=1}^{K} [-\p(\pi_K(Y) = k) \log \p(\phi_{\pi_K}^{\mathcal C, \mathcal{D}_t}(\bX) = \pi_K(Y)| \pi_K(Y) = k)] \cdot \p(\phi_{\pi_K}^{\mathcal C, \mathcal{D}_t}(\bX) = \pi_K(Y)| \pi_K(Y) = k)\,, \notag
		\end{align}
	}%
	where first term after the last equal sign is the p-ITCA. Accordingly, at the sample level, the \textit{$R$-fold CV PE} is
	{\small
		\begin{align}\label{eq:PE_CV}
			\text{PE}^{\text{CV}}(\pi_K; \mathcal{D}, \mathcal{C}) := \frac{1}{R} \sum_{r=1}^R \sum_{k=1}^{K} &- \frac{\sum\limits_{(\bX_i, Y_i)\in \mathcal{D}_v^r} \1\left(\phi_{\pi_K}^{\mathcal C, \mathcal{D}_t^r}(\bX_i)=\pi_K(Y_i)=k\right)}{|\mathcal{D}_v^r|} \\
			& \cdot \log \left( \frac{\sum\limits_{(\bX_i, Y_i)\in \mathcal{D}_v^r} \1\left(\phi_{\pi_K}^{\mathcal C, \mathcal{D}_t^r}(\bX_i)=\pi_K(Y_i)=k\right)}{|\mathcal{D}_v^r|} \right) \,. \notag
		\end{align}
	}
	In the following text, we refer to $\text{PE}^{\text{CV}}$ as the PE criterion. We argue that the definition of PE is not as intuitive as that of ITCA because PE only considers the data points that have labels correctly predicted; hence, compared to ITCA, PE does not fully capture the classification  resolution information.
	To compute ITCA, ACC, MI, AAC, CKL, and PE, we set the number of folds in CV to $R=5$ in all numerical analyses.
	
	\section{Some theoretical remarks}\label{sec:remarks}
	\par
	The simulation studies and applications have empirically verified the effectiveness of ITCA and the two search strategies: greedy search and BFS.
	One crucial question remains: is ITCA maximized at the true class combination when the sample size $n\to \infty$?
	To investigate the asymptotic property of ITCA, we define the population-level ITCA (p-ITCA) \eqref{eq:p-ITCA} in Appendix~\ref{sec:alternative_criteria}.
	For clarity, in the following text, we refer to the sample-level definition of ITCA in the main text \eqref{eq:s-ITCA} and \eqref{eq:s-ITCA2} as s-ITCA, and we refer to the $\text{ITCA}^\text{CV}$ in the main text \eqref{eq:ITCA_CV} as $\text{s-ITCA}^\text{CV}$. We can easily see that s-ITCA converges to p-ITCA in probability.
	
	In section~\ref{sec:p-ITCA}, we show that p-ITCA dose not always suggest to combine the classes that have the same class-conditional feature distribution, i.e., the classes we will refer to as ``same-distributed classes'' in the following. 
	Although this result seems counter-intuitive, it is aligned with the definition of p-ITCA, which not only considers prediction accuracy but also classification  resolution. 
	If the same-distributed classes dominate in class proportions, p-ITCA may prefer to keep them separate to maintain a not-so-degenerate classification  resolution (a degenerate classification  resolution means that all classes are combined as one).
	To investigate the class combination properties of p-ITCA, we define and analyze the class-combination curves and regions of the oracle classification algorithm and the LDA algorithm as examples.
	The results show that the oracle and LDA algorithms have different class-combination curves and regions.
	They also guide us to enhance the ability of the LDA algorithm for discovering the true class combination as described in section~\ref{subsec:enhance}.    
	In section~\ref{subsec:search property}, we analyze the properties of the search strategies with the oracle classification algorithm and show that BFS is equivalent to the exhaustive search.
	\subsection{Properties of p-ITCA with the oracle and LDA classification algorithms}\label{sec:p-ITCA}
	\begin{definition}[$\pi_K$'s induced partition]
	 	Given $K_0$ observed classes, a class combination $\pi_K$'s induced partition is defined as $K$ subsets of $[K_0]$: $\partition{K}{1}, \ldots, \partition{K}{K}$. That is, $\partition{K}{k} \cap \partition{K}{k'} = \varnothing$ if $1 \le k \neq k' \le K_0$, and $\cup_{k=1}^K \partition{K}{k} = [K_0]$.   
	\end{definition}
\begin{definition}[true class combination $\pi_{K^*}^*$] 
The true class combination $\pi_{K^*}^*$ is defined as the one whose induced partition $\truepartition{1}, \ldots, \truepartition{K}$ satisfies that the observed classes in each $\truepartition{k}$ have the same class-conditional feature distribution, and that the observed classes in $\truepartition{k}$ and $\truepartition{k'}$ have different class-conditional feature distributions if $1 \le k \neq k' \le K^*$. 
\end{definition}
\begin{definition}[set of split true class combinations $\mathcal A^*$] 
Suppose $\truecomb$ is the true class combination.
We define $\mathcal A^*:=\{\pi_K:\forall k \in[K],\;\exists k' \in[K^*] \text{ s.t. } \partition{K}{k} \subset \truepartition{k'} \}$ as the set of split true class combinations such that, in $\mathcal A^*$, each combination $\pi_K$'s induced partition is nested under the true class combination $\pi_{K^*}^*$'s induced partition; that is, each combined class defined by $\pi_K$ is a subset of a combined class defined by $\pi_{K^*}^*$.
\end{definition}
\begin{definition}[oracle classification algorithm $\mathcal{C}^*$]\label{def:oracle}
    	Suppose there are $K_0$ observed classes, and $\truecomb$ is the true class combination.
    	We define $\mathcal{C}^*$ as the oracle classification algorithm if its classifier for any class combination $\pi_K$, denoted by $\phi^{\mathcal{C}^*}_{\pi_{K}}$, satisfies the following property. 
    	\begin{itemize}
    	\item  For any data point $(\bX, Y)$, given $S^*(Y) := \pi_{K^*}^{*-1}( \pi^*_{K^*}(Y)) \subset [K_0]$, i.e., the set of observed classes that have the same class-conditional feature distribution as that of $Y$, 
    	\begin{align*}
    	\phi^{\mathcal{C}^*}_{\pi_{K}}(\bX) = \pi_K(k_0)\,, \text{ where } k_0 \sim \text{Multinomial}&\left(n=1, \text{support}= S^*(Y),\right.\\ &\left.\text{probabilities}=\left\{ \frac{p_{k_0}}{\sum_{k_0' \in S^*(Y)} p_{k_0'}}: k_0\in S^*(Y)\right\}\right)\,,
    	\end{align*}
    	where $p_{k_0}$ is the proportion of the $k_0$-th observed class. That is, $k_0$ is randomly picked from $S^*(Y)$ with probability equal to the proportion of $k_0$-th observed class in $S^*(Y)$. Moreover, $\phi^{\mathcal{C}^*}_{\pi_{K}}(\bX) = \pi_K\left(\phi^{\mathcal{C}^*}_{\pi_{K_0}}(\bX)\right)$.
    	\end{itemize}
    	Then the combined-class-conditional prediction accuracy of $\phi^{\mathcal{C}^*}_{\pi_{K}}$ is
		\begin{align*}\small
		 &\p\left(\phi^{\mathcal C^*}_{\pi_K}(\bX)=k \mid \pi_K(Y)=k\right) =\p\left(\phi^{\mathcal C^*}_{\pi_K}(\bX)=k \mid Y \in  \partition{K}{k}\right)  \\
		  =& \sum_{k_0\in \partition{K}{k}} \p\left(\phi^{\mathcal C^*}_{\pi_K}(\bX)=k \mid Y = k_0\right) \cdot \p(Y = k_0 \mid Y \in  \partition{K}{k})\\
		  =& \sum_{k_0\in \partition{K}{k}} \p\left(\phi^{\mathcal C^*}_{\pi_{K_0}}(\bX) \in \partition{K}{k} \mid Y = k_0\right) \cdot \p(Y = k_0 \mid Y \in  \partition{K}{k})\\
		  =& \sum_{k_0\in \partition{K}{k}} \p(Y = k_0 \mid Y \in  \partition{K}{k}) \sum_{k_0'\in \partition{K}{k}} \p\left(\phi^{\mathcal C^*}_{\pi_{K_0}}(\bX) = k_0' \mid Y = k_0\right) \\
		  =& \sum_{k_0\in \partition{K}{k}} \p(Y = k_0 \mid Y \in  \partition{K}{k}) \sum_{k_0'\in \partition{K}{k} \cap S^*(k_0)} \p\left(\phi^{\mathcal C^*}_{\pi_{K_0}}(\bX) = k_0' \mid Y = k_0\right) \\
		  =& \sum_{k_0\in \partition{K}{k}} \frac{p_{k_0}}{P_{\partition{K}{k}}} \sum_{k_0'\in \partition{K}{k} \cap S^*(k_0)} \frac{p_{k_0'}}{P_{S^*(k_0)}}\,, \quad k=1,\ldots,K\,,
		\end{align*}
		where $P_A := \sum_{k_0 \in   A}p_{k_0}$ denotes the total proportion of the observed classses in $A$.
\end{definition}
 
We give two examples to help readers understand the definition of the oracle classification algorithm. 
\begin{itemize}
    \item \textbf{Example 1}. All $K_0$ observed classes have distinct class-conditional feature distributions, i.e, $\pi_{K^*}^*=\pi_{K_0}$.
    For any data point $(\bX, Y)$, $\p(\phi^{\mathcal C^*}_{\pi_{K_0}}(\bX)=k_0 \mid Y=k_0)=1, \forall k_0 \in [K_0]$.
    Hence, $\phi^{\mathcal C^*}_{\pi_{K_0}}(\bX) = Y$ and $\phi^{\mathcal C^*}_{\pi_{K}}(\bX) = \pi_K(Y)$ with probability $1$; i.e., the oracle classification algorithm predicts perfectly for any class combination.
    \item \textbf{Example 2}. Among the $K_0$ observed classes, $\truepartition{1}=\{1, 2\}$, and the remaining observed classes $[K_0]\setminus \{1, 2\}$ have distinct class-conditional feature distributions. 
    For any data point $(\bX, Y)$, $\p(\phi_{\pi_{K_0}}^{\mathcal C^*}(\bX)=Y \mid Y=1)=p_1/(p_1 + p_2)$, $\p(\phi_{\pi_{K_0}}^{\mathcal C^*}(\bX)=Y \mid Y=2)=p_2/(p_1 + p_2)$, and $\p(\phi_{\pi_{K_0}}^{\mathcal C^*}(\bX)=Y \mid Y\notin \{1, 2\}) = 1$. 
\end{itemize}
\begin{lemma}\label{lemma:p-ITCA with oracle}
	Following the notations in Definition~4, suppose that among $K_0 \geq 2$ observed classes, the observed classes 1 and 2 have distinct class-conditional feature distributions.
	Let $\pi_K$ be a class combination that keeps the observed classes 1 and 2 uncombined, and consider 
	$\pi_{K-1}\in\mathcal{N}(\pi_{K})$ that combines the observed class 1 and 2 into one class. 
	Then we have $\textup{p-ITCA}(\pi_K;\mathcal{C}^*) > \textup{p-ITCA}(\pi_{K-1};\mathcal{C}^*)$. In other words, $\textup{p-ITCA}$ with the oracle classification algorithm would decrease if two distinct classes are combined into one.
\end{lemma}
\begin{proof}
	Following the notations and the combined-class-conditional prediction accuracy of $\phi^{\mathcal{C}^*}_{\pi_{K}}$ in Definition~4, we have
	\[
	\textup{p-ITCA}(\pi_K;\mathcal{C}^*) =  -\frac{p_1^2}{P_{S^*(1)}}\log p_1 - \frac{p_2^2}{P_{S^*(2)}}\log p_2  + \text{REM}\,,
	\]
	where $\text{REM}$ denotes the remaining terms that correspond to the combined classes in $\pi_K$ other than the observed classes $1$ and $2$, which stay as distinct classes in $\pi_K$.
	\begin{align*}
		\textup{p-ITCA}(\pi_{K-1};\mathcal{C}^*) &= - (p_1 + p_2)\log (p_1 + p_2)\left(\frac{p_1}{p_1 + p_2} \cdot \frac{p_1}{P_{S^*(1)}} + \frac{p_2}{p_1 + p_2}\cdot\frac{p_2}{P_{S^*(2)}}\right) +  \text{REM}\\
		& = - \log (p_1 + p_2)\left(\frac{p_1^2}{P_{S^*(1)}} + \frac{p_2^2}{P_{S^*(2)}}\right) +  \text{REM}
	\end{align*}
	Since $\log p_1 < \log(p_1 + p_2)$ and  $\log p_2 < \log(p_1 + p_2)$, we have 
	\[
	\textup{p-ITCA}(\pi_{K};\mathcal{C}^*) > \textup{p-ITCA}(\pi_{K-1};\mathcal{C}^*)\,.
	\]
\end{proof}
\par 
Lemma~\ref{lemma:p-ITCA with oracle} characterizes an important p-ITCA property with the defined oracle classification algorithm: p-ITCA does not combine two classes with distinct class-conditional feature distributions.

	Below we investigate how the proportions of two same-distributed classes, say $p_1$ and $p_2$, affect the class combination decision of p-ITCA: whether p-ITCA would suggest the two classes to be combined. In the space of $(p_1, p_2)$, denoted by 
	\begin{equation}\label{eq:Omega}
		\Omega = \{(p_1, p_2): p_1>0, p_2>0, p_1+p_2<1\} \subset [0,1]^2\,,
	\end{equation}
	we define the class-combination curve and region of a general classification algorithm $\mathcal{C}$; the curve and region are regarding whether the two same-distributed classes should be combined.
	\begin{definition}[class-combination curve and region]
		Among $K_0>2$ observed classes, suppose there are two same-distributed classes $S=\{1, 2\}$, and the other classes all have distinct class-conditional feature distributions.
		Denote by $\pi_{K_0 - 1} = \{(1,2), \dots, K_0\}$ the class combination that only combines the observed classes $1$ and $2$ into one class. We consider data-generating populations with varying $(p_1,p_2) \in \Omega$ in \eqref{eq:Omega}, and for each population we evaluate the performance of p-ITCA for combining the observed classes 1 and 2.
		Given a training dataset $D_t$ (needed for practical algorithms but not for the oracle algorithm) and a classification algorithm $\mathcal{C}$, we refer to the curve in $\Omega$
		\begin{equation}\label{eq:boundary}
			\textup{CC}(\pi_{K_0-1} || \pi_{K_0}; \mathcal{D}_t, \mathcal{C}) := \{(p_1, p_2) \in \Omega: \textup{p-ITCA}(\pi_{K_0}; \mathcal{D}_t, \mathcal{C}, p_1, p_2) =  \textup{p-ITCA}(\pi_{K_0-1}; \mathcal{D}_t, \mathcal{C}, p_1, p_2)\}
		\end{equation}
		as the class-combination curve (CC) of $\mathcal{C}$.
		For notation clarity, $p_1$ and $p_2$ are added to $\textup{p-ITCA}(\pi_{K_0}; \mathcal{D}_t, \mathcal{C})$ and $\textup{p-ITCA}(\pi_{K_0-1}; \mathcal{D}_t, \mathcal{C})$ to indicate that p-ITCA also depends on $p_1$ and $p_2$.
		
		Next, we define the class-combination region (CR) of algorithm $\mathcal{C}$ as 
		\begin{equation}\label{eq:CR}
			\textup{CR}(\pi_{K_0-1} || \pi_{K_0}; \mathcal{D}_t, \mathcal{C}) := \{(p_1, p_2) \in \Omega: \textup{p-ITCA}(\pi_{K_0-1}; \mathcal{D}_t, \mathcal{C}, p_1, p_2) -  \textup{p-ITCA}(\pi_{K_0}; \mathcal{D}_t, \mathcal{C}, p_1, p_2) > 0\}\,,
		\end{equation} 
		where $\pi_{K_0-1}$ improves the p-ITCA of $\pi_{K_0}$ and thus the two same-distributed classes should be combined.
	\end{definition}
	By the above definition, the following proposition holds for the oracle classification algorithm.
	\begin{proposition}[class-combination curve and region of the oracle classification algorithm]\label{prop:oracle_boundary}
		The class-combination curve of the oracle classification algorithm $\mathcal{C}^*$ has the closed form
		\begin{equation}\label{eq:oracle CC}
			\textup{CC}(\pi_{K_0-1} || \pi_{K_0}; \mathcal{C}^*) = \{(p_1, p_2) \in \Omega: 	p_1^2 \log p_1 + p_2^2 \log p_2 = (p_1 + p_2)^2 \log(p_1 + p_2)\}\,,
		\end{equation}
		and the class-combination region is 
		\begin{equation}\label{eq:oracle CR}
			\textup{CR}(\pi_{K_0-1} || \pi_{K_0}; \mathcal{C}^*) = \{(p_1, p_2) \in \Omega: 	p_1^2 \log p_1 + p_2^2 \log p_2 - (p_1 + p_2)^2 \log(p_1 + p_2) > 0\}\,.
		\end{equation} 
		
	\end{proposition}
	\begin{proof}
		Consider the p-ITCA before combing the observed classes 1 and 2, 
		\begin{equation*}
			\text{p-ITCA}(\pi_{K_0}; \mathcal{C}^*, p_1, p_2) = \sum_{k_0=1}^{K_0} -p_{k_0} \log p_{k_0} \cdot \p(\phi^{\mathcal{C}^*}_{\pi_{K_0}}(\bX) = Y | Y = k_0)\,.
		\end{equation*}
		Since $\phi^{\mathcal{C}^*}_{\pi_{K_0}}$ is an oracle classifier, we have
		\begin{align*}
			\p(\phi^{\mathcal{C}^*}_{\pi_{K_0}}(\bX) = Y | Y = 1) &= \frac{p_1}{p_1 + p_2}\,, \nonumber\\ \p(\phi^{\mathcal{C}^*}_{\pi_{K_0}}(\bX) = Y | Y = 2) &= \frac{p_2}{p_1 + p_2}\,, \nonumber\\
			\p(\phi^{\mathcal{C}^*}_{\pi_{K_0}}(\bX) = Y | Y = k_0) &= 1 \,, \quad k_0 = 3, \ldots, K_0\,.
		\end{align*}
		Hence,
		\begin{equation}\label{eq:p-ITCA-K0}
			\text{p-ITCA}(\pi_{K_0}; \mathcal{C}^*, p_1, p_2) = -\frac{p_1^2 \log p_1}{p_1+p_2} - \frac{p_2^2 \log p_2}{p_1+p_2} - \sum_{k_0=3}^{K_0} p_{k_0} \log p_{k_0}\,.
		\end{equation}
		Denote by $\pi_{K_0 - 1}$ the class combination that only combines the observed classes 1 and 2, i.e, $\pi_{K_0 - 1}(1) = \pi_{K_0 - 1}(2) = 1$, $\pi_{K_0 - 1}(k_0) = k_0-1$, $\forall k\in[K_0]\backslash\{1, 2\}$.
		The p-ITCA of $\pi_{K_0 - 1}$ is
		{\small
			\begin{equation*}
				\begin{split}
					\text{p-ITCA}(\pi_{K_0-1}; \mathcal{C}^*, p_1, p_2) = -\p(\pi_{K_0 - 1}(Y) = 1) \log \p(\pi_{K_0 - 1}(Y) = 1) \cdot \p(\phi^{\mathcal{C}^*}_{\pi_{K_0-1}}(\bX) = \pi_{K_0 - 1}(Y)| \pi_{K_0 - 1}(Y) = 1) \\
					- \sum_{k=2}^{K_0 - 1} \p(\pi_{K_0 - 1}(Y) = k) \log \p(\pi_{K_0 - 1}(Y) = k) \cdot \p(\phi^{\mathcal{C}^*}_{\pi_{K_0-1}}(\bX) = \pi_{K_0 - 1}(Y)| \pi_{K_0 - 1}(Y) = k)\,.
				\end{split}
			\end{equation*}
		}
		Since $\phi^{\mathcal{C}^*}_{\pi_{K_0-1}}$ is an oracle classifier, we have
		\begin{align*}
			\p(\phi^{\mathcal{C}^*}_{\pi_{K_0-1}}(\bX) = \pi_{K_0 - 1}(Y)| \pi_{K_0 - 1}(Y) = k) &= 1 \,, \quad \forall k \in [K_0-1]\,.
		\end{align*}
		Hence,
		\begin{equation}\label{eq:p-ITCA-K0minus1}
			\text{p-ITCA}(\pi_{K_0-1}; \mathcal{C}^*, p_1, p_2) = -(p_1 + p_2) \log(p_1+p_2) - \sum_{k_0=3}^{K_0} p_{k_0} \log p_{k_0} \,.
		\end{equation}
		Substituting (\ref{eq:p-ITCA-K0}) and (\ref{eq:p-ITCA-K0minus1}) into the definitions of the class-combination curve (\ref{eq:boundary}) and class-combination region (\ref{eq:CR}) and simplifying the forms, 
		we obtain the class-combination curve (\ref{eq:oracle CC}) and class-combination region \eqref{eq:oracle CR} of the oracle classification algorithm.
		
	\end{proof}
	\begin{figure}[hbpt]
	\centering
	\includegraphics[width=0.95\linewidth]{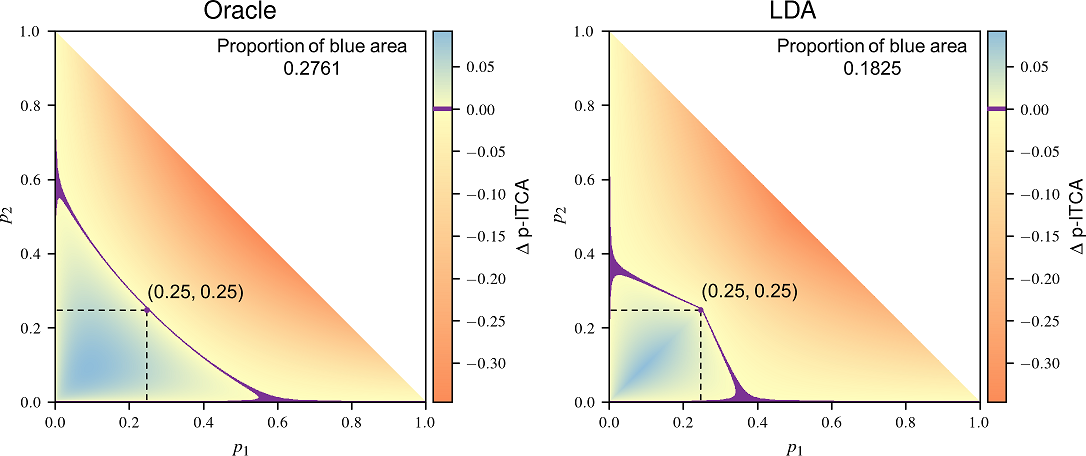}
	\caption{Regarding the combination of two same-distributed classes (with proportions $p_1$ and $p_2$), the improvement of p-ITCA , $\Delta \text{p-ITCA}(p_1, p_2; \mathcal{D}_t, \mathcal{C}) := \textup{p-ITCA}(\pi_{K_0-1}; \mathcal{D}_t, \mathcal{C}, p_1, p_2) -  \textup{p-ITCA}(\pi_{K_0}; \mathcal{D}_t, \mathcal{C}, p_1, p_2)$, of the oracle classification algorithm (left; $\mathcal{D}_t$ not needed; $\mathcal{C}^*$) and the LDA algorithm (right; $\mathcal{D}_\infty$; $\mathcal{C}^{\textup{LDA}}$). The blue areas indicate the class-combination regions of the oracle classification algorithm \eqref{eq:oracle CR} and the LDA classification algorithm \eqref{eq:lda_CR_simple} where $\Delta \text{p-ITCA}(p_1, p_2; \mathcal{D}_t, \mathcal{C}) > 0$ and thus the two classes will be combined. In each panel, the purple boundary ($|\Delta \text{p-ITCA}| < 10^{-3}$) between the blue area and the orange area indicates the class-combination curve of the corresponding algorithm; $(0.25, 0.25)$ is the point where $\Delta \text{p-ITCA}(p_1, p_2; \mathcal{D}_t, \mathcal{C}) = 0$ for both classification algorithms; the area of the class-combination region (the proportion of the blue area) is shown in the upper right corner.}
	\label{fig:p-ITCA_boudnary}
\end{figure}	
	Proposition~\ref{prop:oracle_boundary} shows that for the oracle classification algorithm, when one of $p_1$ and $p_2$ is close to zero and the other is less than $e^{-1/2} \approx 0.6$, by \eqref{eq:oracle CR} p-ITCA would combine the two classes (Supplementary Material Figure~S8 left panel shows the function $f(p) = p^2 \log p$, which monotone decreases for $p \in (0,e^{-1/2})$). Otherwise, if $p_1$ and $p_2$ have a not-too-small minimum and a large maximum, p-ITCA would keep the two classes separate to maintain a not-so-degenerate classification  resolution (Figure~\ref{fig:p-ITCA_boudnary}, left panel). Roughly speaking, unless one of the two classes is extremely small, p-ITCA would not suggest the two classes to be combined if their total proportion $p_1+p_2$ dominates. This result suggests that, even under the ideal scenario (i.e., with the oracle algorithm), p-ITCA would not suggest a combination of two non-trivial classes that would result in a dominant class.
	
	\par
	Next, we investigate the class-combination curve and region of the LDA algorithm.
	Different from the oracle classification algorithm, the LDA algorithm has its accuracy depending on the joint distribution of $(\bX, Y)$, on which we will make some additional assumptions.
	\begin{proposition}[class-combination curve and region of LDA]\label{prop:lda_boundary}
		Suppose there are $K_0=3$ observed classes, which contain two same-distributed classes.
		Without loss of generality, we assume that classes 1 and 2 have the same class-conditional feature distribution $ \mathcal{N}(\boldsymbol{0}, \sigma^2 \mathbf{I}_d)$, and that class $3$ has the class-conditional feature distribution $\mathcal{N}(\boldsymbol{\mu}, \sigma^2 \mathbf{I}_d)$ with $\boldsymbol{\mu} \neq \boldsymbol{0}$; hence, $||\boldsymbol{\mu}||$ is the Euclidean distance between the mean vectors of the two Gaussian distributions.  
		Denote by $\mathcal{D}_\infty$ the training dataset $\mathcal{D}_t$ with an infinite sample size, i.e., the LDA model trained on $\mathcal{D}_\infty$ has parameter estimates as the true parameters. 
		Then, for deciding whether $\pi_3=\{1, 2, 3\}$ should be combined as $\pi_2=\{(1, 2), 3\}$, the class-combination curve of the LDA algorithm $\mathcal{C}^{\mathrm{LDA}}$ has the following form
		\begin{alignat}{2}\label{eq:lda CC}
			& \textup{CC}(\pi_{2} || \pi_{3};\mathcal{D}_\infty&&, \mathcal{C}^{\textup{LDA}}) \notag \\
			= & \Big\{ (p_1, p_2) \in \Omega: &&\Phi\left(\frac{||\boldsymbol{\mu}||}{2\sigma} + \frac{\sigma}{||\boldsymbol{\mu}||} \log \frac{p_{1 \vee 2}}{p_3}\right) p_{1 \vee 2} \log p_{1 \vee 2} + \Phi\left(\frac{||\boldsymbol{\mu}||}{2\sigma} - \frac{\sigma}{||\boldsymbol{\mu}||} \log \frac{ p_{1 \vee 2}}{p_3}\right) p_3 \log p_3 =		\notag \\
			& &&\Phi\left(\frac{||\boldsymbol{\mu}||}{2\sigma} + \frac{\sigma}{||\boldsymbol{\mu}||} \log \frac{p_{1 + 2}}{p_3}\right) p_{1 + 2}\log p_{1 + 2} +
			\Phi\left(\frac{||\boldsymbol{\mu}||}{2\sigma} - \frac{\sigma}{||\boldsymbol{\mu}||} \log \frac{p_{1 + 2}}{p_3}\right)  p_3 \log p_3 \Big\},
		\end{alignat}		
		where $p_{1 \vee 2} := p_1 \vee p_2 = \max(p_1, p_2)$, $p_{1 + 2} := p_1 + p_2$, $p_3 = 1- p_{1+2}$, and $\Phi$ is the cumulative distribution function (CDF) of the univariate standard Gaussian distribution.
		The corresponding class-combination region is
		\begin{alignat}{2}\label{eq:lda CR}
			& \textup{CR}(\pi_{2} || \pi_{3};\mathcal{D}_\infty&&, \mathcal{C}^{\textup{LDA}}) \notag \\
			= & \Big\{ (p_1, p_2) \in \Omega: &&\Phi\left(\frac{||\boldsymbol{\mu}||}{2\sigma} + \frac{\sigma}{||\boldsymbol{\mu}||} \log \frac{p_{1 \vee 2}}{p_3}\right) p_{1 \vee 2} \log p_{1 \vee 2} + \Phi\left(\frac{||\boldsymbol{\mu}||}{2\sigma} - \frac{\sigma}{||\boldsymbol{\mu}||} \log \frac{ p_{1 \vee 2}}{p_3}\right) p_3 \log p_3
			\notag \\
			& &&-\Phi\left(\frac{||\boldsymbol{\mu}||}{2\sigma} + \frac{\sigma}{||\boldsymbol{\mu}||} \log \frac{p_{1 + 2}}{p_3}\right) p_{1 + 2}\log p_{1 + 2} 
			-
			\Phi\left(\frac{||\boldsymbol{\mu}||}{2\sigma} - \frac{\sigma}{||\boldsymbol{\mu}||} \log \frac{p_{1 + 2}}{p_3}\right)  p_3 \log p_3  > 0
			\Big\}\,.
		\end{alignat}	
		\par
		When $||\boldsymbol{\mu}|| \gg \sigma$, the class-combination curve (\ref{eq:lda CC}) reduces to
		\begin{equation}\label{eq:lda_boundary_simple}
			\textup{CC}(\pi_2 || \pi_{3}\,;\mathcal{D}_\infty, \mathcal{C}^{\textup{LDA}}) = \big\{(p_1, p_2) \in \Omega:\, p_{1 \vee 2}\log p_{1 \vee 2} = p_{1 + 2} \log p_{1 + 2}\big\}\,.   
		\end{equation}
		\par
		and the class-combination region (\ref{eq:lda CR}) reduces to
		\begin{equation}\label{eq:lda_CR_simple}
			\textup{CR}(\pi_2 || \pi_3\,;\mathcal{D}_\infty, \mathcal{C}^{\textup{LDA}}) = \big\{(p_1, p_2) \in \Omega:\, p_{1 \vee 2}\log p_{1 \vee 2} - p_{1 + 2} \log p_{1 + 2} > 0 \big\}\,.
		\end{equation}   
		Note that \eqref{eq:lda_boundary_simple} and \eqref{eq:lda_CR_simple} hold in general for $K_0 \ge 3$ observed classes regarding whether $\pi_{K_0}$ should be combined as $\pi_{K_0 - 1}$, which combines classes $1$ and $2$, when every classes $k$ with class-conditional feature distribution $\mathcal{N}(\boldsymbol{\mu}_k, \sigma^2 \mathbf{I}_d)$ satisfies that $||\boldsymbol{\mu}_k|| \gg \sigma$, $k=3, \ldots, K_0$.
	\end{proposition}
	\begin{proof}
		Without loss of generality, we assume that $\boldsymbol{\mu} =(||\boldsymbol{\mu}||, 0, \dots, 0)$ because one can always rotate $\boldsymbol{\mu}$ to obtain $(||\boldsymbol{\mu}||, 0, \dots, 0)$. 
		With the sample size $n\to \infty$, the parameter estimates of the LDA model converge to the true parameters in probability. 
		In our setting, without class combination (i.e., $\pi_3$), the LDA algorithm has the following three decision functions for the three observed classes.
		\begin{align*}
			\delta_1^{\pi_3}(\bX) &=  \log \p(Y=1|\bX) \text{ up to a constant} = \log p_1\,,  \\
			\delta_2^{\pi_3}(\bX) &=  \log \p(Y=2|\bX) \text{ up to a constant} = \log p_2\,,  \\
			\delta_3^{\pi_3}(\bX) &=  \log \p(Y=3|\bX) \text{ up to a constant} =  \frac{1}{\sigma^2}\bX^T\boldsymbol{\mu} - \frac{||\boldsymbol{\mu}||^2}{2\sigma^2} + \log p_3\,.
		\end{align*}
		Given a new data point $\boldsymbol{X}$, the LDA classifier predicts its label as $\phi_{\pi_3}^{\mathcal{C}^{\textup{LDA}},\mathcal{D}_\infty}(\bX) = \argmax_{k_0\in[3]}\, \delta_{k_0}^{\pi_3}(\bX)$.
		Consider the p-ITCA definition before combining the observed classes 1 and 2,
		\begin{equation*}
			\textup{p-ITCA}(\pi_3; \mathcal{D}_\infty, \mathcal{C}^{\textup{LDA}}, p_1, p_2) = \sum_{k_0=1}^3 -p_{k_0} \log p_{k_0}\p(\phi_{\pi_3}^{\mathcal{C}^{\textup{LDA}},\mathcal{D}_\infty}(\bX)=Y \mid Y=k_0)\,.
		\end{equation*}
		Without loss of generality, when $p_1 > p_2$, we immediately have $\delta_1^{\pi_3}(\bX) > \delta_2^{\pi_3}(\bX),\,\forall \bX \in \R^d$.
		Hence, we have the following class-conditional prediction accuracies. Conditional on $Y=1$,
		\begin{align*}
			\p(\phi_{\pi_3}^{\mathcal{C}^{\textup{LDA}},\mathcal{D}_\infty}(\bX)=Y\mid Y=1)&= \p(\delta_3^{\pi_3}(\bX) < \delta_1^{\pi_3}(\bX) \text{ and } \delta_2^{\pi_3}(\bX) < \delta_1^{\pi_3}(\bX) \mid Y=1)\\
			&= \p(\delta_3^{\pi_3}(\bX) < \delta_1^{\pi_3}(\bX) \mid Y=1)\\
			& = \p\left( \frac{1}{\sigma^2}\bX^T\boldsymbol{\mu} - \frac{||\boldsymbol{\mu}||^2}{2\sigma^2} + \log p_3 < \log p_1 \,\middle\vert\, Y = 1\right)\\
			& = \p\left(X_1 < \frac{||\boldsymbol{\mu}||}{2} + \frac{\sigma^2}{||\boldsymbol{\mu}||} \log \frac{p_1}{p_3} \,\middle\vert\, Y = 1 \right)\\
			&= \Phi\left(\frac{||\boldsymbol{\mu}||}{2\sigma} + \frac{\sigma}{||\boldsymbol{\mu}||} \log \frac{p_1}{p_3}\right)\,,
		\end{align*}
		where $X_1$ is the first element of $\bX$, and $X_1 \mid Y=1 \sim \mathcal{N}(0, \sigma^2)$.
		The fourth equation is based on $\bX^T\boldsymbol{\mu} = ||\boldsymbol{\mu}||$ because $\boldsymbol{\mu} =(||\boldsymbol{\mu}||, 0, \dots, 0)$. Conditional on $Y=2$,
		\begin{equation*}
			\p(\phi_{\pi_3}^{\mathcal{C}^{\textup{LDA}},\mathcal{D}_\infty}(\bX)=Y\mid Y=2) = \p(\delta_3^{\pi_3}(\bX) < \delta_2^{\pi_3}(\bX) \text{ and } \delta_1^{\pi_3}(\bX) < \delta_2^{\pi_3}(\bX) \mid Y=2) = 0\,.
		\end{equation*}
		Conditional on $Y=3$,
		\begin{align*}
			\p(\phi_{\pi_3}^{\mathcal{C}^{\textup{LDA}},\mathcal{D}_\infty}(\bX)=Y \mid Y=3) &= \p(\delta_1^{\pi_3}(\bX) < \delta_3^{\pi_3}(\bX) \text{ and } \delta_2^{\pi_3}(\bX) < \delta_3^{\pi_3}(\bX) \mid Y=3)\\
			&= \p(\delta_1^{\pi_3}(\bX) < \delta_3^{\pi_3}(\bX) \mid Y=3)\\
			& = \p\left(\log p_1 < \frac{1}{\sigma^2}\bX^T\boldsymbol{\mu} - \frac{||\boldsymbol{\mu}||^2}{2\sigma^2} + \log p_3 \,\middle\vert\, Y = 3\right)\\
			& = \p\left(X_1 > \frac{||\boldsymbol{\mu}||}{2} + \frac{\sigma^2}{||\boldsymbol{\mu}||} \log \frac{p_1}{p_3} \,\middle\vert\, Y = 3 \right)\\
			&= \Phi\left(\frac{||\boldsymbol{\mu}||}{2\sigma} - \frac{\sigma}{||\boldsymbol{\mu}||} \log \frac{p_1}{p_3}\right)\,,
		\end{align*}
		where the last equation holds because $X_1 \mid Y=3 \sim \mathcal{N}(||\boldsymbol{\mu}||, \sigma^2)$.
		
		Hence, p-ITCA before combining classes $1$ and $2$ is
		\begin{align}\label{eq:p-ITCA_LDA_pi3}
			&\textup{p-ITCA}(\pi_3; \mathcal{D}_\infty, \mathcal{C}^{\textup{LDA}}, p_1, p_2)\notag\\
			=& \left\{ \begin{array}{ll}
				-\Phi\left(\frac{||\boldsymbol{\mu}||}{2\sigma} + \frac{\sigma}{||\boldsymbol{\mu}||} \log \frac{p_1}{p_3}\right) p_1\log p_1 - \Phi\left(\frac{||\boldsymbol{\mu}||}{2\sigma} - \frac{\sigma}{||\boldsymbol{\mu}||} \log \frac{p_1}{p_3}\right) p_3 \log p_3  &  \text{if } p_1 > p_2 \\
				-\Phi\left(\frac{||\boldsymbol{\mu}||}{2\sigma} + \frac{\sigma}{||\boldsymbol{\mu}||} \log \frac{p_2}{p_3}\right) p_2\log p_2 - \Phi\left(\frac{||\boldsymbol{\mu}||}{2\sigma} - \frac{\sigma}{||\boldsymbol{\mu}||} \log \frac{p_2}{p_3}\right) p_3 \log p_3  &  \text{otherwise}
			\end{array}   \right. \notag\\
			=& -\Phi\left(\frac{||\boldsymbol{\mu}||}{2\sigma} + \frac{\sigma}{||\boldsymbol{\mu}||} \log \frac{p_{1 \vee 2}}{p_3}\right) p_{1 \vee 2}\log p_{1 \vee 2} - \Phi\left(\frac{||\boldsymbol{\mu}||}{2\sigma} - \frac{\sigma}{||\boldsymbol{\mu}||} \log \frac{p_{1 \vee 2}}{p_3}\right) p_3 \log p_3\,. 
		\end{align}
		
		After combining classes $1$ and $2$, the definition of p-ITCA becomes
		\begin{align*}
			\textup{p-ITCA}(\pi_2; \mathcal{D}_\infty, \mathcal{C}^{\textup{LDA}}, p_1, p_2) = & -p_{1+2} \log p_{1+2} \p(\phi_{\pi_2}^{\mathcal{C}^{\textup{LDA}},\mathcal{D}_\infty}(\bX)=\pi_2(Y) \mid \pi_2(Y)=1)\\
			& - p_3 \log p_3 \p(\phi_{\pi_2}^{\mathcal{C}^{\textup{LDA}},\mathcal{D}_\infty}(\bX)=\pi_2(Y) \mid \pi_2(Y)=2)\,.
		\end{align*}
		
		With $\pi_2$, the decision functions of the LDA algorithm becomes
		\begin{align*}
			\delta_1^{\pi_2}(\bX) &=  \log \p(\pi_2(Y)=1|\bX) \text{ up to a constant} = \log p_{1+2}\,,  \\
			\delta_2^{\pi_2}(\bX) &=  \log \p(\pi_2(Y)=2|\bX) \text{ up to a constant} =  \frac{1}{\sigma^2}\bX^T\boldsymbol{\mu} - \frac{||\boldsymbol{\mu}||^2}{2\sigma^2} + \log p_3\,.
		\end{align*}
		Hence, the LDA classifier $\phi_{\pi_2}^{\mathcal{C}^{\textup{LDA}},\mathcal{D}_\infty}$ has the following class-conditional prediction accuracies.
		Conditional on $\pi_2(Y)=1$,
		\begin{align*}
			\p(\phi_{\pi_2}^{\mathcal{C}^{\textup{LDA}},\mathcal{D}_\infty}(\bX)=\pi_2(Y) \mid \pi_2(Y)=1) &= \p(\delta_2^{\pi_2}(\bX) < \delta_1^{\pi_2}(\bX) \mid \pi_2(Y)=1)\\
			& = \p\left( \frac{1}{\sigma^2}\bX^T\boldsymbol{\mu} - \frac{||\boldsymbol{\mu}||^2}{2\sigma^2} + \log p_3 < \log p_{1+2} \,\middle\vert\, \pi_2(Y) = 1\right)\\
			& = \p\left(X_1 < \frac{||\boldsymbol{\mu}||}{2} + \frac{\sigma^2}{||\boldsymbol{\mu}||} \log \frac{p_{1+2}}{p_3} \,\middle\vert\, \pi_2(Y) = 1 \right)\\
			&= \Phi\left(\frac{||\boldsymbol{\mu}||}{2\sigma} + \frac{\sigma}{||\boldsymbol{\mu}||} \log \frac{p_{1+2}}{p_3}\right)\,,
		\end{align*}
		where the last equation holds because $X_1 \mid \pi_2(Y) = 1 \sim \mathcal{N}(0, \sigma^2)$.
		
		Conditional on $\pi_2(Y) = 2$,
		\begin{align*}
			\p(\phi_{\pi_2}^{\mathcal{C}^{\textup{LDA}},\mathcal{D}_\infty}(\bX)=\pi_2(Y) \mid \pi_2(Y)=2) &= \p(\delta_2^{\pi_2}(\bX) > \delta_1^{\pi_2}(\bX) \mid \pi_2(Y)=2)\\
			& = \p\left( \frac{1}{\sigma^2}\bX^T\boldsymbol{\mu} - \frac{||\boldsymbol{\mu}||^2}{2\sigma^2} + \log p_3 > \log p_{1+2} \,\middle\vert\, \pi_2(Y) = 1\right)\\
			& = \p\left(X_1 > \frac{||\boldsymbol{\mu}||}{2} + \frac{\sigma^2}{||\boldsymbol{\mu}||} \log \frac{p_{1+2}}{p_3} \,\middle\vert\, \pi_2(Y) = 1 \right)\\
			&= \Phi\left(\frac{||\boldsymbol{\mu}||}{2\sigma} - \frac{\sigma}{||\boldsymbol{\mu}||} \log \frac{p_{1+2}}{p_3}\right)\,,
		\end{align*}
		where the last equation holds because $X_1 \mid \pi_2(Y)=2 \sim \mathcal{N}(||\boldsymbol{\mu}||, \sigma^2)$.
		
		Hence, p-ITCA after combining classes $1$ and $2$ becomes 
		\begin{equation}
			\small
			\label{eq:p-ITCA_LDA_pi2}
			\textup{p-ITCA}(\pi_2; \mathcal{D}_\infty, \mathcal{C}^{\textup{LDA}}, p_1, p_2) = -\Phi\left(\frac{||\boldsymbol{\mu}||}{2\sigma} + \frac{\sigma}{||\boldsymbol{\mu}||} \log \frac{p_{1+2}}{p_3}\right) p_{1+2} \log p_{1+2} - \Phi\left(\frac{||\boldsymbol{\mu}||}{2\sigma} - \frac{\sigma}{||\boldsymbol{\mu}||} \log \frac{p_{1+2}}{p_3}\right)p_3 \log p_3\,.
		\end{equation}
		By \eqref{eq:p-ITCA_LDA_pi3} and \eqref{eq:p-ITCA_LDA_pi2}, we obtain the class-combination curve of LDA in (\ref{eq:lda CC}) and the class-combination region in \eqref{eq:lda CR}.
		When $||\boldsymbol{\mu}||\gg \sigma$, it is straightforward to see that (\ref{eq:lda CC}) and (\ref{eq:lda CR}) reduce to $(\ref{eq:lda_boundary_simple})$ and $(\ref{eq:lda_CR_simple})$, respectively.
		
		By similar derivations, we can show that $(\ref{eq:lda_boundary_simple})$ and $(\ref{eq:lda_CR_simple})$ hold in general for $K_0 \ge 3$ if the classes 1 and 2 are distinct from the other classes.
	\end{proof}
	
	Similar to Proposition~\ref{prop:oracle_boundary}, Proposition~\ref{prop:lda_boundary} shows that when one of $p_1$ and $p_2$ is close to zero and the other is less than $e^{-1}\approx 0.37$, by \eqref{eq:lda_CR_simple} p-ITCA would combine the two classes (Figure~S8, right panel shows the function $f(p) = p \log p$, which monotone decreases for $p \in (0, e^{-1})$).
	If $p_1$ and $p_2$ have a not-too-small minimum and a large maximum, p-ITCA would keep the two classes separate to maintain a not-so-degenerate classification  resolution (Figure~S8, right panel).
	The class-combiniation curve of LDA is quite different from that of the oracle classification algorithm. Moreover, compared with the oracle classification algorithm, LDA has a smaller chance of discovering the true class combination (shown by the smaller blue area in Figure~\ref{fig:p-ITCA_boudnary}).
	\subsection{Improvement of LDA as soft LDA for discovering class combination}\label{subsec:enhance}
	\par
	Seeing that the LDA algorithm is less powerful than the oracle classification algorithm for discovering the true combination, we consider modifying the LDA algorithm to improve its power.
	We first review how LDA predicts the class label of a data point $\bX$.
	Without loss of generality, we assume that two same-distributed classes $k_1, k_2 \in  [K_0]$ are distinguishable from the remaining $(K_0-2)$ classes.
	Suppose that classes $k_1$ and $k_2$ have a class-conditional feature distribution $\mathcal{N}(\boldsymbol{\mu}, \sigma^2\boldsymbol{I})$.
	Note the decision function $\delta_k^{\pi_{K_0}}(\bX) = \frac{1}{\sigma^2} \bX^T \boldsymbol{\mu} - \frac{||\boldsymbol{\mu}||^2}{2\sigma^2} + \log p_k$ for $k=k_1$ or $k_2$.
	Hence, if $p_{k_1} > p_{k_2}$, then $\delta_{k_1}^{\pi_{K_0}}(\bX) > \delta_{k_2}^{\pi_{K_0}}(\bX)$ holds for all $\bX \in \R^d$.
	Therefore, LDA would never predict that $Y=k_2$.
	In contrast, the oracle classification algorithm acts differently: if  a data point $(\bX, Y)$ has $Y\in \{k_1, k_2\}$, then the oracle classification algorithm would predict $Y$ as $k_1$ or $k_2$ with probability $p_1$ or $p_2$.
	
	We define soft LDA to mimic the oracle classification algorithm to enhance the ability of LDA for discovering the true class combination.
	The only difference between soft LDA and LDA is that the soft LDA predicts $Y$ randomly.
	Specifically, given a data point $\bX$, soft LDA computes the decision scores $\delta_k^{\pi_{K_0}}(\bX)$ for $k\in [K_0]$.
	Then soft LDA predicts $Y$ by drawing a sample with size one from a multinomial distribution $\text{Multinomial}(1, [K_0], \text{softmax}(\boldsymbol{\delta}^{\pi_{K_0}}(\bX)))$, where $\boldsymbol{\delta}^{\pi_{K_0}}(\bX):=(\delta_1^{\pi_{K_0}}(\bX), \dots, \delta_{K_0}^{\pi_{K_0}}(\bX))$ is a $K_0$-dimensional vector, and
	the $\text{softmax}$ function normalizes $\boldsymbol{\delta}^{\pi_{K_0}}(\bX)$ to a vector of probabilities that sum up to $1$.
	\par
	Interestingly, we show in Proposition~\ref{prop:soft_lda} that when the data-generating distribution is the LDA model with perfectly separated classes, the soft LDA algorithm trained with an infinite sample size equates the oracle classification algorithm. 
	\begin{proposition}\label{prop:soft_lda}
		Under the same setting in Proposition~\ref{prop:lda_boundary}, when $||\boldsymbol{\mu}||/\sigma \to \infty$, the soft LDA classification algorithm is the same as the oracle classification algorithm in Definiton~\ref{def:oracle}.
	\end{proposition}
	\begin{proof}
		The decision score of the soft LDA classification algorithm is the same as that of LDA:
		\begin{equation*}
			\boldsymbol{\delta}^{\pi_3}(\bX) = (\log p_1,~\log p_2,~\delta_3^{\pi_3}(\bX))\,,
		\end{equation*}
		where 
		\[ \delta_3^{\pi_3}(\bX) = \frac{1}{\sigma^2}\bX^T\boldsymbol{\mu} -\frac{||\boldsymbol{\mu}||^2}{2\sigma^2}  + \log p_3 = \frac{||\boldsymbol{\mu}||}{\sigma^2} X_1 -\frac{||\boldsymbol{\mu}||^2}{2\sigma^2}  + \log p_3 \,,
		\] and
		\begin{equation*}
			\textup{softmax}(\boldsymbol{\delta}^{\pi_3}(\bX))=\left(\frac{p_1}{p_1 + p_2 + \exp(\delta_3^{\pi_3}(\bX))},~ \frac{p_2}{p_1 + p_2 + \exp(\delta_3^{\pi_3}(\bX))}, ~\frac{\exp(\delta_3^{\pi_3}(\bX))}{p_1 + p_2 + \exp(\delta_3^{\pi_3}(\bX))}\right)\,.
		\end{equation*}
		Since 
		\begin{align*}
			X_1 \mid Y\in \{1,2\} \sim \mathcal{N}(0, \sigma^2) \quad \text{and} \quad X_1 \mid Y = 3 \sim \mathcal{N}(||\boldsymbol{\mu}||, \sigma^2)\,,
		\end{align*}
		we have
		\begin{align*}
			\delta_3^{\pi_3}(\bX) \mid Y\in \{1,2\} \sim \mathcal{N}\left(-\frac{||\boldsymbol{\mu}||^2}{2\sigma^2}  + \log p_3, \frac{||\boldsymbol{\mu}||^2}{\sigma^2}\right) \quad \text{and} \quad \delta_3^{\pi_3}(\bX) \mid Y = 3 \sim \mathcal{N}\left(\frac{||\boldsymbol{\mu}||^2}{2\sigma^2}  + \log p_3, \frac{||\boldsymbol{\mu}||^2}{\sigma^2}\right)\,.   
		\end{align*}
		Hence, when $||\boldsymbol{\mu}||/\sigma \to \infty$, we have the following limits in probability.
		\begin{align*}
			\textup{softmax}(\boldsymbol{\delta}^{\pi_3}(\bX))|Y\in\{1, 2\}\rightarrow \left(\frac{p_1}{p_1 + p_2}, \frac{p_2}{p_1 + p_2}, 0\right) \quad \text{and} \quad \textup{softmax}(\boldsymbol{\delta}^{\pi_3}(\bX))|Y=3\rightarrow (0, 0, 1)\,.
		\end{align*}
		Recall that soft LDA predicts $Y$ by drawing a sample with size one from $\text{ }(1, [3], \textup{softmax}(\boldsymbol{\delta}^{\pi_3}(\bX) ))$.
		Hence, we have 
		\begin{align*}
			&\p(\phi_{\pi_3}^{\mathcal{C}^{\textup{soft LDA}}, \mathcal{D}_\infty}(\bX) = 1 \mid Y \in \{1, 2\})  \rightarrow \frac{p_1}{p_1 + p_2}\,, \\
			&\p(\phi_{\pi_3}^{\mathcal{C}^{\textup{soft LDA}}, \mathcal{D}_\infty}(\bX) = 2 \mid Y \in \{1, 2\})  \rightarrow \frac{p_2}{p_1 + p_2}\,,\\
			&\p(\phi_{\pi_3}^{\mathcal{C}^{\textup{soft LDA}}, \mathcal{D}_\infty}(\bX) = 3 \mid Y=3) \rightarrow 1\,.
		\end{align*}
		Similarly, we have
		\begin{align*}
			&\p(\phi_{\pi_2}^{\mathcal{C}^{\textup{soft LDA}}, \mathcal{D}_\infty}(\bX) = 1 \mid \pi_2(Y) =1)  \rightarrow 1\,, \\
			&\p(\phi_{\pi_2}^{\mathcal{C}^{\textup{soft LDA}}, \mathcal{D}_\infty}(\bX) = 2 \mid \pi_2(Y) =2)  \rightarrow 1\,.
		\end{align*}
		Hence, the soft LDA algorithm approaches the oracle classification algorithm when $||\boldsymbol{\mu}||/\sigma \to \infty$.
	\end{proof}
	
	To numerically verify this result, we generate an array of simulated datasets with $K_0=3$, $K^*=2$, $l=5$, $n=5000$, and varying $(p_1, p_2)$, following the procedure described in Section \ref{sec:exp} of the main text.
	Each dataset, one per $(p_1, p_2)$ combination, contains 3 classes, with
	classes 1 and 2 as same-distributed.
	We set $0.1 \le p_1, p_2 \le 0.7$ and $p_1 + p_2 \le 0.8$ for numerical stability.
	Then we apply LDA and soft LDA to each simulated dataset and compute the improvement of s-ITCA$^{\textup{CV}}$ by combining classes 1 and 2.
	\begin{figure}[hpt]
		\centering
		\includegraphics[width=0.9\linewidth]{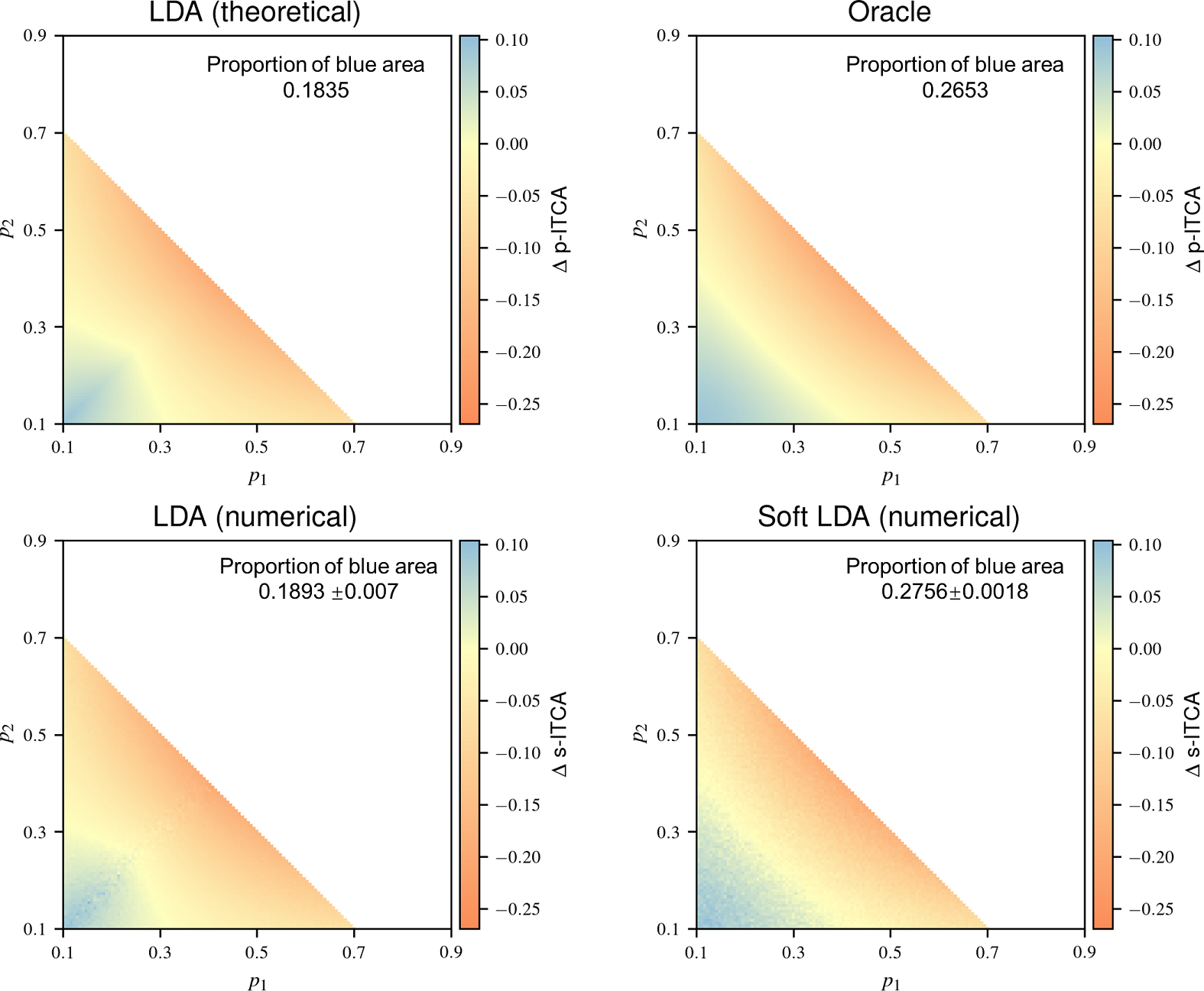}
		\caption{Regarding the combination of two same-distributed classes (with proportions $p_1$ and $p_2$), the improvement of s-ITCA$^{\textup{CV}}$ by class combination, i.e., $\Delta \text{s-ITCA}^{\textup{CV}}(p_1, p_2; \mathcal{D}_t, \mathcal{C}) := \textup{s-ITCA}^{\textup{CV}}(\pi_{K_0-1}; \mathcal{D}_t, \mathcal{C}, p_1, p_2) -  \textup{s-ITCA}^{\textup{CV}}(\pi_{K_0}; \mathcal{D}_t, \mathcal{C}, p_1, p_2)$ for $0.1 \le p_1, p_2 \le 0.7$ and $p_1 + p_2 \le 0.8$, of the LDA algorithm (bottom left; $\mathcal{C}^{\textup{LDA}}$) and the soft LDA algorithm (bottom right; $\mathcal{C}^{\textup{soft LDA}}$). 
			The blue areas indicate the class-combination regions where $\Delta \text{s-ITCA}^{\textup{CV}}(p_1, p_2; \mathcal{D}_t, \mathcal{C}) > 0$ and thus the two classes will be combined. 
			For comparison, $\Delta \text{p-ITCA}$ of the LDA and oracle classification algorithm are shown in the same range of $(p_1, p_2)$.
			In each panel, the yellow boundary between the blue area and the orange area indicates the class-combination curve of the corresponding algorithm; the area of the class-combination region (the proportion of the blue area) is shown in the upper right corner.
			In the second row, the numerical results are the average of 5-fold CV and the standard deviations of the five folds are shown (divided by $\sqrt{5}$).
		}\label{fig:s-ITCA_boudnary}
	\end{figure}	
	
	Figure \ref{fig:s-ITCA_boudnary} shows that the numerical results of s-ITCA$^{\textup{CV}}$ using LDA (lower left panel) match well with the results of p-ITCA using LDA (upper left panel). It is obvious that soft LDA (lower right panel) improves LDA (lower left panel) in terms class combination power (the proportion of blue area). This numerical result is also consistent with the theoretical result in Proposition~\ref{prop:soft_lda} about the agreement of soft LDA (lower right panel) with oracle (upper right panel).
	
	The improvement of soft LDA over LDA suggests that, for probabilistic classification algorithms, soft prediction of class labels has better power for class combination.
	
	We also used the simulated datasets to numerically approximate the class-combination regions of three commonly used classification algorithms (see Supplementary Material Section~4.2), including random forest (0.2228), gradient boosting trees (0.1961) and neural networks (0.1972); the number in each parenthesis indicates the proportion of blue area.
	Among the five classification algorithms (LDA, soft LDA, random forest, gradient boosting trees, and neural networks), random forest is second only to soft LDA in terms of finding the true class combination, possibly due to its ensemble nature.
\subsection{Properties of search strategies with the oracle classification algorithm}\label{subsec:search property}
\par We investigate the properties of the two local search strategies---greedy search and BFS---with the oracle classification algorithm. 
\begin{lemma}\label{lemma:ineq}
Consider $f(x):=-x^2\log x$. For $\forall p_1,p_2,p_3\in(0,1)$, such that $\sum_{i=1}^3 p_i <1$, if $f(p_1)+f(p_2) > f(p_1 + p_2)$, then $f(p_1 )+f(p_2 + p_3) > f(p_1+p_2+p_3)$.
\end{lemma}
\begin{proof}
	We formulate this problem as a nonlinear-constrained optimization problem and solve it numerically. 
	Consider the following optimization problem
	\begin{equation*}
		\begin{aligned}
			\min\quad & F(p_1, p_2, p_3) = f(p_1) + f(p_2 + p_3) - f(p_1 + p_2 + p_3)\\
			\text{s.t.}\quad & f(p_1)+f(p_2) > f(p_1 + p_2)\,, \\
			& p_1 + p_2 + p_3 < 1\,, \\
			& p_1, p_2, p_3 \in (0, 1)\,. \\
		\end{aligned}
	\end{equation*}
	We solve this nonlinear-constrained optimization problem numerically with the trust region method and find that $\min F(p_1, p_2, p_3) > 0$. Hence, $f(p_1 + p_2) > f(p_1) + f(p_2)$.
\end{proof}
\begin{theorem}\label{thm:search}
	Suppose there are $K_0$ observed classes. 
	Denote the class combinations found by the exhaustive search, BFS and greedy search with the oracle classification algorithm by $\escomb$, $\bfscomb$ and $\gscomb$, which correspond to $K_{\rm{ES}}$, $K_{\rm{BFS}}$ and $K_{\rm{GS}}$ combined classes, respectively.
	Then $\escomb, \bfscomb, \gscomb \in\mathcal{A}^*$, the set of split true class combinations, and $\escomb=\bfscomb$.
\end{theorem}
\begin{proof}
By Lemma~\ref{lemma:p-ITCA with oracle}, p-ITCA would decrease if two distinct classes are combined into one. 
Hence, $\escomb, \bfscomb, \gscomb \in\mathcal{A}^*$.
We will then show that $\bfscomb=\escomb$ by proving that there exists a path from $\pi_{K_0}$ to $\escomb$, denoted as $P^*:=\{\pi_{K_0 - i}\}_{i=0}^{K_0 - K_{\rm{ES}}}\subset \mathcal{A}^*$, that satisfies $\pi_{K_0 - i} \in \mathcal{N}(\pi_{K_0 - i + 1})$ for $i\in[K_0 - K_{\rm{ES}}]$, and
	\begin{equation}\label{eq:path_ineq}
		\text{p-ITCA}(\pi_{K_0}; \mathcal{C}^*) < \text{p-ITCA}(\pi_{K_0-1}; \mathcal{C}^*)<\cdots<\text{p-ITCA}(\pi_{K_{\rm{ES}} + 1}; \mathcal C^*) < \text{p-ITCA}(\pi_{K_{\rm{ES}}}; \mathcal C^*)\,. 
	\end{equation}
where $\pi_{K_{\rm{ES}}}:=\escomb$.
\par
We will show the existence of $P^*$ using a counterproof. Suppose that $P^*$ does not exist. Then for any path $P$ from $\pi_{K_0}$ to $\escomb$, there must exist two class combinations $\pi_{K_0 - i + 1}$, $\pi_{K_0 - i} \in P$ such that $\pi_{K_0 - i} \in \mathcal{N}(\pi_{K_0 - i + 1})$, and
\begin{equation*}
\text{p-ITCA}(\pi_{K_0 - i + 1}; \mathcal{C}^*) \geq \text{p-ITCA}(\pi_{K_0 - i}; \mathcal{C}^*)\,.     
\end{equation*}
Without loss of generality, we assume that $\pi_{K_0 - i}$ combines $\pi_{K_0 - i + 1}$'s classes $1$ and $2$.
Following Definition~\ref{def:oracle} and its notations, we have
\[
	\text{p-ITCA}(\pi_{K_0 - i + 1}; \mathcal{C}^*) = -\frac{p_{1}^2}{P_{S^*(1)}}\log p_{1} - \frac{p_{2}^2}{P_{S^*(2)}}\log p_{2} +  \text{REM}\,,
\]
where $p_i$ is the proportion of $\pi_{K_0 - i + 1}$'s class $i$, $\text{REM}$ indicates the remaining terms based on $\pi_{K_0 - i + 1}$'s other classes (except classes $1$ and $2$), and $P_{S^*(i)}$ is the total proportion of the classes that have the same distribution as $\pi_{K_0 - i + 1}$'s class $i$, $i=1, 2$.

Denote $f(p):=-p^2 \log p$, where $0<p<1$. 
Since $\pi_{K_0-i} \in \mathcal{A}^*$, $\pi_{K - i + 1}$'s classes $1$ and $2$ must follow the same class-conditional feature distribution, so $P_{S^*(1)} = P_{S^*(2)}$.
Then
	\begin{align*}
		\text{p-ITCA}(\pi_{K_0 - i + 1}; \mathcal{C}^*) &= \frac{f(p_{1}) + f(p_{2})}{P_{S^*(1)}}  + \text{REM}\,,\\
		\text{p-ITCA}(\pi_{K_0 - i}; \mathcal{C}^*) &= \frac{f(p_{1}+ p_{2})}{P_{S^*(1)}}  + \text{REM}\,.
	\end{align*}
Hence
\[
\text{p-ITCA}(\pi_{K_0 - i + 1}; \mathcal{C}^*) - \text{p-ITCA}(\pi_{K_0 - i}; \mathcal{C}^*) \geq 0
\]
implies that
\begin{align}\label{ineq:p1_p2}
    f(p_{1}) + f(p_{2}) \ge f(p_{1}+ p_{2})\,.
\end{align}
\par
Since $\escomb$ is in the same path $P$ as $\pi_{K_0 - i + 1}$ and $\pi_{K_0 - i}$, $\escomb$ must have a class, denoted by $k_{\rm ES}$, that contains $\pi_{K_0 - i + 1}$'s classes $1$ and $2$, i.e., $\pi_{K_0 - i + 1}^{-1}(1) \subset (\escomb)^{-1}(k_{\rm ES})$ and $\pi_{K_0 - i + 1}^{-1}(2) \subset (\escomb)^{-1}(k_{\rm ES})$.
 Since $\escomb \in \mathcal{A}^*$, the observed classes contained in $k_{\rm ES}$ must be in $S^*(1)$, the set of observed classes that have the same class-conditional feature distribution as that of $\pi_{K-i+1}$'s class $1$.

Now we consider a class combination $\pi'_{K_{\rm ES} + 1}$ such that $\escomb \in \mathcal{N}(\pi'_{K_{\rm ES} + 1})$ and $\pi'_{K_{\rm ES} + 1}$ separates  $\pi_{K_0 - i + 1}$'s class $1$ from $\escomb$'s class $k_{\rm ES}$.
Denote the proportion of $\escomb$'s class $k_{\rm ES}$ as $p_{k_{\rm ES}}$, and define $p_{3}:=p_{k_{\rm ES}} - p_{1} - p_{2}$.
Then
	\begin{align*}
		\text{p-ITCA}(\pi'_{K_{\rm ES} + 1}; \mathcal{C}^*) &= \frac{f(p_{1}) + f(p_{2} + p_{3})}{P_{S^*(1)}}  + \text{REM}\,,\\
		\text{p-ITCA}(\escomb; \mathcal{C}^*) &= \frac{f(p_{1}+ p_{2} + p_{3})}{P_{S^*(1)}}  + \text{REM}\,,
	\end{align*}
where $\text{REM}$ indicates the remaining terms based on $\pi'_{K_{\rm ES} + 1}$'s other classes.
Then by \eqref{ineq:p1_p2} and Lemma~\ref{lemma:ineq}, we have
\begin{equation*}\label{eq:thm_ineq2}
 \text{p-ITCA}(\pi'_{K_{\rm ES} + 1}; \mathcal{C}^*) - \text{p-ITCA}(\escomb; \mathcal{C}^*) =  \frac{f(p_{1}) + f(p_{2} + p_{3}) - f(p_{1}+ p_{2} + p_{3})}{P_{S^*(1)}}\geq0\,,  
\end{equation*}
contradicting the fact that $\escomb$ maximizes p-ITCA by definition. 
Hence, \eqref{eq:path_ineq} holds, and 
BFS can find $\escomb$ through the path $P^*$.
Therefore $\bfscomb = \escomb$.
\end{proof}
\par
Theorem~\ref{thm:search} shows that with the oracle classification algorithm, all three search strategies would not combine two distinct classes, and BFS is equivalent to the exhaustive search. 
Our simulation results also empirically show that the greedy search and BFS work well with the LDA classification algorithm.

\section{Search space pruning}\label{sec:pruning}
	Compared with the exhaustive search, the greedy search and BFS  significantly reduce the number of times needed to compute s-ITCA.
	We note that if the following assumption holds for a classification algorithm, one can further prune the search space of possible class combinations.  
	\begin{assumption}[classification algorithm property]\label{assumpiton:phi}
		Consider a classification algorithm $\mathcal{C}$ and a training dataset $\mathcal{D}_t$. The property states that, for any $K=3,\ldots,K_0 \ge 3$, the algorithm $\mathcal{C}$ satisfies the following inequality for any $i,j \in [K]$ and the $\pi_{K-1}^{(i,j)}$ that only combines $\pi_K$'s (observed or combined) classes $i$ and $j$ into a new combined class 1. 
		\begin{align}
			& \sum_{k=2}^{K-1} \left[-\p\left(\pi_{K-1}^{(i,j)}(Y) = k\right) \log \p\left(\pi_{K-1}^{(i,j)}(Y) = k\right)\right] \cdot \p\left(\phi_{\pi_{K-1}^{(i,j)}}^{\mathcal C, \mathcal D_t}(\bX) = \pi_{K-1}^{(i,j)}(Y) \,\middle\vert\, \pi_{K-1}^{(i,j)}(Y) = k\right)\notag\\
			\ge & \sum_{k\in [K]\backslash \{i, j\}} \left[-\p(\pi_{K}(Y) = k) \log \p(\pi_{K}(Y) = k)\right] \cdot \p\left(\phi_{\pi_{K}}^{\mathcal{C}, \mathcal D_t}(\bX) = \pi_{K}(Y)\mid \pi_{K}(Y) = k\right)\,. 
		\end{align}
		In other words, the total contribution of $\pi_K$'s other classes (except classes $i$ and $j$) to p-ITCA does not decrease after $\pi_K$'s classes $i$ and $j$ are combined.
	\end{assumption}
	\par
	A sufficient condition for Assumption \ref{assumpiton:phi} is that a classification algorithm has the same class-conditional prediction accuracies, before and after class combination, for the classes that are not combined. We can easily verify that the oracle, LDA, and soft LDA algorithms satisfy this sufficient condition and thus Assumption \ref{assumpiton:phi}.
	In addition, Assumption \ref{assumpiton:phi} holds for the classification algorithms that use the one-vs-the-rest scheme. 
	
	Under Assumption \ref{assumpiton:phi}, we derive the following condition for class combination by p-ITCA.
	
	\begin{proposition}[class combination condition]\label{prop:criterion}
		If Assumption \ref{assumpiton:phi} holds, p-ITCA will guide $\pi_K$'s classes $i$ and $j \in [K]$ to be combined if and only if:
		\begin{align}\label{eq:criterion}
			& \; \p\left(\phi_{\pi_{K-1}^{(i,j)}}^{\mathcal C, \mathcal D_t}(\bX) = \pi_{k-1}^{(i,j)}(Y) \,\middle\vert\, \pi_K(Y)\in \{i, j\}\right) \notag\\ 
			> & \; \frac{p_i (\log p_i) \p\left(\phi_{\pi_K}^{\mathcal C, \mathcal D_t}(\bX)=\pi_K(Y) \mid \pi_K(Y)=i\right) + p_j (\log p_j) \p\left(\phi_{\pi_K}^{\mathcal C, \mathcal D_t}(\bX)=\pi_K(Y) \mid \pi_K(Y)=j\right)}{(p_i + p_j)\log(p_i + p_j)}\,,
		\end{align}
		where $p_i$ and $p_j$ are the proportions of $\pi_K$'s classes $i$ and $j$, respectively.
	\end{proposition}
	\begin{proof}
		The proof is straightforward. Classes $i$ and $j$ will be combined if and only if
		\begin{equation}
			\text{p-ITCA}\left(\pi_{K-1}^{(i,j)}; \mathcal{D}_t, \mathcal{C}\right) > \text{p-ITCA}(\pi_{K}; \mathcal{D}_t, \mathcal{C})\,,
		\end{equation}
		which is equivalent to
		\begin{align}\label{eq:compare_p-ITCA1}
			& \sum_{k=1}^{K-1} \left[-\p\left(\pi_{K-1}^{(i,j)}(Y) = k\right) \log \p\left(\pi_{K-1}^{(i,j)}(Y) = k\right)\right] \cdot \p\left(\phi_{\pi_{K-1}^{(i,j)}}^{\mathcal C, \mathcal D_t}(\bX) = \pi_{K-1}^{(i,j)}(Y) \,\middle\vert\, \pi_{K-1}(Y) = k\right) \notag\\ 
			>& \sum_{k=1}^K [-\p(\pi_{K}(Y) = k) \log \p(\pi_{K}(Y) = k)] \cdot \p\left(\phi_{\pi_{K}}^{\mathcal C, \mathcal D_t}(\bX) = \pi_{K}(Y) \mid \pi_{K}(Y) = k\right)\,.
		\end{align}
		Under Assumption~\ref{assumpiton:phi}, it is straightforward to see that \eqref{eq:criterion} is a sufficient condition for  (\ref{eq:compare_p-ITCA1}), which completes the proof.
	\end{proof}
	
	Proposition~\ref{prop:criterion} provides a rule for pruning the search space. The left-hand side of \eqref{eq:criterion} must be no greater than $1$. Hence, if the accuracies $\p(\phi_{\pi_K}^{\mathcal C, \mathcal D_t}(\bX)=Y \mid Y=i)$ and $\p(\phi_{\pi_K}^{\mathcal C, \mathcal D_t}(\bX)=Y \mid Y=j)$ are high enough such that the right-hand side is greater than 1, there is no way for \eqref{eq:criterion} to hold, and we can remove $\pi_{K-1}^{(i,j)}$ from the search space.

	Our simulation results show that this pruning strategy is effective and can reduce the number of s-ITCA evaluations by about half (Tables \ref{table:search-8} in the main text and Table~S3 in Supplementary Material). 
\end{appendices}
\stopcontents
\clearpage
\setcounter{section}{0}
\setcounter{figure}{0}
\setcounter{table}{0}
\setcounter{equation}{0}
\makeatletter 
\renewcommand{\thefigure}{S\@arabic\c@figure}
\renewcommand{\thetable}{S\@arabic\c@table}
\renewcommand{\theequation}{S\@arabic\c@equation}
\makeatother
\section*{\Large Supplementary Material}
\section{Approximate KL divergence between two Gaussian mixture models}\label{sec:kl_gmm}
The KL divergence from a Gaussian distribution $\mathcal{N}(\boldsymbol{\mu}_2, \boldsymbol{\Sigma}_2)$ to another Gaussian distribution $\mathcal{N}(\boldsymbol{\mu}_1, \boldsymbol{\Sigma}_1)$ has the following closed-form
\begin{equation}
	D_{\textup{KL}}(\mathcal{N}(\boldsymbol{\mu}_1, \boldsymbol{\Sigma}_1)~||~ \mathcal{N}(\boldsymbol{\mu}_2, \boldsymbol{\Sigma}_2)) =\frac{1}{2}\log\frac{|\boldsymbol{\Sigma}_2|}{|\boldsymbol{\Sigma}_1|} + \frac{1}{2}\tr(\boldsymbol{\Sigma}_2^{-1}\boldsymbol{\Sigma}_1) + \frac{1}{2}(\boldsymbol{\mu}_1 - \boldsymbol{\mu}_2)^T \boldsymbol{\Sigma}_2^{-1}(\boldsymbol{\mu}_1 - \boldsymbol{\mu}_2) -\frac{d}{2}\,. 
\end{equation} 
\par However, there exists no closed-form KL divergence for Gaussian mixture models (GMMs).
Durrieu \textit{et al.} gave the lower and upper bounds for approximating the KL divergence from one GMM to another \citep{Durrieu2012} . 
Let $\mathcal F:= \sum_{i=1}^{K} p^f_i \mathcal{N}(\boldsymbol{\mu}_i^f, \boldsymbol{\Sigma}_i^f)$ be a Gaussian mixture model;
denote $\mathcal F_i :=\mathcal{N}(\boldsymbol{\mu}_i^f, \boldsymbol{\Sigma}_i^f)$ and $f_i$ as the probability density function (PDF) of $\mathcal F_i$. 
Accordingly, denote $\mathcal G := \sum_{j=1}^{K'} p^g_j \mathcal{N}(\boldsymbol{\mu}_j^g, \boldsymbol{\Sigma}_j^g)$, $\mathcal G_j :=\mathcal{N}(\boldsymbol{\mu}_j^g, \boldsymbol{\Sigma}_j^g)$, and $g_j$ as the PDF of $\mathcal G_j$. 
The lower bound of the KL divergence from $\mathcal G$ to $\mathcal F$ is
\begin{equation}
	D_{\textup KL}(\mathcal F||\mathcal G) \geq 
	\underbrace{
		\sum_{i=1}^{K} p_i^f \log \frac{\sum_{l=1}^K p_l^f \exp(-D_{\textup KL}(\mathcal F_i||\mathcal F_l))}{\sum_{j=1}^{K'} p_j^g t_{ij}} - \sum_{i=1}^K p_i^f H(\mathcal F_i)
	}_{\mathclap{D_{\text{lower}}(\mathcal F||\mathcal G)}}\,,
\end{equation}
where $t_{ij}$ is the normalization constant:
\begin{equation}
	t_{ij} := \int_{\mathcal{X}} f_i(\boldsymbol{x})g_j(\boldsymbol{x}) d\boldsymbol{x}\,,
\end{equation}
which is given by \citep{Ahrendt2005}:
\begin{equation}
	\log t_{ij} = -\frac{d}{2} \log 2\pi -\frac{1}{2} \log|\boldsymbol{\Sigma}_i^f + \boldsymbol{\Sigma}_j^g| -
	\frac{1}{2}(\boldsymbol{\mu}_j^g - \boldsymbol{\mu}_i^f)^T (\boldsymbol{\Sigma}_i^f + \boldsymbol{\Sigma}_j^g)^{-1}(\boldsymbol{\mu}_j^g - \boldsymbol{\mu}_i^f)\,,
\end{equation}
and $H(\cdot)$ is the entropy of Gaussian distribution:
\begin{equation}
	H(\mathcal F_i) = \frac{1}{2} \log (2\pi e)^d |\boldsymbol{\Sigma}_i^f|
\end{equation}
Similarly, the upper bound of the KL divergence from $\mathcal G$ to $\mathcal F$ is:
\begin{equation}
	D_{\textup KL}(\mathcal F||\mathcal G) \leq  \underbrace{
		\sum_{i=1}^{K} p_i^f \log \frac{\sum_{l=1}^K p_l^f t_{il}}{\sum_{j=1}^{K'} p_j^g \exp(-D_{\textup KL}(\mathcal F_i||\mathcal G_j))}
	}_{\mathclap{D_{\text{upper}}(\mathcal F||\mathcal G)}}\,. 
\end{equation}
\cite{Durrieu2012} proposed using the average of $D_{\text{upper}}$ and $D_{\text{lower}}$ to approximate $D_{\text{KL}}$:
\begin{equation}\label{eq:gmm_kl}
	D_{\text{approx}}(\mathcal F||\mathcal G) := \frac{D_{\text{upper}}(\mathcal F||\mathcal G) +  D_{\text{lower}}(\mathcal F||\mathcal G)}{2}\,.
\end{equation} 
The experimental results in \cite{Durrieu2012} show that  (\ref{eq:gmm_kl}) works well. 
Hence, we use $D_{\text{approx}}$ to approximate the two KL divergence functions in the definition of CKL \eqref{eq:CKL} in Appendix~\ref{sec:alternative_criteria}. 

\section{More simulation details}
\subsection{Additional results of the six criteria's performance on simulated datasets}
\par Here we present the performance of the six criteria on the simulated data. 
The datasets are generated by the procedures described in the main text Section~\ref{subsec:sim}.
Table~\ref{table:table-sim-6} shows the performance of the six criteria when the simulated datasets are generated with $K_0=6$.
The results are consistent with those in the main text Table~\ref{table:table-sim-8}; ITCA outperforms alternative class combination criteria and only failed when $K^*=2$.
Table~\ref{table:search-6} shows the proposed search strategies, namely greedy search and BFS, are almost as effective as the exhaustive search in finding the true class combinations where $K_0=8$.
Table~\ref{table:serach-large} shows that the two search strategies works well even when the number of observed classes $K_0$ is large.
\begin{table}[hpt]
	\centering 
	\caption{The performance of six criteria on the $31$ simulated datasets with $K_0=6$. The best result in each column is boldfaced.}\label{table:table-sim-6}
	\begin{threeparttable}
		\begin{tabular}{lrccrcc}
			\toprule
			\multirow{3}{*}{Criterion} & \multicolumn{1}{c}{\# successes} & Average & Max & \multicolumn{1}{c}{\# successes} & Average & Max\\
			\cmidrule(r){2-2} \cmidrule(r){5-5}
			& \multicolumn{1}{c}{\# datasets} & Hamming & Hamming & \multicolumn{1}{c}{\# datasets} & Hamming & Hamming\\
			\cmidrule(r){2-4} \cmidrule(r){5-7}
			& \multicolumn{3}{c}{LDA} &  \multicolumn{3}{c}{RF} \\
			\midrule
			ACC&1/31&2.03&4&1/31&2.03&4\\
			MI&8/31&1.42&4&6/31&1.65&4\\
			AAC&9/31&1.03&3&8/31&1.30&3\\
			CKL&7/31&2.42&5&1/31&2.13&4\\
			PE&22/31&0.55&3&22/31&0.42&3\\
			ITCA&\textbf{26/31}&\textbf{0.23}&\textbf{2}&\textbf{26/31}&\textbf{0.19}&\textbf{2}\\	
			\bottomrule
		\end{tabular}
	\end{threeparttable}
\end{table}
\begin{table}[htp]
	\centering
	\caption{Performance of ITCA using five search strategies and LDA on the $31$ simulated datasets with $K_0 = 6$.}
	\label{table:search-6}
	\begin{tabular}{lcccr}
		\toprule
		\multirow{2}{*}{Strategy} &  \# successes & Average & Max & Average \# class \\
		\cmidrule(r){2-2}
		& \# datasets & Hamming & Hamming & combinations examined\\
		\midrule
		Exhaustive &26/31&0.23&2&31.00 \\
		Greedy search&26/31 &0.23 &2 &12.13 \\
		BFS & 26/31 &0.19 &2 &19.19  \\
		Greedy (pruned)& 26/31 &0.19 &2 &5.71\\
		BFS (pruned) &26/31 &0.19 &2 &8.84 \\
		\bottomrule		
	\end{tabular}
\end{table}
\begin{table}[htp]
	\small
	\caption{Performance of ITCA using five search strategies and LDA on the $50$ simulated datasets with $K_0 = 20$.}
	\label{table:serach-large}
	\centering	
	\begin{tabular}{lcccr}
		\toprule
		\multirow{2}{*}{Strategy} &  \# successes & Average & Max & Average \# class \\
		\cmidrule(r){2-2}
		& \# datasets & Hamming & Hamming & combinations examined\\
		\midrule
		Greedy &50/50 &0.00 &0 &150.08 \\
		BFS & 50/50 &0.00 &0 &27226.84 \\
		Greedy (pruned)& 50/50 &0.00 &0 &87.70 \\
		BFS (pruned)& 50/50 &0.00 &0 &17155.82 \\
		\bottomrule
	\end{tabular}
\end{table}

\subsection{Additional results where ITCA missed the true class combination}
\par Table~\ref{table:table-sim-6} and Table~\ref{table:table-sim-8} in the main text have shown that ITCA is effective for finding the true class combination; ITCA using the LDA classification algorithm missed the true combination on 5 out of 31 datasets when $K_0=6$ and on 7 out of 120 datasets when $K_0=8$.
\par
Tables~\ref{table:failed_k0=6} and \ref{table:failed_k0=8} list the $5$ and $7$ true class combinations missed by ITCA using LDA for $K_0=6$ and $8$, respectively.
Notably, all these true class combinations belong to $K^*=2$, i.e., the scenario with only two combined classes, where obviously at least one combined class must have a proportion no less than $0.5$. 
Based on our theoretical analysis in Appendix \ref{sec:p-ITCA}, we know that ITCA would not combine two same-distributed classes when the combined class' proportion is large (see Appendix Figure~\ref{fig:p-ITCA_boudnary}, right panel). Hence, it is expected that ITCA is unlikely to find the true class combination when $K^*=2$. In other words, ITCA is unsuitable for combining observed classes, even if ambiguous, into a large class that dominates in proportion.

On the other hand, ITCA has successfully found the true class combinations when $K^* \ge 3$ for both $K_0=6$ and $8$, demonstrating its effectiveness.
\begin{table}[hpt]
	\centering
	\caption{True class combinations missed by ITCA in Table~\ref{table:table-sim-6} (simulation study with $K_0=6$).}\label{table:failed_k0=6}
	\begin{tabular}{ll}
		\toprule
		True combination&  ITCA-guided combination\\
		\midrule
		$\{(1, 2, 3, 4, 5), 6\}$& $\{1, 2, (3, 4), (5, 6)\}$\\
		$\{(1, 2, 3, 4), (5, 6)\}$ & $ \{1, 2, (3, 4), (5, 6)\}$\\
		$\{(1, 2, 3), (4, 5, 6)\}$& $\{1, 2, (3, 4), (5, 6)\}$ \\
		$ \{(1, 2), (3, 4, 5, 6)\}$ & $\{1, 2, (3, 4), (5, 6)\}$\\
		$\{1, (2, 3, 4, 5, 6)\}$ & $ \{1, 2, (3, 4), (5, 6)\}$\\
		\bottomrule
	\end{tabular}
\end{table}

\begin{table}[hpt]
	\centering
	\caption{True class combinations missed by ITCA with LDA in Table~\ref{table:table-sim-8} (simulation study with $K_0=8$).}\label{table:failed_k0=8}
	\begin{tabular}{ll}
		\toprule
		True combination&  ITCA-guided combination\\
		\midrule
		$ \{(1, 2, 3, 4, 5, 6, 7), 8\}$& $ \{1, 2, 3, 4, (5, 6), (7, 8)\}$\\
		$\{(1, 2, 3, 4, 5, 6), (7, 8)\}$ & $ \{1, 2, 3, 4, (5, 6), (7, 8)\}$\\
		$\{(1, 2, 3, 4, 5), (6, 7, 8)\}$& $ \{1, 2, 3, 4, (5, 6), (7, 8)\}$ \\
		$\{(1, 2, 3, 4), (5, 6, 7, 8)\}$ & $ \{1, 2, 3, 4, (5, 6), (7, 8)\}$\\
		$\{(1, 2, 3), (4, 5, 6, 7, 8)\}$ & $ \{1, 2, 3, 4, (5, 6), (7, 8)\}$\\
		$\{(1, 2), (3, 4, 5, 6, 7, 8)\}$ & $ \{1, 2, 3, 4, (5, 6), (7, 8)\}$\\
		$ \{1, (2, 3, 4, 5, 6, 7, 8)\}$ & $ \{1, 2, 3, 4, (5, 6), (7, 8)\}$\\
		\bottomrule
	\end{tabular}
\end{table}

\subsection{Comparison of the six criteria}
\par
\begin{figure}[hp]
	\centering
	\includegraphics[width=0.9\linewidth]{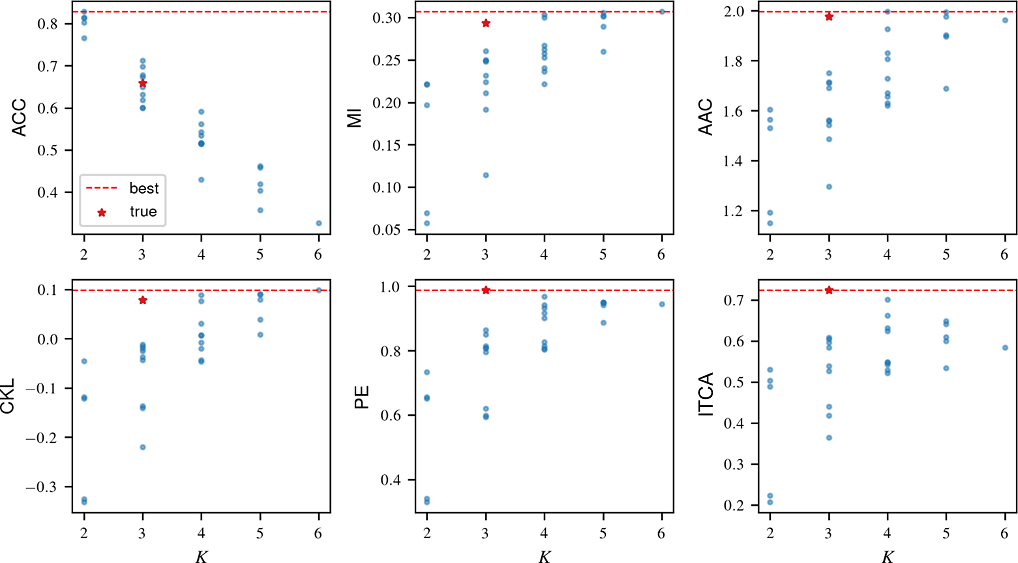}
	\caption{Comparison of ITCA and five other criteria using RF as the classification algorithm. The dataset is generated with $K_0=6$, $K^*=3$, $l=3$, $\sigma=1.5$, $n=2000$, and $d=5$. The true class combination is $\pi^*_3=\{(1, 2), (3, 4), (5, 6)\}$. For each criterion (panel), the $31$ blue points correspond to the $31$ class combinations $\pi_K$'s with $K=2,\ldots,6$. The true class combination $\pi_{K^*}^*$ is marked with the red star, and the best value for each criterion is indicated by a horizontal dashed line. The true class combination is only found by PE and ITCA without close ties.}
	\label{figure:sim_RF}
\end{figure}

 We also evaluate ITCA using random forest (RF) as the classification algorithm. 
For the true class combination $\pi_3^*=\{(1,2),(3,4),(5,6)\}$, the comparison results of ITCA versus the five alternative criteria using RF (Figure \ref{figure:sim_RF}) are consistent with those using LDA (Figure~\ref{fig:sim_LDA} in the main text).
\par
\begin{figure}[hp]
	\centering
	\includegraphics[width=0.9\linewidth]{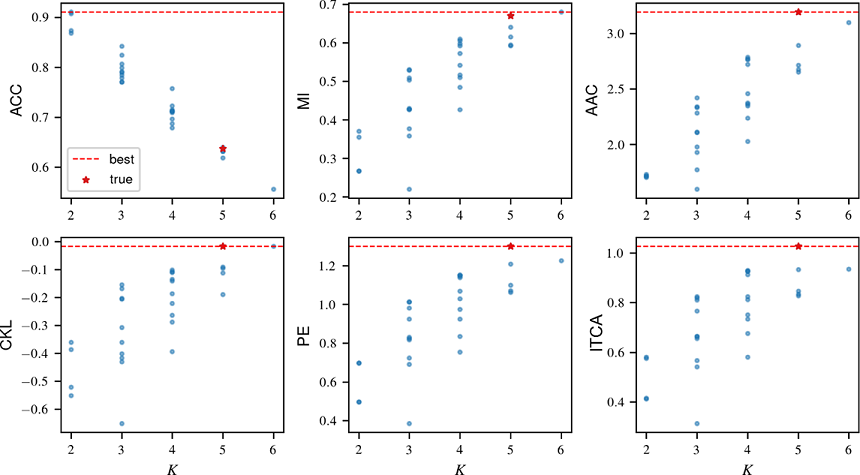}
	\caption{Comparison of ITCA and five other criteria using LDA as the classification algorithm. The dataset is generated with $K_0=6$, $K^*=5$, $l=3$, $\sigma=1.5$, $n=2000$, and $d=5$. The true class combination is $\pi^*_5=\{(1, 2), 3, 4, 5, 6\}$. For each criterion (panel), the $31$ blue points correspond to the $31$ class combinations $\pi_K$'s with $K=2,\ldots,6$. The true class combination $\pi_{K^*}^*$ is marked with the red star, and the best value for each criterion is indicated by a horizontal dashed line. The true class combination is found by AAC, PE and ITCA without close ties.}
	\label{figure:sim_K*=5}
\end{figure}

\begin{figure}[hp]
	\centering
	\includegraphics[width=0.9\linewidth]{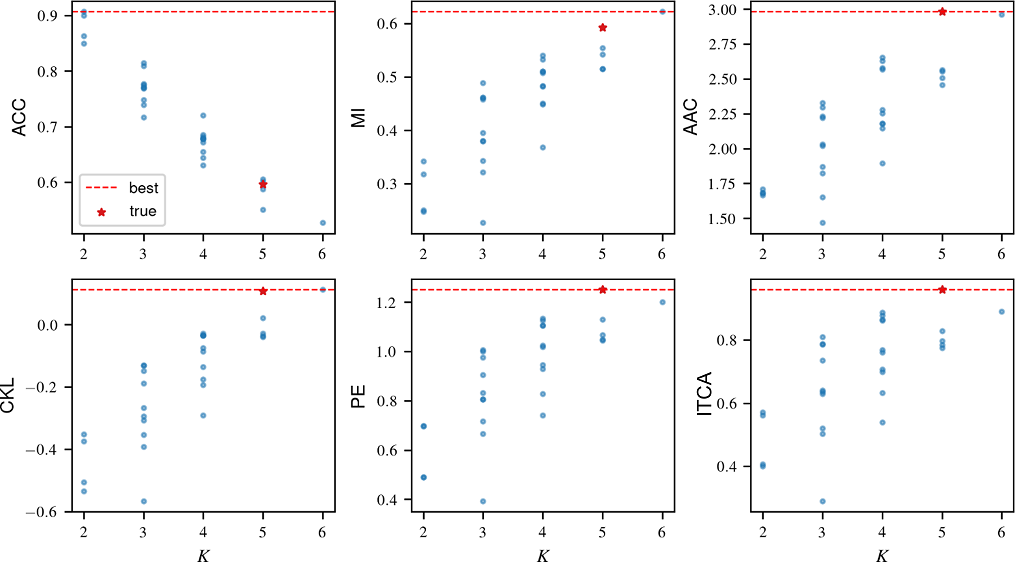}
	\caption{Comparison of ITCA and five other criteria using RF as the classification algorithm. The dataset is generated with $K_0=6$, $K^*=5$, $l=3$, $\sigma=1.5$, $n=2000$, and $d=5$. The true class combination is $\pi^*_5=\{(1, 2), 3, 4, 5, 6\}$. For each criterion (panel), the $31$ blue points correspond to the $31$ class combinations $\pi_K$'s with $K=2,\ldots,6$. The true class combination $\pi_{K^*}^*$ is marked with the red star, and the best value for each criterion is indicated by a horizontal dashed line. The true class combination is found by PE and ITCA without close ties.}
	\label{figure:sim_K*=5RF}
\end{figure}

 For another true class combination $\pi^*_5=\{(1, 2), 3, 4, 5, 6\}$, the results of the six criteria using LDA are shown in Figure \ref{figure:sim_K*=5}.
Among the six criteria, only AAC, PE and ITCA find the true class combination. 
ITCA outperforms the alternative criteria including PE by having the largest gap between the true class combination and the other class combinations. 
Specifically, the ITCA value of the true combination is 9\% higher than the value of the second-best class combination, while this improvement percentage is only 5.1\% for PE. 
When RF is used as the classification algorithm, the results stay consistent (Figure \ref{figure:sim_K*=5RF}).

\subsection{Alternative definition of the adjusted accuracy (AAC)}\label{sec:aac}
\par 
In Appendix~\ref{sec:alternative_criteria}, we define the AAC by assigning each (observed or combined) class the weight as the inverse of the class proportion. In other words, smaller classes receive larger weights because they are intuitively more difficult to predict.
Here we refer to this definition as ``AAC (proportion)'' for clarity. 

An alternative approach is to weigh each (observed or combined) class by the inverse of the number of observed classes it corresponds to. For example, an observed classes would have a weight of $1$, while a class combined from two observed classes would have a weight of $1/2$. The intuition is that a class is easier to predict if it is combined from more observed classes.
Hence, we refer to this alternative definition as ``AAC (cardinality)'': 
\begin{equation}
	\text{AAC (cardinality)}^{\text{CV}}(\pi_K; \mathcal{D}, \mathcal{C}) := \frac{1}{R} \sum_{r=1}^R \frac{1}{|\mathcal{D}_v^r|} \sum\limits_{(\bX_i, Y_i)\in \mathcal{D}_v^r} \frac{\1\left(\phi_{\pi_K}^{\mathcal C, \mathcal{D}_t^r}(\bX_i)=\pi_K(Y_i)\right)}{\left|\pi_K^{-1}(\pi_K(Y_i))\right|}\,,
\end{equation}
where the dataset $\mathcal{D}$ is randomly split into $R$ equal-sized folds, with the $r$-th fold $\mathcal{D}_v^r$ serving as the validation data and the union of the remaining $R-1$ folds $\mathcal{D}_t^r$ serving as the training data, and the denominator $\left|\pi_K^{-1}(\pi_K(Y_i))\right|$ indicates the number of observed classes contained in the combined class $\pi_K(Y_i)$.
In the following text, we refer to $\text{AAC (cardinality)}^{\text{CV}}$ as the AAC (cardinality) criterion.
\begin{figure}[hpt]
	\centering
	\includegraphics[width=0.75\linewidth]{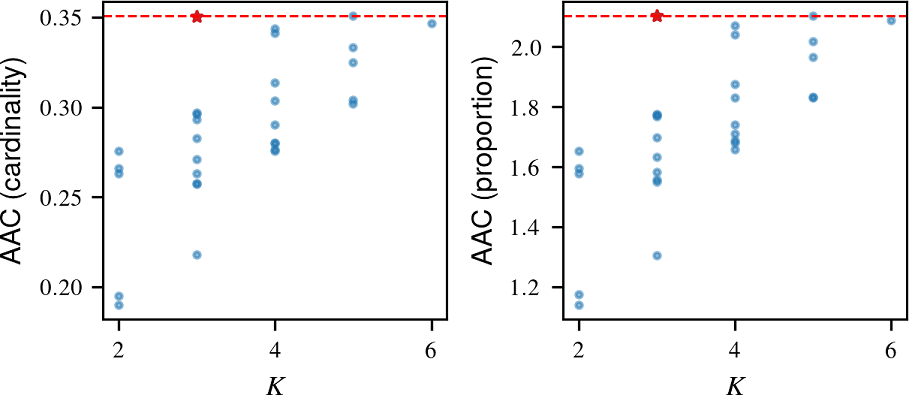}
	\caption{Comparison of AAC (cardinality) and AAC (proportion) using LDA as the classification algorithm. The dataset is generated with $K_0=6$, $K^*=5$, $l=3$, $\sigma=1.5$, $n=2000$, and $d=5$. The true class combination is $\pi^*_5=\{(1, 2), 3, 4, 5, 6\}$. 
		For each criterion (panel), the $31$ blue points correspond to the $31$ class combinations $\pi_K$'s with $K=2,\ldots,6$. The true class combination $\pi_{K^*}^*$ is marked with the red star, and the best value for each criterion is indicated by a horizontal dashed line.}
	\label{figure:sim_LDA_aac}
\end{figure}

\begin{table}[h]
	\centering 
	\caption{Comparison of two definitions of AAC using LDA on simulated datasets}\label{table:sim-lda}
	\begin{tabular}{lcccccc}
		\toprule
		\multirow{3}{*}{Criterion} & \# successes & Average & Max & \# successes & Average & Max\\
		\cmidrule(r){2-2} \cmidrule(r){5-5}
		& \# datasets & Hamming & Hamming & \# datasets & Hamming & Hamming\\
		\cmidrule(r){2-4} \cmidrule(r){5-7}
		& \multicolumn{3}{c}{$K_0=6$} &  \multicolumn{3}{c}{$K_0=8$} \\
		\midrule
		AAC (cardinality)&1/31&1.90&3&5/127&2.56&6\\
		AAC (proportion)&9/31&1.03&3&15/127&2.02&6\\
		\bottomrule
	\end{tabular}
\end{table} 
\par 
Figure \ref{figure:sim_LDA_aac} shows the comparison of AAC (cardinality) and AAC (proportion) for finding $\pi_5^*=\{(1,2),3,4,5,6\}$.
Table \ref{table:sim-lda} lists the overall comparison results of ACC (cardinality) and ACC (proportion) in the simulation studies with $K_0=6$ and $8$ (Section~\ref{subsec:sim} in the main text).
The results show that AAC (proportion) outperforms AAC (cardinality) by finding more true class combinations. Hence, we use AAC (proportion) in the main text.

\section{More application details}
\subsection{Prognosis of rehabilitation outcomes of traumatic brain injury patients}
\par 
The Casa Colina dataset includes Functional Independence Measure (FIM) of 17 activities at admission and discharge.
For each activity, the discharge FIM is coded as an ordinal outcome with $K_0=7$ levels by physical therapists.
Table \ref{table:casa17} lists the 17 activities.
\begin{table}[ht]
	\centering
	\caption{17 activities in the Casa Colina dataset}\label{table:casa17}
	\begin{tabular}{m{3cm} m{4cm}}
		\toprule
		Category                    & Activity\\
		\midrule
		& Eating                \\
		& Grooming              \\
		& Bathing               \\
		& Dressing - upper body \\
		& Dressing - lower body \\
		& Toileting             \\
		& Bladder control       \\
		& Bower control         \\
		& Transfer - bed        \\
		& Transfer - toilet     \\
		& Transfer - tub        \\
		\multirow{-12}{*}{Motor}    & Stairs                \\
		\midrule
		\multirow{5}{*}{Cognition}& Comprehension         \\
		& Expression            \\
		& Social interaction    \\
		& Problem solving       \\
		& Memory   \\          
		\bottomrule
	\end{tabular}
\end{table}

We use RF with 1000 trees as the classification algorithm (with default hyperparameters in the Python \texttt{sklearn} package \citep{Pedregosa2011}), and we compute the ITCA for all allowed class combinations.
Since there are seven ordinal outcomes, the number of allowed class combinations is  $2^{7-1} -1 = 63$.
\par
\begin{figure}[hp]
	\includegraphics[width=\textwidth]{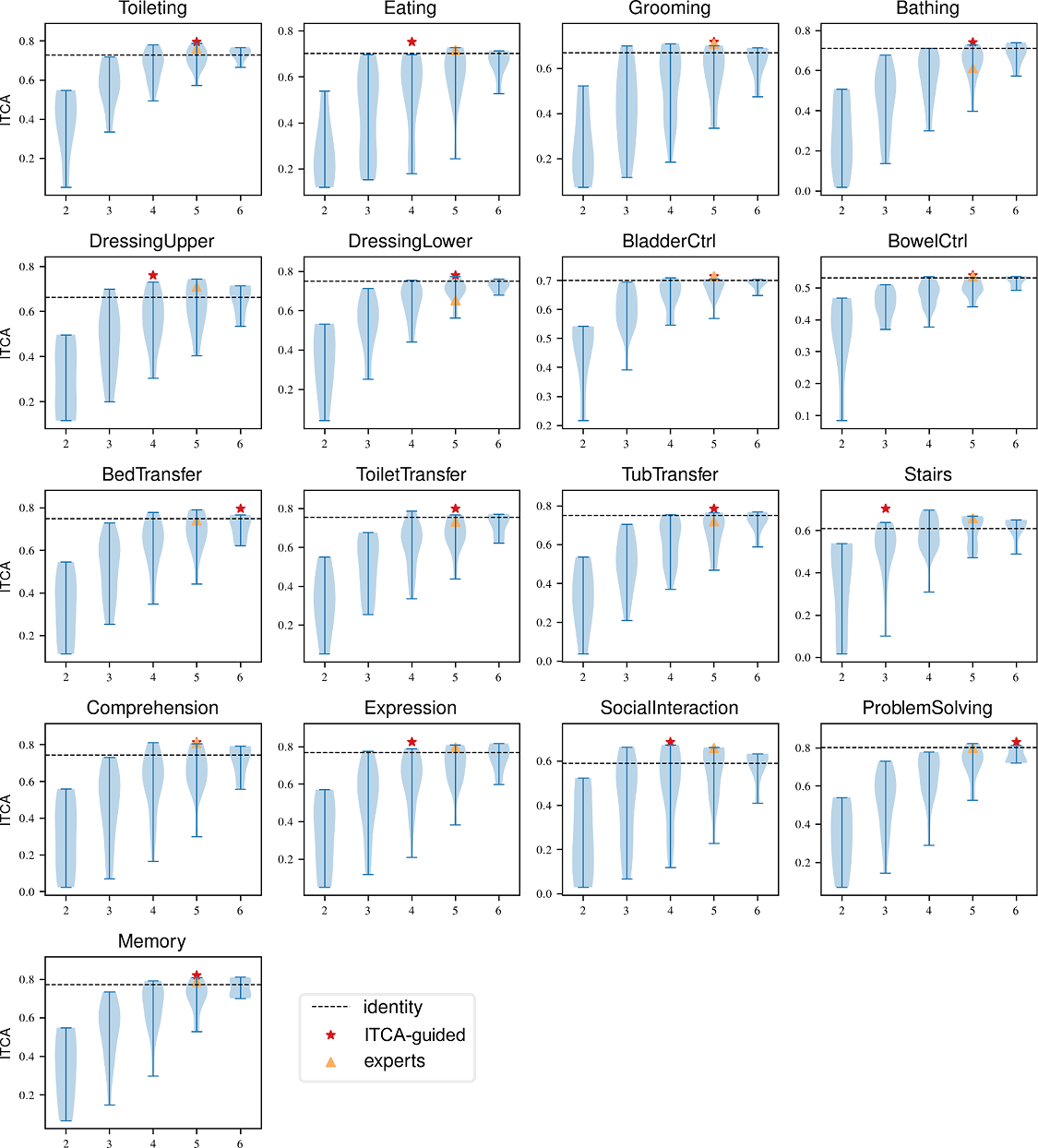}
	\caption{ITCA values of the 63 allowed class combinations for the 17 activities in the Casa Colina dataset. 
		The violin plots show the distribution of ITCA values for each combined class number $K$. 
		The dashed horizontal lines indicate the ITCA values of the identity class combination (i.e., $\pi_{7}$ with no classes combined).}
	\label{fig:casa_itca}
\end{figure}

Figure \ref{fig:casa_itca} shows the ITCA values of all allowed class combinations for each activity.
Interestingly, ITCA suggests that most activities should have their outcomes combined into four or five classes, largely consistent with the experts' suggestion (Table \ref{table:casa_selected}).
ITCA-guided class combination leads to larger-than-expected increases in the prediction accuracy for many activities, e.g., DressingUpper (Table \ref{table:casa_comparison}).
\begin{table}[hp]
	\centering
	\caption{Class combinations suggested by ITCA for the 17 activities.}
	\label{table:casa_selected}
	\begin{tabular}{llc}
		\toprule
		Activity &\multicolumn{1}{c}{$\pi^*_K$}& $K$\\
		\midrule
		Toileting&\{1, 2, (3, 4), 5, (6, 7)\}&5\\
		Eating&\{(1, 2, 3, 4), 5, 6, 7\}&4\\
		Grooming&\{1, (2, 3, 4), 5, 6, 7\}&5\\
		Bathing&\{1, (2, 3), 4, 5, (6, 7)\}&5\\
		DressingUpper&\{(1, 2), (3, 4), 5, (6, 7)\}&4\\
		DressingLower&\{1, (2, 3), 4, 5, (6, 7)\}&5\\
		BladderCtrl&\{1, (2, 3, 4), 5, 6, 7\}&5\\
		BowelCtrl&\{1, 2, (3, 4, 5), 6, 7\}&5\\
		BedTransfer&\{1, (2, 3), 4, 5, 6, 7\}&6\\
		ToiletTransfer&\{1, (2, 3), 4, 5, (6, 7)\}&5\\
		TubTransfer&\{1, (2, 3), 4, 5, (6, 7)\}&5\\
		Stairs&\{1, (2, 3, 4), (5, 6, 7)\}&3\\
		Comprehension&\{(1, 2), (3, 4), 5, 6, 7\}&5\\
		Expression&\{(1, 2, 3), (4, 5), 6, 7\}&4\\
		SocialInteraction&\{(1, 2, 3), (4, 5), 6, 7\}&4\\
		ProblemSolving&\{1, (2, 3), 4, 5, 6, 7\}&6\\
		Memory&\{(1, 2, 3), 4, 5, 6, 7\}&5\\
		\bottomrule
	\end{tabular}
\end{table}
\begin{table}[htp]
	\centering
	\caption{ACC and ITCA values of the original classes (identical class combination), expert-suggested class combination (the same $5$ combined classes for all activities), and ITCA-guided class combinations (listed in Table~\ref{table:casa_selected}; specific to each activity; the number of combined classes $K$ in the last column).}
	\label{table:casa_comparison}
	\resizebox{\textwidth}{!}{
		\begin{tabular}{lccccccc}
			\toprule
			& \multicolumn{3}{c}{ACC (\%)}&\multicolumn{3}{c}{ITCA (\%)}	\\
			\cmidrule(lr){2-4} \cmidrule(lr){5-7}
			Activity &Original&Experts&ITCA&Original&Experts&ITCA&$K$\\
			\midrule
			Toileting&44.12(1.26)&55.33(1.13)&52.50(1.05)&72.70(3.29)&75.76(3.16)&79.39(0.87)&5\\
			Eating&57.31(1.36)&57.08(1.18)&61.21(1.24)&70.14(3.94)&71.71(3.20)&75.35(0.54)&4\\
			Grooming&46.36(2.44)&52.44(1.08)&49.94(0.49)&66.91(2.20)&70.97(2.19)&71.73(3.99)&5\\
			Bathing&44.25(1.35)&62.18(1.16)&50.10(1.83)&71.15(1.76)&61.18(2.47)&74.21(2.35)&5\\
			DressingUpper&42.20(1.98)&51.62(1.30)&57.28(1.11)&66.33(3.07)&70.89(1.86)&76.11(3.85)&4\\
			DressingLower&39.76(3.69)&58.45(1.40)&49.74(1.05)&74.88(4.33)&65.23(2.96)&77.93(2.07)&5\\
			BladderCtrl&58.19(0.92)&57.67(2.01)&58.32(2.08)&70.06(3.09)&71.54(2.51)&71.54(2.51)&5\\
			BowelCtrl&65.14(1.43)&65.01(1.90)&64.81(0.79)&53.13(1.28)&53.56(1.27)&53.93(1.96)&5\\
			BedTransfer&42.85(2.21)&55.04(0.95)&46.75(1.46)&74.91(1.30)&74.04(1.54)&79.57(2.86)&6\\
			ToiletTransfer&45.87(1.53)&57.18(2.26)&51.27(0.72)&75.45(1.59)&73.19(2.99)&79.95(3.18)&5\\
			TubTransfer&48.80(0.98)&58.61(1.48)&50.42(1.21)&75.07(2.42)&71.97(2.39)&78.56(2.09)&5\\
			Stairs&52.79(1.26)&59.36(1.47)&65.72(1.02)&60.89(2.39)&65.79(0.87)&70.32(1.36)&3\\
			Comprehension&50.78(1.98)&58.80(1.36)&56.43(2.09)&74.42(1.62)&80.98(3.32)&81.41(2.32)&5\\
			Expression&48.21(1.72)&58.87(1.57)&62.48(1.18)&76.94(2.57)&79.60(2.92)&82.52(2.89)&4\\
			SocialInteraction&51.07(1.07)&56.40(1.80)&60.17(2.35)&59.01(2.81)&65.86(1.30)&68.63(3.93)&4\\
			ProblemSolving&45.03(2.17)&58.84(1.50)&50.03(1.62)&80.13(2.46)&79.98(3.70)&83.03(3.87)&6\\
			Memory&43.34(2.15)&60.04(2.35)&55.33(1.93)&77.23(2.25)&78.83(2.37)&82.02(4.01)&5\\	
			\bottomrule
		\end{tabular}
	}
\end{table}
\subsection{Prediction of glioblastoma cancer patients' survival time}\label{sec:gbm}
\par\noindent
\textbf{Description of the GBM survival dataset}.
The original data consists of 577 patients and 23 features.
The dataset contains diagnosis age, gender, gene expression subtypes, therapy and other clinical information.  
\par
\noindent
\textbf{Data processing}. 
We first drop the features less relevant to survival prediction, including ``Study ID'', ``Patient ID'', ``Sample ID'', ``Cancer Type Detailed'', ``Number of Samples Per Patient'', ``Oncotree Code'', ``Somatic Status'', ``Sample Type'' and ``Cancer Type''.
We then impute the missing entries by the feature means and use one-hot coding to represent the categorical features.
Since some patients received multiple types of therapies (out of $10$ types), we use $10$-dimensional binary vectors to indicate patients' therapy types. 
After data processing, there are 36 features in total.
\par
\begin{figure}[hpt]
	\centering
	\includegraphics[width=0.375\linewidth]{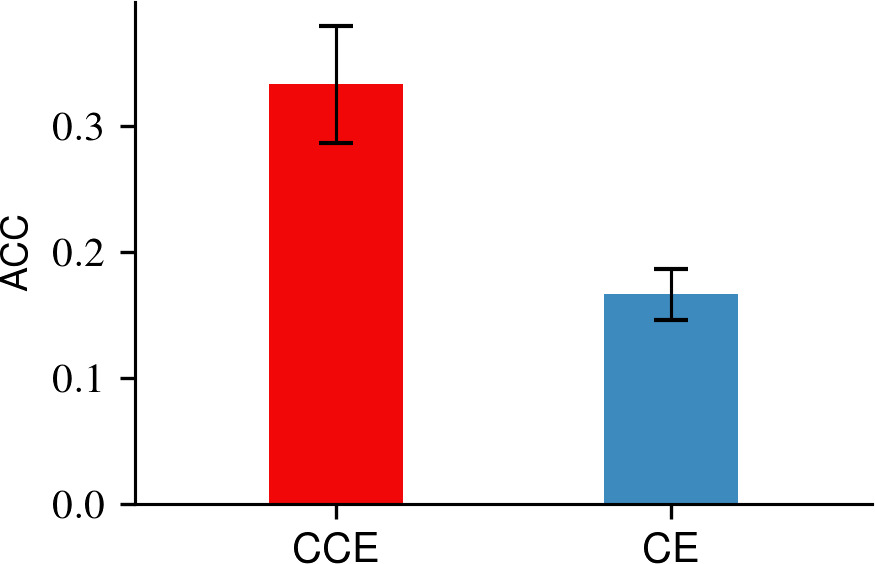}
	\caption{Performance of neural networks trained with the CCE and CE losses. 
		There are $K=12$ ordinal classes.
		The neural networks have three hidden layers with 40 hidden units and the ReLU activation function.}
	\label{fig:CCE_vs_CE}
\end{figure}

\noindent 
\textbf{Censored cross entropy loss function}.
The neural network is configured to output $K$ values, and it uses a softmax function to normalize the outputs as $K$ probabilities that sum to one.
The most commonly used loss function for classification is the cross entropy (CE) defined as
\begin{equation}
	\textup{CE} = -\sum_{i=1}^n\sum_{k=1}^K  \1(\pi_K(Y_i)=k) \log[\phi^{\text{NN}, \mathcal{D}_t}_{\pi_K}(\bX_i)]_k\,,
\end{equation}
where $\{(\bX_i, Y_i)\}_{i=1}^n$ is the validation dataset, $\mathcal{D}_t$ is the training dataset, $[\phi^{\text{NN},\mathcal{D}_t}_{\pi_K}(\bX_i)]_k$ is the $k$-th entry of the output vector of the neural network trained for predicting the combined classes indicated by $\pi_K$.
However, the survival data usually contain many right censored data.
The CE loss function is not compatible with such data.
To make full use of the censorship information, we introduce the censored cross entropy (CCE):
\begin{equation}\small
	\text{CCE} = -\smashoperator{\sum_{i=1}^n}\left[O_i \sum_{k=1}^K I(\pi_K(Y_i)=k)  \log[\phi^{\text{NN}, \mathcal{D}_t}_{\pi_K}(\bX_i)]_k + (1 - O_i) \smashoperator{\sum_{k > \pi_K(Y_i)}} \frac{\hat{p}_k}{1 - \sum_{l \leq \pi_K(Y_i)}\hat{p}_l} \log[\phi^{\text{NN}, \mathcal{D}_t}_{\pi_K}(\bX_i)]_k\right],
\end{equation}
where $O_i$ is binary with $O_i = 0$ indicating that $\bX_i$ is right censored, and $\hat{p}_k$ is the proportion of $\pi_K$'s $k$-th class in $\mathcal{D}$. 
When $\bX_i$ is not censored, its contribution to CCE is the same as to CE; 
when $\bX_i$ is censored, we compute its contribution to CCE as the cross entropy between the output sub-vector (for the classes later than $Y_i$) and the conditional distribution that the disease occurs later than $Y_i$.
The empirical results show that the neural network trained with the CCE loss outperforms that with the CE loss (Figure \ref{fig:CCE_vs_CE}).

\par
\noindent 
\textbf{Experiment setting}.
We implement the neural network with PyTorch \citep{Paszke2017}.
Specifically, we use a three-layered neural network with 40 hidden units each layer.
Experimental results show that the ReLU activation outperforms the sigmoid activation function. 
We use the SGD optimizer with 0.001 learning rate, 0.9 momentum and 0.01 weight decay.
Batch size is 64, and we stop the training after 150 epochs.

\subsection{Prediction of user demographics using mobile phone behavioral data}\label{sec:user}
\par
\begin{figure}
	\includegraphics[width=\linewidth]{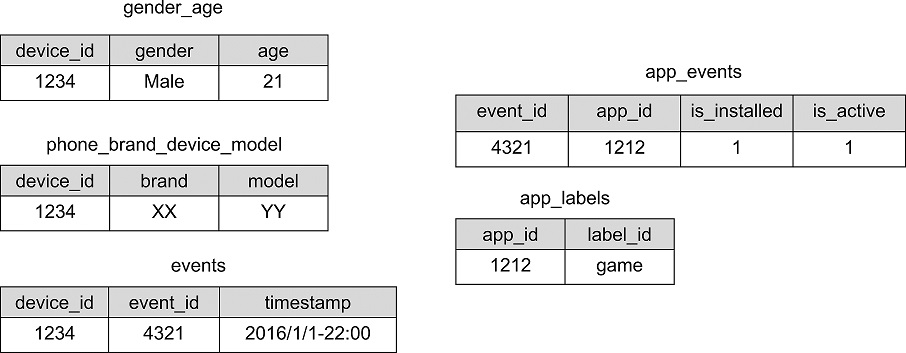}
	\caption{A view of the TalkingData mobile user demographics dataset. 
		We use five CSV files to construct the features and class labels.
		Here the five tables are excerpts from the five files, and only the relevant fields are listed.}\label{figure:talkingdata}
\end{figure}

\noindent
\textbf{Description of the TalkingData mobile user demographics dataset}.
This dataset includes several comma separate values (CSV) files as shown in Figure \ref{figure:talkingdata}.
Our training dataset contains 74645 unique device id.
The task is to predict a user's gender and age group (gender\_age table in Figure \ref{figure:talkingdata}) from their phone brand and the applications installed on their phone.
Readers may refer to the official website\footnote{\url{https://www.kaggle.com/c/talkingdata-mobile-user-demographics/data}} for a complete description of the dataset. 
\par\noindent
\textbf{Data processing}.
We first construct the device types by concatenating the mobile phone brand and model strings, resulting in around 1600 different types. 
We denote the device types that appear less than 50 times in the dataset as ``others".
There are  440 device types remaining.
\par
We then construct the features of the user behavioral data. 
Specifically, we use event logs to count which applications are installed on the device.
Since there are too many applications, we use the labels of applications instead (app\_label in Figure \ref{figure:talkingdata}).
We also count the users' earliest, latest, and most used time periods of phone usage every day (24 hours) by the events log.
The applications installed are represented by a binary vector, and the device types are encoded with one-hot encoding.
We note that the constructed feature vectors are very sparse and there are many users who do not have any activities.
We filter out the users whose features vectors have fewer than 5 nonzero values.
In summary, there are  $23{,}556$ users and 818 features after the data processing. 
\par\noindent
\textbf{Experiment setting}. 
We use the gradient boost decision tree model (GBT) in XGBoost \citep{Chen2016} as the classification algorithm.
GBT has several critical hyperparameters that may influence the prediction performance, including ``subsample" corresponding to the subsample ratio of the training instances, ``colsample\_bytree" corresponding to the subsample ratio of columns when constructing each tree, and ``gamma" corresponding to the partition on a leaf node.  
We adopt a greedy strategy to choose the best hyperparameters by the estimated accuracy on the test dataset.
Specifically, we first use grid search to find the best value of ``subsample" and then fix it to find the best value of ``colsample\_bytree".
We fix the three hyperparameters as  0.9, 1.0 and 0, respectively. The learning rate is 0.05 in all experiments.
\section{More theoretical remarks}

\subsection{Class-combination regions of the oracle and LDA classification algorithms}
\par 
Denote $f_1(p) := p^2 \log p $. The class-combination region of the oracle classification algorithm can be rewritten as
\begin{equation}\label{eq:oracle CR_2}
	\textup{CR}(\pi_{K_0-1} || \pi_{K_0}; \mathcal{C}^*) = \{(p_1, p_2) \in \Omega: 	f_1(p_1) + f_1(p_2) - f_1(p_{1+2}) > 0\}\,,
\end{equation} 
where $\Omega = \{(p_1, p_2): p_1>0, p_2>0, p_1+p_2<1\} \subset [0,1]^2$, and $p_{1+2} = p_1 + p_2$.

\begin{figure}[hpt]
	\centering
	\includegraphics[width=\linewidth]{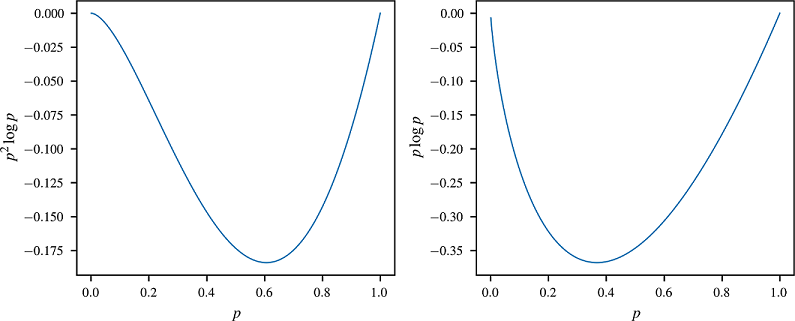}
	\caption{Plots of $f_1(p) = p^2 \log p$ and $f_2(p) = p \log p$ where $p \in (0, 1]$.}\label{fig:curves}
\end{figure}

Denote $f_2(p) := p \log p$. The class-combination region (as $||\boldsymbol{\mu}||/\sigma^2 \to \infty$) of the LDA classification algorithm can be rewritten as
\begin{equation}\label{eq:lda_CR_simple}
	\textup{CR}(\pi_{K_0-1} || \pi_{K_0}\,;\mathcal{D}_\infty, \mathcal{C}^{\textup{LDA}}) = \big\{(p_1, p_2) \in \Omega:\, f_2(p_{1 \vee 2}) - f_2(p_{1 + 2}) > 0 \big\}\,.
\end{equation} 
where $p_{1 \vee 2} := p_1 \vee p_2$. 
Figure~\ref{fig:curves} shows the plots of $f_1(p)$, and $f_2(p)$ where $p\in(0, 1]$.
$f_1(p)$ monotone decreases for $p \in (0, e^{-1/2})$ (left panel) and $f_2(p)$ monotone decreases for $p\in (0, e^{-1})$ (right panel).

\subsection{Class-combination curves and regions of other classification algorithms}\label{sec:three algorithms}
\par 
We also investigate the class-combination curves of three other commonly used classification algorithms, including random forest (RF), gradient boosting trees (GBT), and neural network (NN).
Specifically, we use RF with 50 trees, GBT with 20 trees, and a two-layer NN with 20 hidden units per layer and the ReLU activation function.
We use the same procedure to generate the simulated datasets as described in the Figure \ref{fig:s-ITCA_boudnary} in the Appendix. 
\begin{figure}[hpt]
	\includegraphics[width=\textwidth]{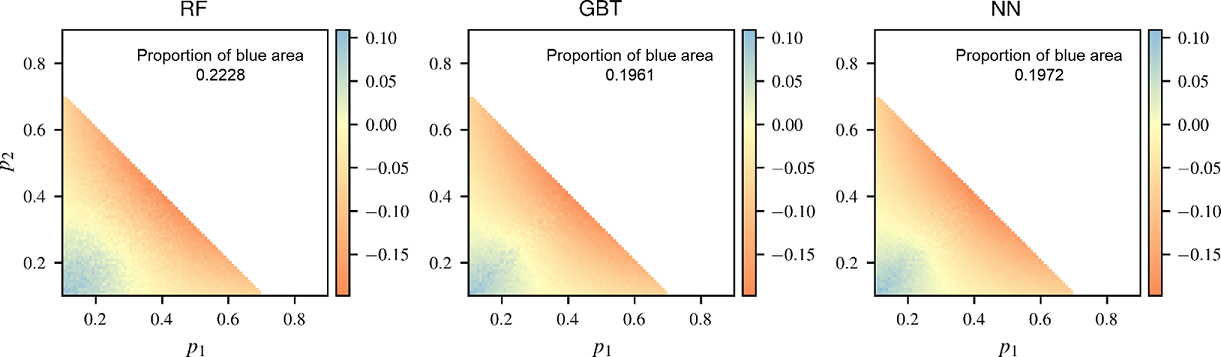}
	\caption{Regarding the combination of two same-distributed classes (with proportions $p_1$ and $p_2$), the improvement of ITCA , $\Delta \text{ITCA}(p_1, p_2; \mathcal{D}_t, \mathcal{C}) := \textup{ITCA}(\pi_{K_0-1}; \mathcal{D}_t, \mathcal{C}, p_1, p_2) -  \textup{ITCA}(\pi_{K_0}; \mathcal{D}_t, \mathcal{C}, p_1, p_2)$, of the RF algorithm (bottom left; $\mathcal{C}^{\textup{RF}}$), the GBT algorithm (middle; $\mathcal{C}^{\textup{GBT}}$) and the NN algorithm (right; $\mathcal{C}^{\textup{NN}}$). 
		The blue areas indicate the class-combination regions where $\Delta \text{ITCA}(p_1, p_2; \mathcal{D}_t, \mathcal{C}) > 0$ and thus the two classes will be combined. 
		In each panel, the yellow boundary between the orange area and the blue area is the class-combination curve of the corresponding algorithm. The proportion of the area of the class-combination region (the blue area) is shown in the upper right corner.}
	\label{fig:3clfs_boundary}
\end{figure}
\par 
The empirical results of the class-combination regions are shown in Figure \ref{fig:3clfs_boundary}.
We can see that the results of GBT and NN are similar to that of LDA.
Compared with GBT and NN, RF is more likely to find the true class combination (the proportion of the blue is the largest).
We note that the prediction of RF is based on the majority voting of decision trees.
Hence, intuitively, RF has a ``soft'' nature, putting it in the middle of the oracle classification algorithm and LDA.
\clearpage
\end{document}